\documentclass{article}

\usepackage[T1]{fontenc}
\usepackage[utf8]{inputenc}
\usepackage{microtype}
\usepackage{graphicx}
\usepackage{subcaption}
\usepackage{booktabs}
\usepackage{multirow}
\usepackage{makecell}
\usepackage{hyperref}
\usepackage{longtable}

\usepackage[accepted]{icml2026}

\usepackage{amsmath,amsfonts,bm}

\def\eqref#1{equation~\ref{#1}}
\def\1{\bm{1}}

\def\vx{{\bm{x}}}

\DeclareMathAlphabet{\mathsfit}{\encodingdefault}{\sfdefault}{m}{sl}
\SetMathAlphabet{\mathsfit}{bold}{\encodingdefault}{\sfdefault}{bx}{n}

\usepackage[capitalize,noabbrev]{cleveref}
\usepackage{caption}
\usepackage{fvextra}
\usepackage{paracol}
\usepackage{listings}
\usepackage{xcolor}
\usepackage[frozencache, cachedir=_minted]{minted}
\usepackage{threeparttable}
\usepackage{algorithm}
\usepackage{algorithmic}

\definecolor{bg}{rgb}{0.95,0.95,0.95}
\icmltitlerunning{ImprovEvolve: Basin-Hopping Meets LLM-Guided Evolutionary Search}

\begin{document}

\twocolumn[
\icmltitle{ImprovEvolve: Basin-Hopping Meets LLM-Guided Evolutionary Search}

\begin{icmlauthorlist}
\icmlauthor{Alexey Kravatskiy}{miri}
\icmlauthor{Valentin Khrulkov}{fusion,axxx}
\icmlauthor{Ivan Oseledets}{inm,axxx}
\end{icmlauthorlist}

\icmlaffiliation{miri}{MIRIAI, Russia}
\icmlaffiliation{fusion}{FusionBrain Lab, Russia}
\icmlaffiliation{axxx}{AXXX, Russia}
\icmlaffiliation{inm}{Institute of Numerical Mathematics, Russia}

\icmlcorrespondingauthor{Alexey Kravatskiy}{kravatskii.a@miriai.org}

\icmlkeywords{genetic programming, optimization, evolutionary computation, large language models, AlphaEvolve, mathematical discovery}

\vskip 0.3in
]

\printAffiliationsAndNotice{}

\newcommand{\alphaevolve}{\emph{AlphaEvolve}}
\newcommand{\improvevolve}{\emph{ImprovEvolve}}
\newcommand{\improvevolvepluse}{\emph{ImprovEvolve+E}}
\newcommand{\improvevolvextwo}{\emph{ImprovEvolve++}}
\newcommand{\gigaevo}{\emph{GigaEvo}}
% Value with citation on the following line (saves horizontal space in tables)
\newcommand{\valcite}[2]{\begin{tabular}[t]{@{}c@{}}#1\\[-0.1em]{\scriptsize\cite{#2}}\end{tabular}}

\begin{abstract}

LLM-guided evolutionary computation, most notably \alphaevolve, has been remarkably successful in discovering novel mathematical constructions by solving challenging optimization problems. The standard approach is to evolve a monolithic program that directly outputs a candidate solution. We present \textbf{ImprovEvolve}, an algorithmic alternative that drastically reduces cognitive load on the LLM. Instead of prompting the model for an end-to-end optimizer, we evolve a program with three specialized operators of initialization, local improvement, and perturbation. We then approach the optimum by iteratively applying local improvements and intensity-scheduled perturbations, effectively driving a basin-hopping search with LLM-evolved subroutines. For hexagon in hexagon packing, \improvevolve\ discovers new state-of-the-art packings of 11, 12, 15, and 16 hexagons, and additionally for 14, 17, and 23 hexagons after minimal expert tuning of the generated code. For the second autocorrelation inequality, the evolved and human-scaled program pushes the lower bound from $0.96102$ to $0.96258$. For spherical codes, the \improvevolve\ program lowers the best-known maximum cosine for the majority of 90 randomly chosen diverse state-of-the-art spherical codes, achieving relative improvements of up to 2.4\%.
\end{abstract}

\section{Introduction}

\begin{figure*}
    \centering
    \includegraphics[width=0.95\linewidth]{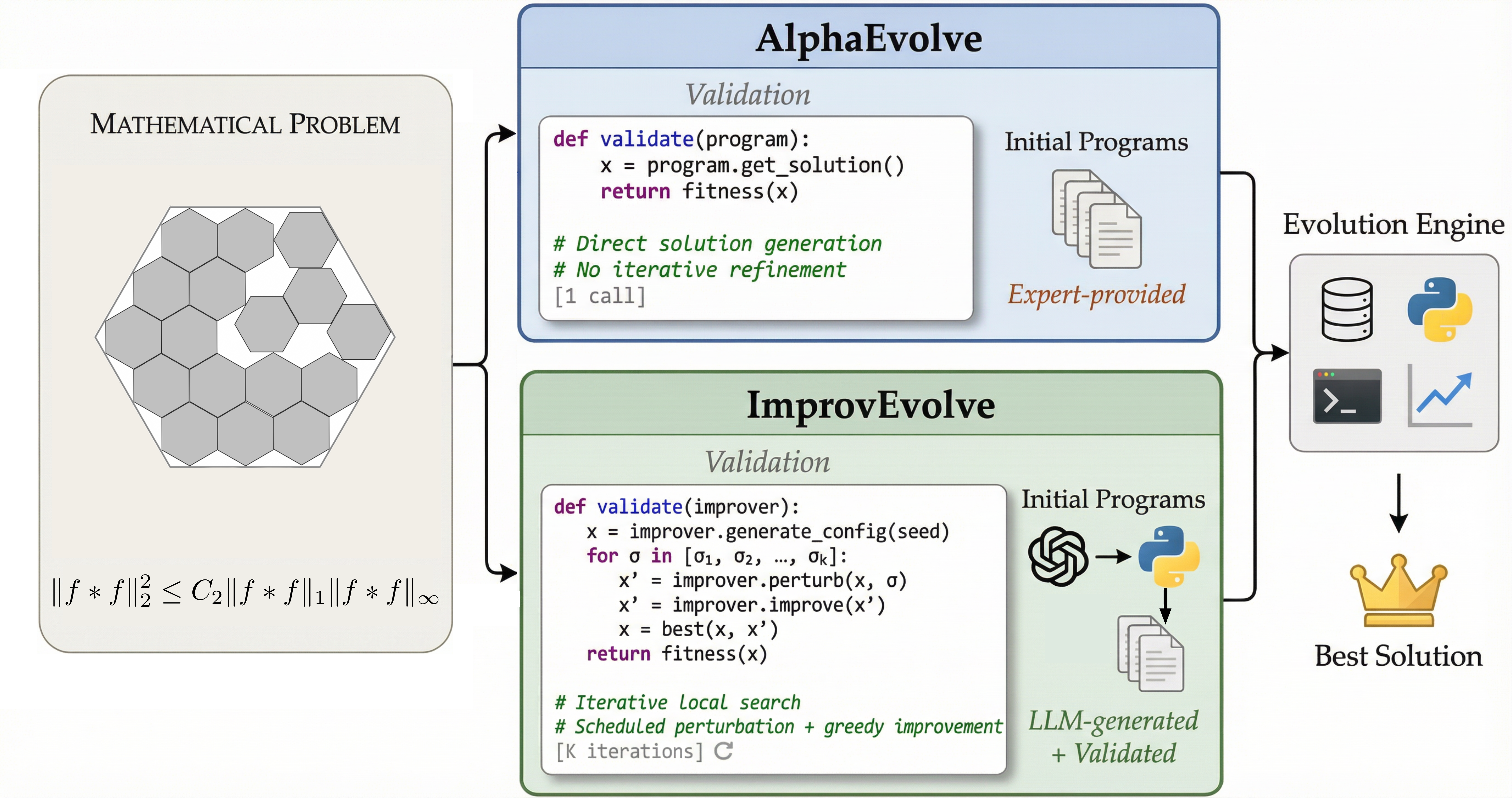}
    \caption{\textbf{Comparison of \alphaevolve\ and \improvevolve.} \emph{Top:} In \alphaevolve, the LLM evolves a program that directly outputs a candidate solution, requiring the model to design an end-to-end optimization algorithm including initialization, search strategy, and termination criteria. \emph{Bottom:} In \improvevolve, the LLM evolves a class with three modular methods: \texttt{generate\allowbreak\_config} (initialization), \texttt{improve} (local optimization), and \texttt{perturb} (exploration), which are combined via basin-hopping dynamics with a scheduled perturbation intensity. This decomposition reduces the cognitive burden on the LLM by separating distinct optimization concerns into tractable subproblems. The pseudocode shown for \improvevolve\ is simplified; see Algorithm~\ref{alg:validation} for the full description.}
    \label{fig:diagram}
\end{figure*}

\alphaevolve~\cite{novikov2025alphaevolve} combines powerful Large Language Models (LLMs) with the classical evolutionary algorithm MAP-Elites~\cite{mouret2015illuminatingsearchspacesmapping} to solve mathematical optimization problems. Given a task, the user specifies a validation function and a set of seed programs; candidate solutions are then iteratively refined by selecting promising parents from the archive and recombining them with an LLM-based mutation operator. The original implementation has not been released, but several open-source reimplementations are available~\cite{openevolve,lange2025shinkaevolve,assumpção2026codeevolveopensourceevolutionary,khrulkov2025gigaevo}.

This work focuses on the ability of LLMs to generate optimization code for problems with extremely rugged landscapes, whose optima often exhibit intricate structure, a phenomenon familiar from packing problems. Designing an effective optimizer \emph{ab initio} is difficult for an LLM: it must produce an end-to-end procedure together with a non-trivial initialization scheme. In practice, generated programs lean on off-the-shelf routines such as \texttt{L-BFGS-B}, which are inadequate for the non-convex, non-smooth landscapes that arise once all constraints are enforced. As a result, LLM-generated programs blend mathematical insight with optimization heuristics rather than rigorously solve the optimization problem.

When such a program yields only an imperfect solution, or when a non-trivial solution has already been constructed by domain experts, local search can usually improve it. This motivates \textbf{ImprovEvolve}: instead of evolving a program that directly outputs an optimal solution (as in \alphaevolve), we evolve a class with the following interface:
\begin{itemize}
    \item \texttt{improve}$(x)$: returns $x'$ with fitness at least that of $x$;
    \item \texttt{perturb}$(x, \sigma)$: randomly explores the search space at intensity $\sigma$;
    \item \texttt{generate\allowbreak\_config}$(seed)$: returns a feasible initial solution from a given random seed.
\end{itemize}
We then approach the optimum as follows: sample initial configurations with \texttt{generate\_config}, refine each via \texttt{improve} and select the best, then iteratively apply \texttt{perturb} and \texttt{improve} with a scheduled intensity, greedily accepting only improvements (\Cref{fig:diagram}).

This decomposition offers three advantages. First, it reduces the LLM's cognitive load by separating initialization, local improvement, and exploration into distinct concerns, each with a natural slot for domain knowledge. Second, the modular structure makes evolved programs easier for both humans and LLMs to interpret and debug. Third and most important, the evolved class can be applied to \emph{any} starting solution, such as one provided by a domain expert and representing a sophisticated construction beyond the LLM's immediate grasp.

\section{Related Work}\label{sec:related}

Evolutionary search has a long history in machine learning. Most directly relevant to this work, \citet{romera2024mathematical, novikov2025alphaevolve} introduced \emph{FunSearch} and \alphaevolve\ to evolve programs that solve mathematical problems, with mutations guided by powerful LLMs such as Gemini 2.5 Pro~\cite{comanici2025gemini}. \emph{FunSearch} evolves a single Python function that constructs a solution; \alphaevolve\ generalises this to arbitrary mutable code blocks with richer evolutionary feedback. \citet{georgiev2025mathematical} subsequently evaluated \alphaevolve\ at scale on cutting-edge mathematical challenges.

\alphaevolve\ has inspired numerous open-source implementations and extensions~\citep{openevolve,lange2025shinkaevolve,khrulkov2025gigaevo,assumpção2026codeevolveopensourceevolutionary,wang2025thetaevolve,yan2026pacevolve,yuksekgonul2026learning}. Most of these refine the evolutionary loop or replace it altogether, e.g., via reinforcement learning~\citep{wang2025thetaevolve,yuksekgonul2026learning}. We instead address a more fundamental obstacle: the difficulty of asking an LLM to generate effective optimization code at all. We do so by reformulating the program the LLM is asked to produce; to our knowledge, this is the first work to apply such a reformulation in the context of LLM-guided evolution. As the backbone of \improvevolve, we adopt the open-source \gigaevo\ framework\footnote{\url{https://github.com/FusionBrainLab/gigaevo-core}} of \citet{khrulkov2025gigaevo}, chosen for its flexibility and its comprehensive collection of pre-implemented mathematical problems.

A complementary line of work clarifies the difficulty of these landscapes from the classical-optimization side: \citet{berthold2026solvers} cast several combinatorial geometry problems (including hexagon packing) as nonlinear programs and obtained state-of-the-art results using commercial solvers such as FICO Xpress and SCIP. We tackle the same problems from the opposite direction, showing that incorporating classical global-optimization techniques, namely basin-hopping, directly into an LLM-guided evolutionary framework yields superior solutions; with lightweight human adjustments, \improvevolve\ matches or exceeds these formalized global-solver outputs.

\section{Methodology}\label{sec:method}
Our approach reformulates what the LLM is asked to produce so that classical global-optimization machinery does the heavy lifting on the search side, leaving domain-specific structure to the evolved code.

\paragraph{Connection to Basin-Hopping.}
Basin-hopping~\cite{wales1997basinhopping} is a well-known global-optimization technique, particularly effective for high-dimensional landscapes with many local minima separated by large barriers. The algorithm iterates (1)~a random perturbation of the current solution, (2)~local optimization from the perturbed point, and (3)~Metropolis acceptance at temperature~$T$; at $T = 0$ it reduces to \emph{Monotonic Basin-Hopping}, accepting only improvements.

\improvevolve\ can be viewed as a variant of basin-hopping in which the perturbation and local-optimization subroutines are themselves \emph{evolved} by an LLM rather than fixed generic operators: \texttt{perturb}$(x,\sigma)$ replaces the random coordinate step (with $\sigma$ controlling intensity), \texttt{improve}$(x)$ replaces the local minimizer (e.g., L-BFGS), and we use $T=0$ (monotonic acceptance) to accelerate convergence during validation. This shifts the LLM's task from designing an end-to-end optimizer to producing domain-specific \texttt{improve} and \texttt{perturb} operators, thereby setting a more tractable objective that directly leverages the LLM's ability to encode mathematical structure and heuristics.

\paragraph{Validation Scheme.}
The validation scheme for \improvevolve\ (Algorithm~\ref{alg:validation}) proceeds in two stages.
In \emph{Stage~A}, $K$ candidate solutions are generated with distinct random seeds and immediately refined via \texttt{improve}; the highest-fitness candidate becomes the starting point~$\vx^*$. Because the final result depends heavily on which basin the search starts in, sampling multiple initial configurations and keeping the best of them reduces sensitivity to poor initializations.
In \emph{Stage~B}, the algorithm performs $R$ rounds of perturbation and improvement. Within each round, the perturbation intensity~$\sigma$ decays geometrically from $\sigma_{\max}$ to $\sigma_{\min}$ over $M$ iterations, encouraging broad exploration early (large perturbations to escape local minima) and fine-grained refinement later (small perturbations to converge inside the current basin). Each perturbed solution is refined via \texttt{improve} and accepted only if it strictly improves the current best (monotonic acceptance, $T=0$). Restarting the $\sigma$ schedule on every round lets the search repeatedly attempt large jumps while still refining the incumbent.

\begin{algorithm}[tb]
  \caption{Validation Scheme for \improvevolve}
  \label{alg:validation}
  \begin{algorithmic}
    \STATE \textbf{Input:} Evolved class with methods \texttt{generate\_config}, \texttt{improve}, \texttt{perturb}; evaluation function \texttt{fitness}; number of rounds $R$
    \STATE \textbf{Output:} Solution $\vx^*$ with highest fitness found
    \STATE \COMMENT{Hyperparameters: $K \gets 10$, $M \gets 10$, $\sigma_{\max} \gets 100$, $\sigma_{\min} \gets 0.001$}
    \STATE \COMMENT{Stage A: Initialization}
    \FOR{$s = 1$ \textbf{to} $K$}
      \STATE $\vx_s \gets \texttt{improve}(\texttt{generate\_config}(s))$
    \ENDFOR
    \STATE $\vx^* \gets \arg\max_{s \in \{1,\ldots,K\}} \texttt{fitness}(\vx_s)$
    \STATE \COMMENT{Stage B: Basin-hopping with scheduled perturbation}
    \FOR{round $= 1$ \textbf{to} $R$}
      \FOR{$t = 1$ \textbf{to} $M$}
        \STATE $\sigma \gets \sigma_{\max} \cdot (\sigma_{\min} / \sigma_{\max})^{(t-1)/(M-1)}$
        \STATE $\vx' \gets \texttt{improve}(\texttt{perturb}(\vx^*, \sigma))$
        \IF{$\texttt{fitness}(\vx') \geq \texttt{fitness}(\vx^*)$}
          \STATE $\vx^* \gets \vx'$ \COMMENT{monotonic acceptance}
        \ENDIF
      \ENDFOR
    \ENDFOR
    \STATE \textbf{return} $\vx^*$
  \end{algorithmic}
\end{algorithm}

\paragraph{Initialization Strategy.}
To accelerate evolution without human assistance, we use \emph{Gemini~3~Pro} to generate $N=5$ initial programs and iteratively refine them in the same chat to fix validator errors (e.g., out-of-bounds configurations). Since running everything with a top-tier model is prohibitively expensive~\citep{georgiev2025mathematical}, we adopt a two-tier approach: \emph{Gemini~3~Pro} for initialization, and the faster \emph{Gemini~3~Flash~Preview} for evolution. The initial programs produced \emph{from the same prompt} already exhibit a diverse set of approaches and are competitive out of the box, consistent with~\citet{gideoni2025random}.

\paragraph{Evolutionary setup.}
We build on the open-source \gigaevo\ framework~\cite{khrulkov2025gigaevo}, which provides a modular evolutionary engine combining a Redis-backed candidate store, an asynchronous DAG execution engine for evaluation pipelines, a MAP-Elites loop with configurable island topologies, and an LLM-based mutation operator. We use a \emph{single-island} MAP-Elites configuration with fitness as the sole behavior descriptor, discretized into $150$ bins; this untuned resolution proved sufficient for state-of-the-art results on all benchmarks, while preliminary multi-island experiments (e.g., with code complexity as a second descriptor) offered no benefit and only added system complexity. We use \gigaevo's default mutation pipeline with system prompts unchanged from the repository: \texttt{InsightsStage} (LLM-generated improvement hints), \texttt{LineageInsights} (parent--child fitness feedback), and aggregate statistics.

It is useful to distinguish two nested levels of optimization. The \emph{outer loop} (MAP-Elites) maintains a population of candidate \emph{programs}, each implementing the \texttt{generate\_config}/\texttt{improve}/\texttt{perturb} interface. The \emph{inner loop} (Algorithm~\ref{alg:validation}) executes a single candidate via basin-hopping to produce a \emph{solution}, and assigns the program a fitness equal to the best solution quality found during that run. For \gigaevo\ baselines, each program is invoked in a single call to directly produce a candidate solution.
At each generation, $N_{\text{elites}}$ programs are sampled from the archive with probability proportional to fitness, and $N_{\text{offspring}}$ offspring are produced by passing $N_{\text{parents}}$ uniformly sampled elites to the LLM, together with their metrics and the pipeline context above; each offspring is evaluated by Algorithm~\ref{alg:validation} and inserted into the archive if its fitness warrants. The number of basin-hopping rounds~$R$ controls the trade-off between evaluation cost and solution quality: we use a smaller~$R$ during evolution for rapid fitness estimation and a larger~$R$ for final validation; problem-specific values appear below.

\paragraph{Computational cost.}
The wall-clock time of a single evolutionary run is dominated by the cost of each \texttt{improve} call. For hexagon packing, where the inner optimization is a moderate-dimensional constrained problem ($3n$ variables), a full run takes roughly 10 hours on a single machine. The second autocorrelation inequality is considerably more expensive: each \texttt{improve} call performs high-dimensional L-BFGS optimization over step functions with up to tens of thousands of parameters, giving runs of roughly 40 hours.

\section{Benchmark problems}
We evaluate \improvevolve\ on three challenging optimization problems: hexagon packing, the second autocorrelation inequality, and spherical codes.

As the backbone LLM we use \textit{Gemini 3 Pro} (via chat at \url{aistudio.google.com}) to generate initial programs, and \textit{Gemini 3 Flash Preview} (via \url{openrouter.ai}, default settings, temperature $T=1$) to run the evolution. Common evolution parameters are $N_{\text{elites}}=6$, $N_{\text{parents}}=2$, $N_{\text{offspring}}=10$, and all programs run on CPU to allow parallelization. During evolution, a program is discarded if its \texttt{improve} function returns an invalid configuration; this requirement is relaxed during final validation, since even reliable optimizers can occasionally diverge after an unlucky perturbation over an extended run.

\begin{table}[t]
    \centering
    \begin{threeparttable}
        \caption{Comparison of best known results for hexagon packing (HEX $n$) and the second autocorrelation inequality (ACI 2).}
        \label{tab:new_sota_transposed}
        \footnotesize
        \setlength{\tabcolsep}{4pt}
        \begin{tabular}{@{}lccc@{}}
            \toprule
            \textbf{Method} & \textbf{HEX 11} ($\downarrow$) & \textbf{HEX 12} ($\downarrow$) & \textbf{ACI 2} ($\uparrow$) \\
            \midrule
            \textbf{Human} & 3.9434 & 4.0000 & 0.94136 \\[-0.1em]
            & \multicolumn{2}{c}{\scriptsize\cite{friedman2025packing}} & {\scriptsize\cite{jaech2025secondcorrhuman}} \\[0.3em]
            
            \textbf{\alphaevolve} & 3.9301 & 3.9419 & 0.96102 \\[-0.1em]
            & \multicolumn{2}{c}{\scriptsize\cite{novikov2025alphaevolve}} & {\scriptsize\cite{georgiev2025mathematical}} \\[0.3em]
            
            \textbf{\emph{ThetaEvolve}} & -- & -- & 0.94690 \\[-0.1em]
            {\scriptsize\cite{wang2025thetaevolve}} & & & \\[0.3em]
            
            \textbf{\emph{CodeEvolve}} & 3.9379 & 4.0000 & 0.88110 \\[-0.1em]
            {\scriptsize\cite{assumpção2026codeevolveopensourceevolutionary}} & & & \\[0.3em]

            \textbf{Global Solvers} & 3.9249 & 3.9417 & -- \\[-0.1em]
            {\scriptsize\cite{berthold2026solvers}} & & & \\[0.3em]

            \textbf{\gigaevo} (baseline) & 3.9296 & invalid\tnote{1} & 0.9478 \\[-0.1em]
            {\scriptsize\cite{khrulkov2025gigaevo}} & & & \\[0.3em]
            
            \textbf{\improvevolve} & \textbf{3.9245} & \textbf{3.9416}\tnote{2} & 0.9512 \\[0.3em]
            \textbf{\improvevolvepluse}\tnote{3} & \textbf{3.9245} & \textbf{3.9416} & \textbf{0.96258} \\
            \bottomrule
        \end{tabular}
        \begin{tablenotes}
            \scriptsize
            \item[1] Hexagon overlap encountered when validating program evolved for $n = 11$ on $n = 12$.
            \item[2] Value obtained by validating code evolved for $n = 11$ on $n = 12$.
            \item[3] \improvevolvepluse: the evolved program was lightly edited by a human expert. For HEX, the optimizer and convergence parameters were adjusted (see \Cref{sec:hex_improver_code}); for ACI 2, the program was modified to support an effective start from \alphaevolve's solution (see \Cref{sec:autocorr_improver_code}).
        \end{tablenotes}
    \end{threeparttable}
\end{table}

\subsection{Hexagon packing}
\begin{table}[t]
    \centering
    \caption{\improvevolve\ hyperparameters for HEX~11 and ACI~2 problems.}
    \label{tab:hyperparameters}
    \scriptsize
    \setlength{\tabcolsep}{1.5pt}
    \begin{tabular}{@{}l@{\hspace{2pt}}c@{\hspace{2pt}}c@{}}
        \toprule
        \textbf{Parameter} & \textbf{HEX 11} & \textbf{ACI 2} \\
        \midrule
        Timeout & 5 min & 20 min \\
        Init.\ seeds ($K$) & 10 & 3 \\
        BH rounds ($R$) & 15 & 5 \\
        Iter.\ per round ($M$) & 11 & 6 \\
        $\sigma$ schedule &
        \begin{tabular}[t]{@{}l@{}}
            \texttt{[1e2, 5e1, 1e1, 5,} \\
            \texttt{1, 5e-1, 1e-1, 5e-2,} \\
            \texttt{1e-2, 5e-3, 1e-3]}
        \end{tabular}
        &
        \begin{tabular}[t]{@{}l@{}}
            \texttt{[1e2, 1e1,} \\
            \texttt{1, 1e-1,} \\
            \texttt{1e-2, 1e-3]}
        \end{tabular} \\
        \bottomrule
    \end{tabular}
\end{table}

The hexagon packing (HEX $n$) problem asks to pack $n$ unit regular hexagons (circumradius $r = 1$) inside a flat-topped regular enclosing hexagon of side length~$L$, minimising~$L$ subject to a no-overlap constraint. Each hexagon is parameterised by its center $(x_i, y_i)$ and rotation angle $\theta_i$, giving a $3n$-dimensional search space. Following~\citet{novikov2025alphaevolve} and~\citet{georgiev2025mathematical}, we evolve programs for $n = 11$---where the best previously known side length was $L = 3.9434$~\citep{friedman2025packing}, subsequently improved to $L = 3.9301$ by \alphaevolve~\citep{novikov2025alphaevolve}---and validate the evolved programs across a range of $n$.

The fitness is $-L$, with any containment or non-overlap violation raising an exception and discarding the configuration. The optimization landscape is highly non-convex: a small perturbation in position or rotation can introduce an overlap or a containment violation, making the problem particularly hard for evolutionary search. Hexagon packing is nevertheless an effective testbed for our approach, since perturbations are geometrically interpretable and configurations are easy to visualise.

\begin{figure}[t]
    \centering
    \begin{subfigure}[b]{0.48\linewidth}
        \centering
        \includegraphics[width=\linewidth]{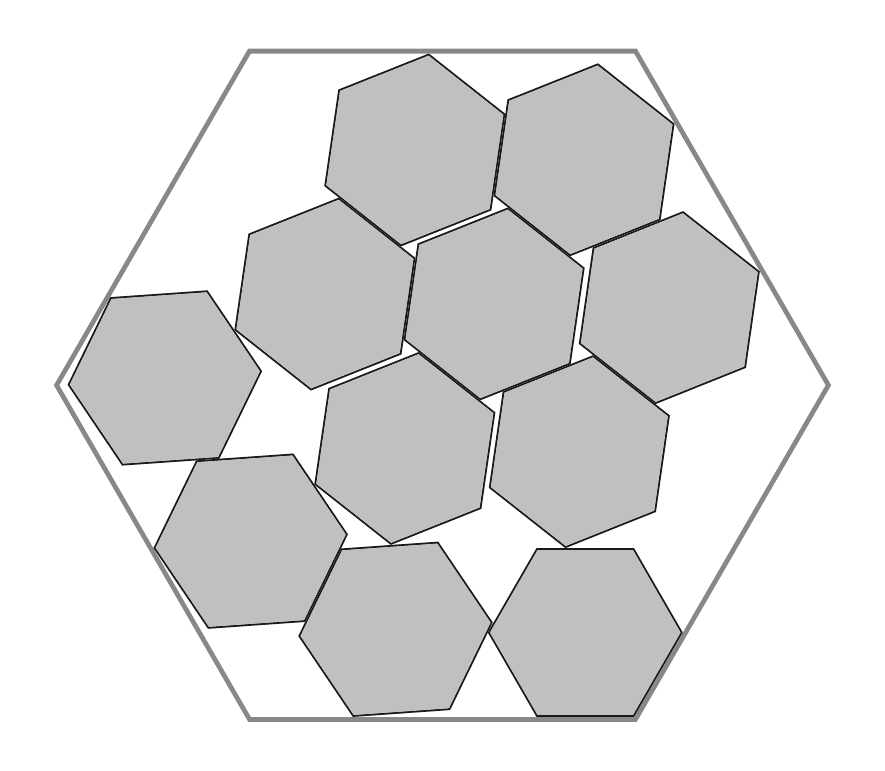}
        \caption{\alphaevolve\ ($L \!=\! 3.9301$)}
        \label{fig:alpha_11}
    \end{subfigure}
    \hfill
    \begin{subfigure}[b]{0.48\linewidth}
        \centering
        \includegraphics[width=\linewidth]{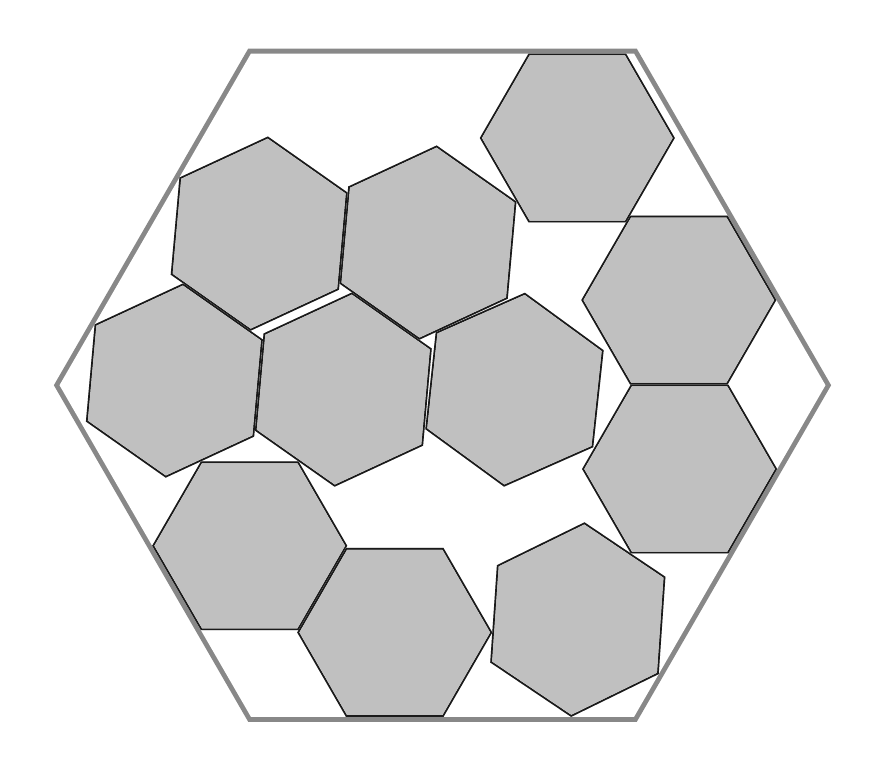}
        \caption{\improvevolve\ ($L \!=\! 3.9269$)}
        \label{fig:medium_sota_hexagons}
    \end{subfigure}
    
    \vspace{0.8em}
    \begin{subfigure}[b]{0.48\linewidth}
        \centering
        \includegraphics[width=\linewidth]{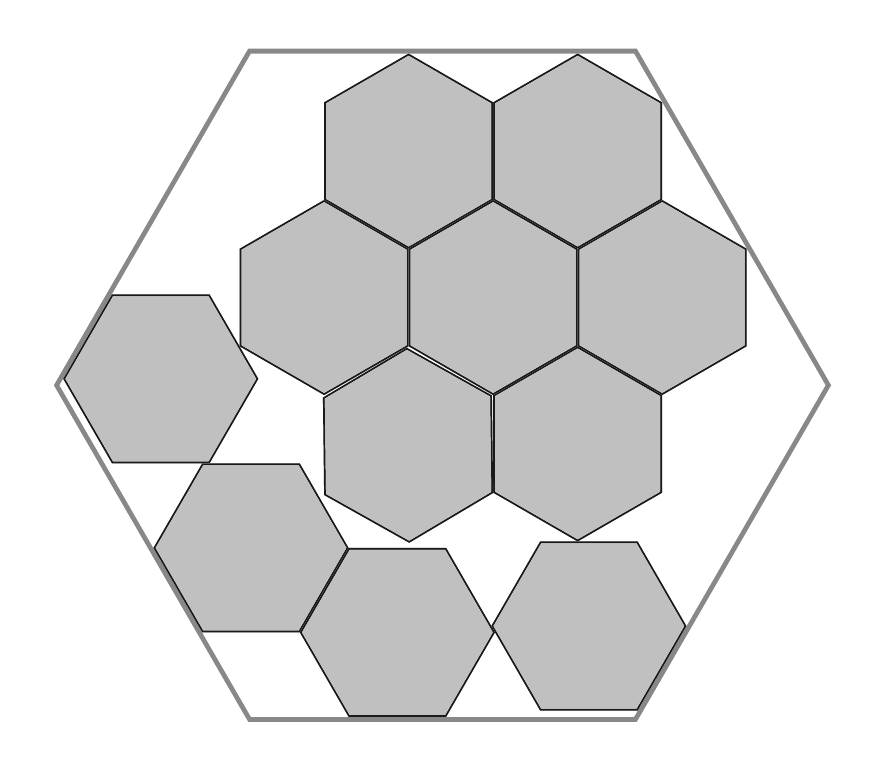}
        \caption{\textbf{\improvevolve\ SOTA} ($L \!=\! 3.9245$)}
        \label{fig:hex11_sota}
    \end{subfigure}
    \caption{Three packings of $n = 11$ unit hexagons. (a)~The previous state-of-the-art by \alphaevolve~\citep{novikov2025alphaevolve}. (b)~A structurally distinct packing found by \improvevolve. (c)~The new overall best packing discovered by \improvevolve, which recovers a structure similar to \alphaevolve\ but finds a deeper optimum.}
    \label{fig:hex_packing11}
\end{figure}

The task description supports any $n$ from $1$ to $25$, but evolution was run for a single fixed value $n = 11$, using the \improvevolve\ hyperparameters in \Cref{tab:hyperparameters}. The run produced two qualitatively different state-of-the-art packings: the first (\Cref{fig:medium_sota_hexagons}) is structurally distinct from \alphaevolve's packing (\Cref{fig:alpha_11}) and gives a noticeably shorter enclosing side, while the second (\Cref{fig:hex11_sota}) closely resembles \alphaevolve's structural basin but reaches a deeper optimum. The existence of two such different high-quality packings indicates that the fitness landscape contains multiple deep basins, underscoring the importance of Stage~A diversity.

We then validated the evolved program on a GPU for all $11 \leq n \leq 24$ under increased computational budgets: $K = 100$ initial seeds, $R = 100$ basin-hopping rounds, and a $\sigma$ schedule of \texttt{np.geom\-space(\allowbreak 1000,\allowbreak{} 0.001,\allowbreak{} 25)}. This yielded new state-of-the-art packings for $n = 12, 15,$ and $16$ (extended results in \Cref{tab:hex_large_scale} and \Cref{fig:all_hex_results}, both in \Cref{sec:hexagons_expanded}). Crucially, the same program evolved for $n = 11$ generalises to unseen problem sizes without retraining or re-evolution, demonstrating the robustness of the \texttt{improve}/\texttt{perturb} interface.

\begin{table}[t]
    \centering
    \caption{Enclosing hexagon side lengths for HEX $13$--$30$. \improvevolvepluse\ denotes minor human edits (see \Cref{sec:hex_improver_code}). Bold digits improve on the global solver baseline. A dash (--) indicates no previously reported packing.}
    \label{tab:hex_large_scale}
    \footnotesize
    \setlength{\tabcolsep}{1.5pt}
    \resizebox{\columnwidth}{!}{%
    \begin{tabular}{@{}lrrrr@{}}
        \toprule
        \textbf{Task} & \begin{tabular}[t]{@{}c@{}}\textbf{Human}\\[-0.1em]{\scriptsize\cite{friedman2025packing}}\end{tabular} & \begin{tabular}[t]{@{}c@{}}\textbf{Solvers}\\[-0.1em]{\scriptsize\cite{berthold2026solvers}}\end{tabular} & \textbf{\improvevolve} & \textbf{\improvevolvepluse} \\
        \midrule
        HEX 13 ($\downarrow$) & 4.0000 & -- & 4.0000 & 4.0000\\
        HEX 14 ($\downarrow$) & 4.2724 & 4.2\textbf{690} & 4.2724 & 4.2\textbf{690}\\
        HEX 15 ($\downarrow$) & 4.4541 & 4.4477 & 4.4\textbf{473} & 4.4\textbf{473} \\
        HEX 16 ($\downarrow$) & 4.5363 & 4.5279 & 4.5\textbf{275} & 4.5\textbf{275} \\
        HEX 17 ($\downarrow$) & 4.6188 & 4.61\textbf{4+} & 4.6188 & 4.61\textbf{36} \\
        HEX 18 \& 19 ($\downarrow$) & 4.6188 & -- & 4.6188 & 4.6188 \\
        HEX 20 \& 21 ($\downarrow$) & 5.0000 & -- & 5.0000 & 5.0000 \\
        HEX 22 ($\downarrow$) & 5.2856 & -- & 5.2857 & 5.2856\\
        HEX 23 ($\downarrow$) & 5.4286 & 5.413+ & 5.4848 & 5.4\textbf{000}\\
        HEX 24 ($\downarrow$) & 5.4848 & -- & 5.4848 & 5.4848 \\
        HEX 25 ($\downarrow$) & -- & -- & 5.6510 & 5.6239 \\
        HEX 26 ($\downarrow$) & -- & -- & 5.7142 & 5.7097 \\
        HEX 27 ($\downarrow$) & -- & -- & 5.7142 & 5.7142 \\
        HEX 28 ($\downarrow$) & -- & -- & 5.9723 & 5.9089 \\
        HEX 29 ($\downarrow$) & -- & -- & 6.0000 & 6.0000 \\
        HEX 30 ($\downarrow$) & -- & -- & 6.0045 & 6.0000 \\
        \bottomrule
    \end{tabular}%
    }
\end{table}

We also consider \textbf{ImprovEvolve+E}, a variant in which a human expert applies minor, targeted edits to the evolved program. For the hexagon task, three changes were made (annotated diff in \Cref{sec:hex_improver_code}): (i)~replacing the \texttt{L-BFGS-B} optimizer with \texttt{SLSQP}, which solves a quadratic subproblem at each step and is better suited to the constrained geometry, though less numerically stable; (ii)~softening the initial penalty weights for a gentler warm-up; and (iii)~relaxing the variable bounds and increasing the iteration limit to allow finer convergence. These edits preserve the overall algorithmic structure produced by the LLM while leveraging domain knowledge to improve robustness. They yield additional state-of-the-art packings for $n = 14, 17,$ and $23$ (\Cref{tab:hex_large_scale}, \Cref{fig:all_hex_results}). We further evaluated the edited program for $25 \leq n \leq 30$, where no packings have previously been reported, and obtained structured, non-chaotic packings (\Cref{fig:all_hex_results}).

\begin{figure}[t]
    \centering
    \includegraphics[width=0.9\linewidth]{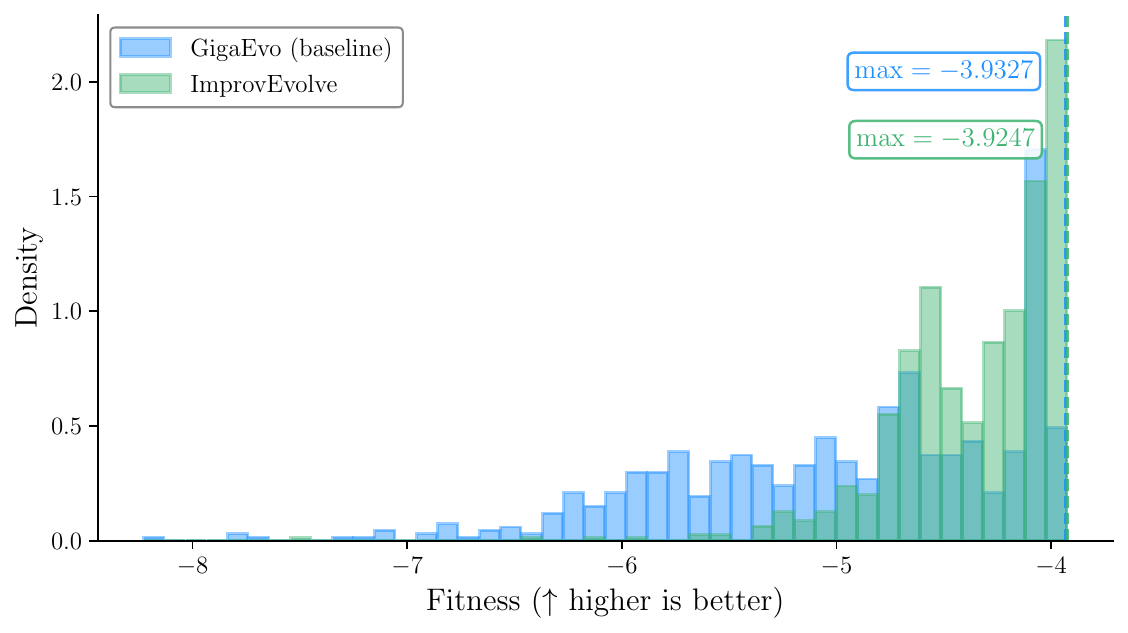}
    \caption{Distribution of validation fitness values ($-L$) across all valid programs produced during evolution for the $n = 11$ hexagon packing problem. \improvevolve\ consistently yields higher-fitness programs compared to the \gigaevo\ baseline under identical evolutionary parameters and time limits.}
    \label{fig:hex11_w_perturb_fitness}
\end{figure}

Under identical evolutionary parameters and time limits, \improvevolve\ proved substantially more effective than the \gigaevo\ baseline; \Cref{fig:hex11_w_perturb_fitness} shows the distribution of validation fitness values across the run. Moreover, validating the best \gigaevo\ program for $n > 11$ produced overlap errors, indicating that its solution did not transfer to other problem sizes. When the full complexity of an optimization procedure is encoded in a single program, the LLM struggles to produce a solution that generalises reliably.

Further details, including the prompt and the evolved program, are provided in \Cref{sec:hexagons_expanded}.

\subsection{Second autocorrelation inequality}
\begin{figure}[t]
    \centering
    \begin{subfigure}[b]{0.48\linewidth}
        \centering
        \includegraphics[width=\linewidth]{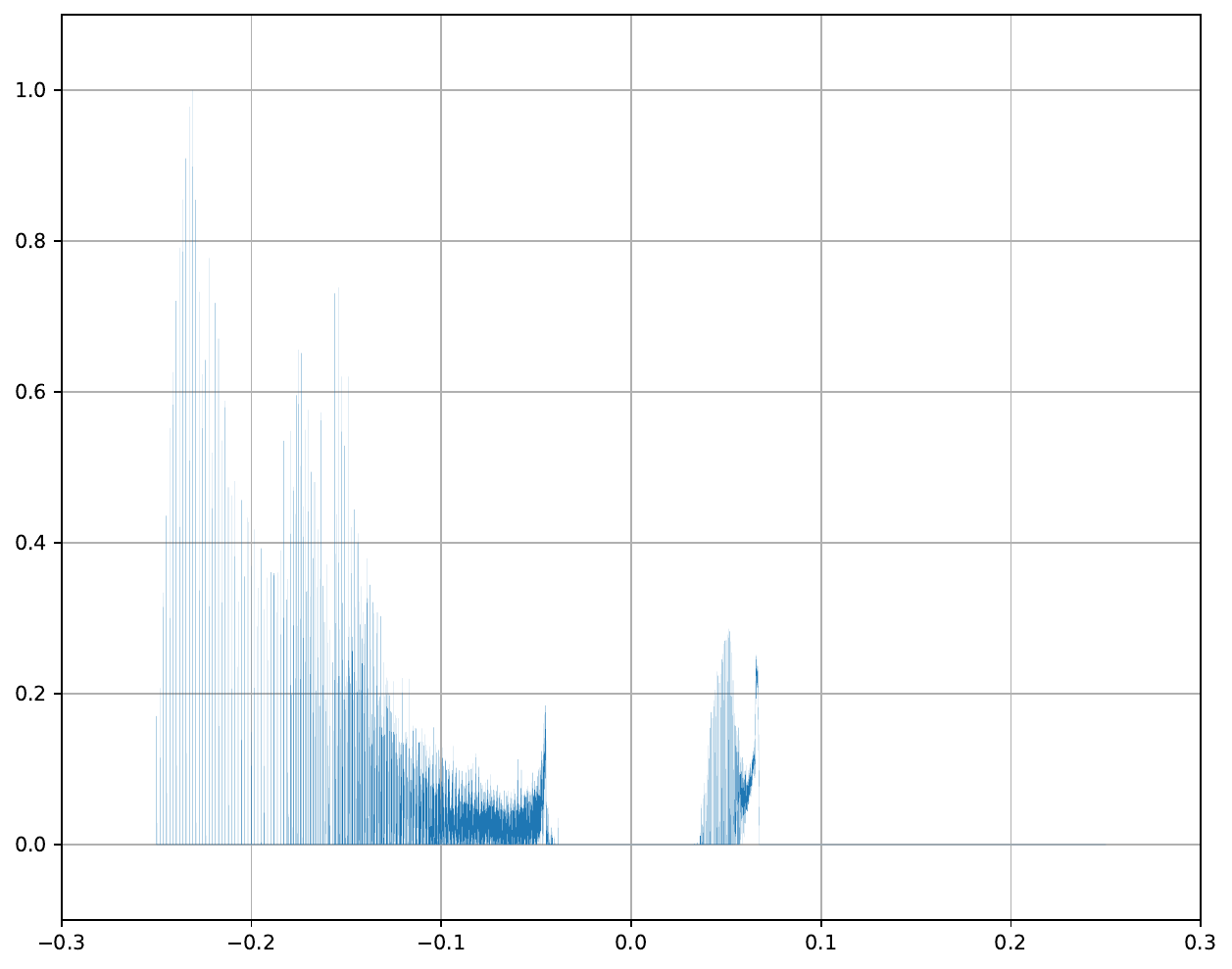}
        \caption{\alphaevolve}
        \label{fig:alpha_func}
    \end{subfigure}
    \hfill
    \begin{subfigure}[b]{0.48\linewidth}
        \centering
        \includegraphics[width=\linewidth]{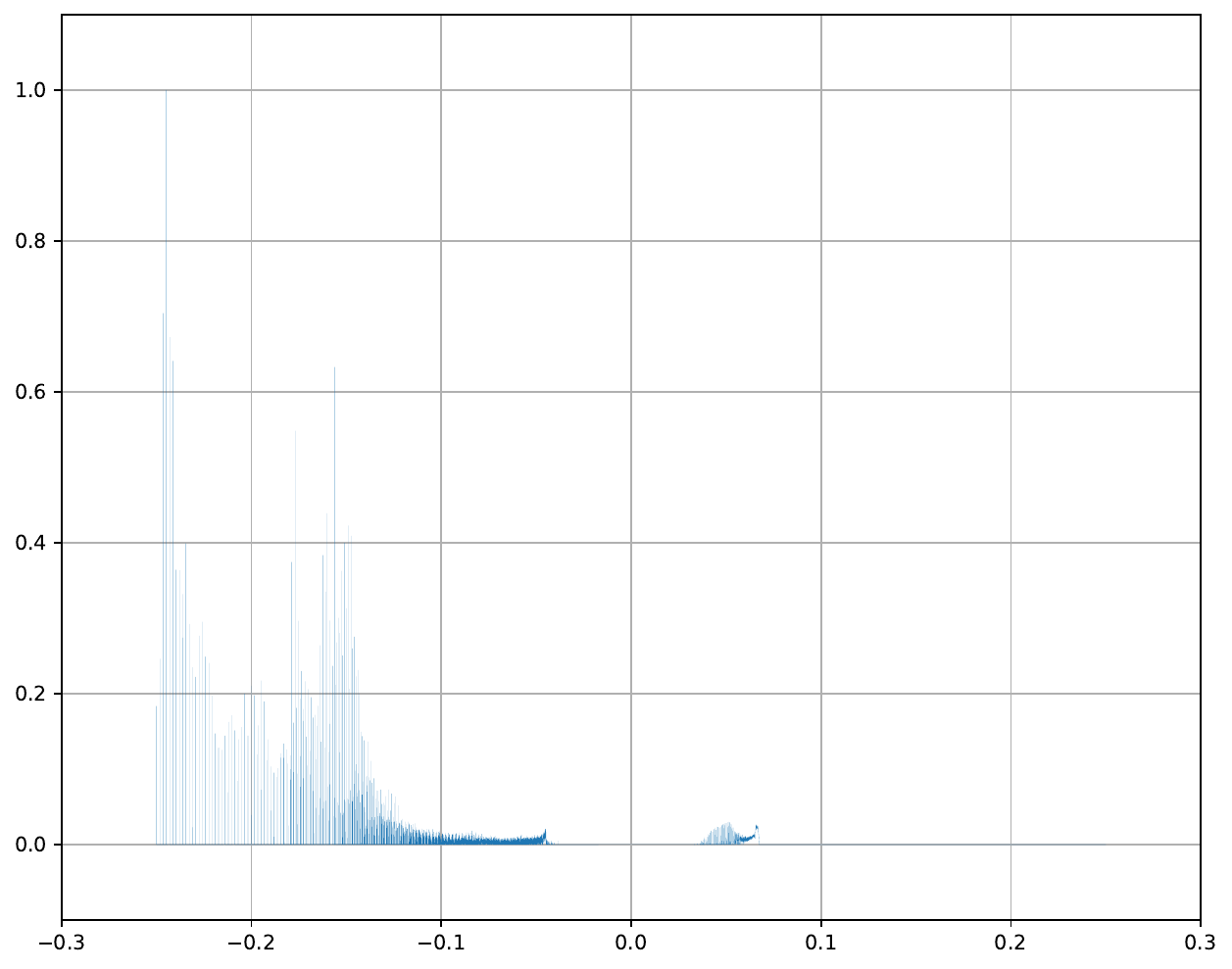}
        \caption{\improvevolvepluse}
        \label{fig:improv_func}
    \end{subfigure}
    \caption{Extremal functions $f$ for the second autocorrelation inequality. (a)~\alphaevolve's solution ($C(f) = 0.96102$, 50\,000 steps)~\citep{georgiev2025mathematical}. (b)~\improvevolvepluse\ solution ($C(f) = 0.96258$, 1.6M steps), obtained by resuming optimization from~(a). Despite the small difference in fitness, the two functions are qualitatively different in structure.}
    \label{fig:func_comparison}
\end{figure}

To test \improvevolve\ on problems with highly parameterised constructions, we consider the second autocorrelation inequality (ACI~2). For a non-negative function $f\colon \mathbb{R}\to\mathbb{R}$, define the autoconvolution
\[
(f * f)(t) = \int_{\mathbb{R}} f(t - x)\, f(x)\, dx,
\]
and let $C_2$ be the smallest constant such that
\[
\|f * f\|_2^2 \;\leq\; C_2\, \|f * f\|_1\, \|f * f\|_\infty
\]
holds for all non-negative~$f$. It is known that $0.88922 \leq C_2 \leq 1$~\citep{georgiev2025mathematical}; the goal is to tighten the lower bound by finding $f$ that maximises the ratio
\[
C(f) = \|f * f\|_2^2 \;\big/\; \bigl(\|f * f\|_1\, \|f * f\|_\infty\bigr),
\]
which serves as the fitness. The state-of-the-art solution found by \alphaevolve~\citep{georgiev2025mathematical} achieves $C(f) = 0.96102$ and is a highly irregular step function of 50\,000 steps (\Cref{fig:alpha_func}), making this benchmark substantially more complex than hexagon packing.

\begin{figure}[t]
    \centering
    \includegraphics[width=0.9\linewidth]{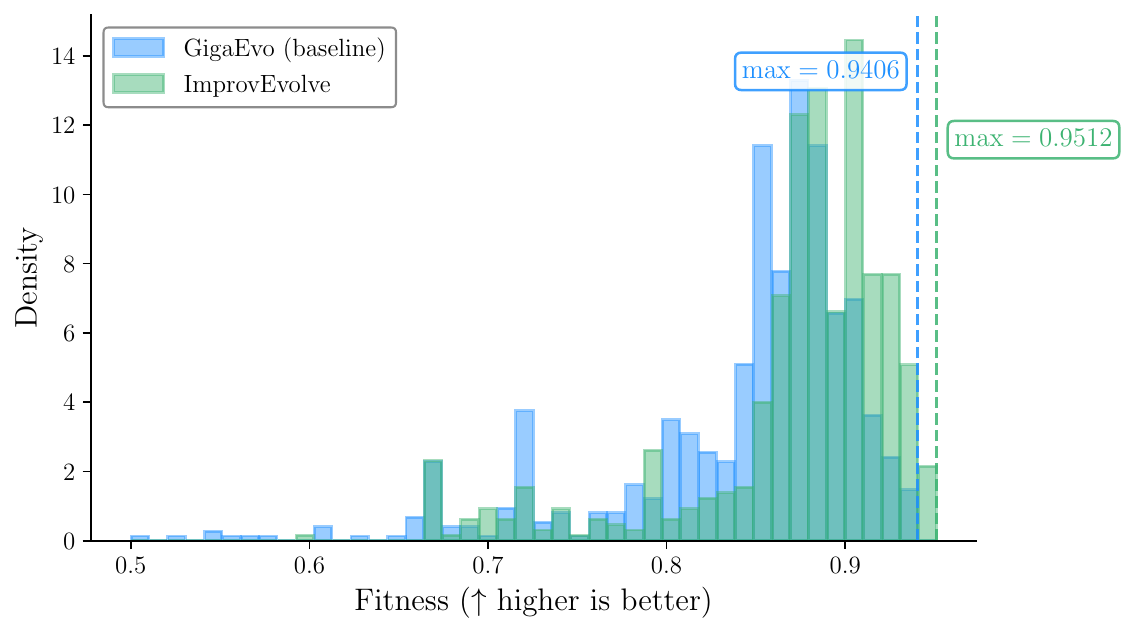}
    \caption{Distribution of validation fitness values $C(f)$ across all valid programs produced during evolution for the ACI~2 problem. \improvevolve\ yields consistently higher fitness than the \gigaevo\ baseline.}
    \label{fig:second_autocorr_w_perturb_fitness}
\end{figure}

The evolved \improvevolve\ program achieved $C(f) = 0.9512$, close to \alphaevolve's $0.96102$ and surpassing all other open-source frameworks~\citep{wang2025thetaevolve,assumpção2026codeevolveopensourceevolutionary} (\Cref{tab:new_sota_transposed}). The \gigaevo\ baseline reached only $C(f) = 0.9478$ under matched conditions (\Cref{fig:second_autocorr_w_perturb_fitness} shows the single-run fitness distribution; \Cref{tab:rebuttal_stats} reports multi-run robustness). \Cref{fig:cartoon_aci_stages} in \Cref{sec:autocorr_expanded} traces the full optimization path, showing how~$f$ and~$f\!*\!f$ evolve across the two-stage validation scheme and how a single perturbation at $\varepsilon = 10^{-3}$ triggers a large jump in~$C(f)$.

A key advantage of \improvevolve's modular design is the built-in flexibility to resume optimization from a given high-quality solution. We demonstrate this with \textbf{ImprovEvolve+E}: starting from \alphaevolve's solution (\Cref{fig:alpha_func}), we apply a human-edited version of the evolved program with three changes (annotated diff in \Cref{sec:autocorr_improver_code}): (i)~replacing the original multi-grid schedule (resolutions up to 32\,768) with a progressive schedule that begins at the input resolution and scales up to 1.6 million steps, so the optimizer can meaningfully refine \alphaevolve's 50\,000-step function; (ii)~removing the lower clipping at $10^{-10}$ that would otherwise erase small input-function steps; and (iii)~raising the L-BFGS iteration cap to allow finer convergence at the higher resolutions. In just fifteen iterations of the \texttt{improve}--\texttt{perturb} loop we obtain a new state-of-the-art lower bound of $C(f) = 0.96258$ (\Cref{fig:improv_func}). As with hexagon packing, this highlights the value of human--LLM collaboration: the LLM produces the core algorithmic structure through evolution, while a domain expert contributes targeted refinements that would be hard for the model to discover unaided---here, the insight that a much finer discretisation is needed to improve on a 50\,000-step solution.

\subsection{Robustness and Reproducibility}
To confirm that the observed gains reflect an algorithmic advantage rather than favorable variance under the stochastic LLM evolution, we ran three independent 100-generation runs of both \gigaevo\ and \improvevolve\ under identical conditions. \Cref{tab:rebuttal_stats} reports the mean, sample standard deviation, and 95\% confidence intervals, along with a Welch's $t$-test on the final fitness scores. \improvevolve\ achieves a statistically significant improvement on both HEX~11 ($p = 0.0086$) and ACI~2 ($p = 0.0493$), confirming that the modular decomposition delivers a reproducible advantage over learning a monolithic end-to-end optimizer.

\begin{table}[htbp]
    \centering
    \caption{Multi-run performance statistics across 3 independent runs per setting. Best single-run values are bolded. Welch's $t$-test confirms that \improvevolve\ yields a statistically significant mean improvement ($p < 0.05$) across runs.}
    \label{tab:rebuttal_stats}
    \resizebox{\columnwidth}{!}{%
    \begin{tabular}{@{}llcccc@{}}
        \toprule
        \textbf{Task} & \textbf{Method} & \textbf{Best} & \textbf{Mean $\pm$ Std} & \textbf{95\% CI} & \textbf{Welch's $t$ / $p$-value} \\
        \midrule
        \multirow{2}{*}{\textbf{HEX 11} ($\downarrow$)} 
        & \gigaevo\ (baseline) & 3.9296 & $3.9309 \pm 0.0016$ & $[3.9270, 3.9349]$ & \multirow{2}{*}{\makecell[c]{$t = -3.96$\\ $\mathbf{p = 0.0086}$}} \\
        & \textbf{\improvevolve} & \textbf{3.9245} & $\mathbf{3.9261 \pm 0.0014}$ & $\mathbf{[3.9227, 3.9295]}$ & \\[1.5ex]
        \midrule
        \multirow{2}{*}{\textbf{ACI 2} ($\uparrow$)} 
        & \gigaevo\ (baseline) & 0.9478 & $0.9430 \pm 0.0041$ & $[0.9328, 0.9532]$ & \multirow{2}{*}{\makecell[c]{$t = 2.43$\\ $\mathbf{p = 0.0493}$}} \\
        & \textbf{\improvevolve} & \textbf{0.9512} & $\mathbf{0.9494 \pm 0.0019}$ & $\mathbf{[0.9447, 0.9541]}$ & \\
        \bottomrule
    \end{tabular}%
    }
\end{table}

\subsection{Spherical codes}\label{sec:spherical_codes}
As a third benchmark, we apply \improvevolve\ to the \emph{spherical codes} (or \emph{Tammes}) problem~\citep{cohnSphericalCodes}, which asks for an arrangement of $N$ points on the unit sphere $S^{d-1}\!\subset\!\mathbb{R}^d$ that maximises the minimum pairwise distance---equivalently, minimises the maximum cosine
\[
\mu(\mathcal{X}) \;=\; \max_{i \neq j}\, \langle x_i, x_j\rangle,\qquad x_i \in S^{d-1},
\]
with fitness $-\mu(\mathcal{X})$. Each configuration $\mathcal{X}=\{x_1,\dots,x_N\}$ lives in an $Nd$-dimensional ambient space subject to $N$ unit-norm constraints, yielding a highly non-convex landscape with rich symmetry structure. Exact optima are known only for $N \leq 2d$; for larger~$N$, the online catalogue of~\citet{cohnSphericalCodes} maintains the best known configurations across a wide range of $(N, d)$ and serves as an extensible benchmark accumulated over decades of expert effort. The closely related \emph{kissing number} problem in $\mathbb{R}^d$ can be cast as deciding the feasibility of $\mu(\mathcal{X}) \leq 1/2$.

\begin{figure}[t]
\centering
\includegraphics[width=\linewidth]{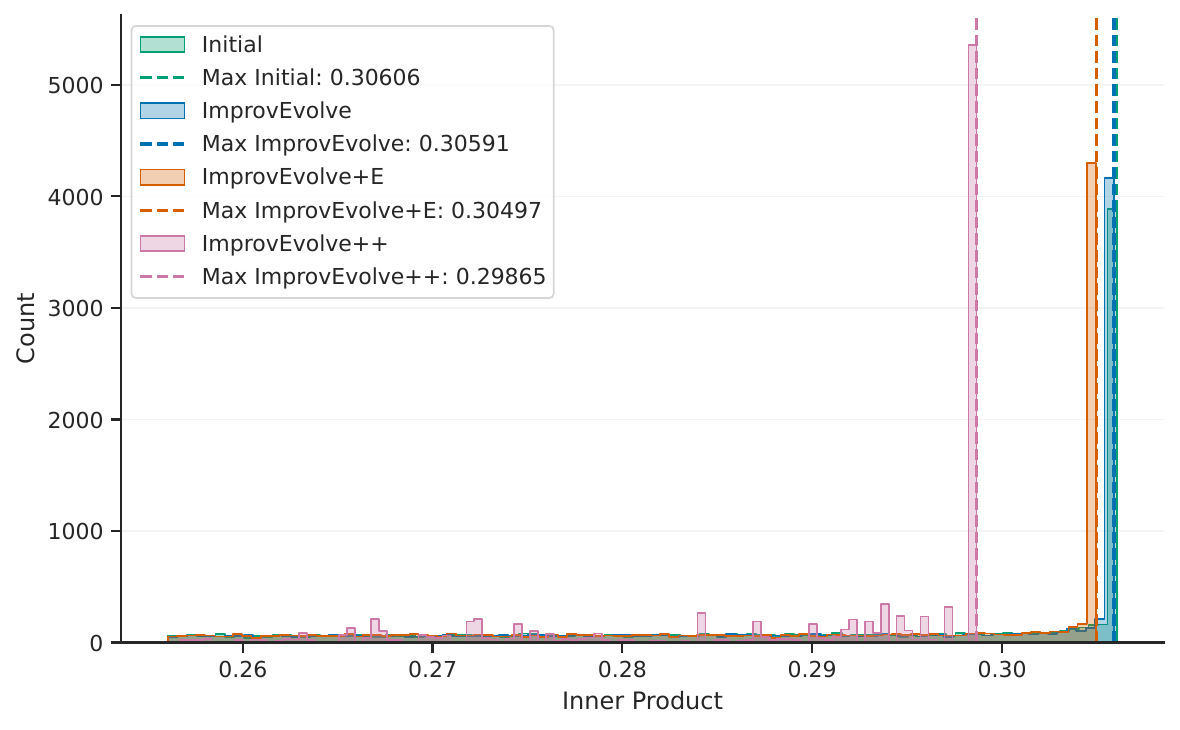}
\caption{Right tail of the inner-product distribution for $(N,d)=(296,16)$ when the \texttt{Improver}s are validated starting from the catalogue of~\citet{cohnSphericalCodes}. The maximum cosine $\mu(\mathcal{X})$ drops from $0.30606$ (Cohn) to $0.30591$ (\improvevolve), $0.30497$ (\improvevolvepluse), and $0.29865$ (\improvevolvextwo, our calibrated general improver); the density at the new, tighter maximum simultaneously grows. Full distribution and UMAP of the final configurations: \Cref{sec:spherical_validation,fig:spherical_umap}.}
\label{fig:spherical_main}
\end{figure}

\paragraph{Setup.} Evolution was run for the single configuration $(N,d)=(600,11)$ with the \gigaevo\ defaults of \Cref{sec:method} and the task prompt of \Cref{sec:spherical_task_prompt}; the program with the highest fitness $-\mu(\mathcal{X})$ becomes the \improvevolve\ \texttt{Improver}, and asking \emph{Gemini~3.1~Pro} to refine it under the minimal-edits prompt of \Cref{sec:llm_tuning_prompt} produces \improvevolvepluse\ (diff and sources in \Cref{sec:spherical_edits}). A separate \gigaevo\ run, evolved jointly across a panel of configurations spanning $d\in[8,16]$ and then tuned by the calibrated search protocol of \Cref{sec:spherical_calibrated_search}, yields the stronger \emph{general} improver \improvevolvextwo. Validation replaces the $K=10$ Stage-A seeds with the best-known catalogue configuration of~\citet{cohnSphericalCodes} and runs Stage-B basin-hopping with $\sigma$ on a geometric schedule under monotone acceptance, directly exercising the resume-from-input interface; per-program protocols and budgets are deferred to \Cref{sec:spherical_setup,sec:spherical_calibrated_search}. The inner-product diagnostics of \Cref{fig:spherical_main} use the high-headroom configuration $(296,16)$.

\paragraph{Results.} On the full benchmark of $90$ catalogue codes spanning $d\in[8,16]$ and $N\in[26,1021]$, the calibrated general improver \improvevolvextwo\ strictly lowers the maximum cosine $\mu(\mathcal{X})$ below the best-known value on $53$ of them, a mean relative improvement of $\mathbf{0.1765\%}$ when non-improvements are scored as zero (\Cref{tab:spherical_cohn})---$2.35\times$ the best prior program (\improvevolve\ $0.0752\%$, \improvevolvepluse\ $0.0618\%$), with the largest gains at $d=13$ and $d=16$. At the diagnostic configuration $(296,16)$, the max cosine drops from $0.30606$ (Cohn) to $0.29865$ (\improvevolvextwo), and the density at the new, tighter maximum grows---more pairs sit at the smaller separation, signalling a structured, discrete-spectrum code (full distribution in \Cref{sec:spherical_validation}). A UMAP of the final configurations (\Cref{fig:spherical_umap}) shows that \improvevolve\ and \improvevolvepluse\ barely move the points, whereas \improvevolvextwo\ relocates a substantial part of the configuration into structurally distinct regions, consistent with its larger gain. To put these figures in perspective, the maximum cosines of the best-known 11D configurations for $N=593$ (the state-of-the-art 11D kissing number \citep{novikov2025alphaevolve}) and $N=596$ differ by a mere $0.18\%$; even fractional-percentage improvements in the maximum cosine thus represent highly non-trivial structural rearrangements equivalent to accommodating multiple additional points.

\section{Ablation study \& Variants}

\subsection{Stage ablation}
\begin{table}[t]
    \centering
    \begin{threeparttable}
        \caption{Ablation study of the \improvevolve\ validation scheme on HEX~11. $L$: best side length; $K$: Stage~A seeds; $R$: Stage~B rounds. $\sigma$ decreases geometrically from $100$ to $0.001$.}
        \label{tab:stage_ablation}
        \footnotesize
        \setlength{\tabcolsep}{4pt}
        \begin{tabular}{@{}l*{5}{c}@{}}
            \toprule
             & \textbf{A only} & \textbf{B only} & \textbf{A{+}B} & \textbf{A{+}B{+}LLM}$^1$ & \textbf{Lat.{+}B}$^2$\\
            \midrule
            Best $L$      & 3.9282 & 3.9269 & 3.92\textbf{45} & 3.9270 & 3.9310 \\
            Seeds $K$     & 1000   & 1      & 100              & 100    & 100    \\
            Rounds $R$    & 0      & 100    & 5                & 100    & 100    \\
            \bottomrule
        \end{tabular}
        \begin{tablenotes}
            \footnotesize
            \item[1] \emph{Gemini~3~Pro} was prompted to edit the evolved program (see \Cref{sec:llm_tuning_prompt}).
            \item[2] Using \texttt{generate\allowbreak\_config} produced by \emph{Gemini~3~Pro}'s edited program.
        \end{tablenotes}
    \end{threeparttable}
\end{table}
Both Stage~A (multi-seed initialization) and Stage~B (basin-hopping) are essential to the validation scheme. \Cref{tab:stage_ablation} reports an ablation on HEX~11. Stage~A alone can produce a prototype of the optimal solution (see \Cref{fig:cartoon_stages} in \Cref{sec:hexagons_expanded}) but cannot refine it. Stage~B alone can polish a single initialization, but without a diverse set of starting points, the probability of reaching the global optimum stays low.

Although basin-hopping with acceptance temperature $T > 0$ and a simulated-annealing schedule is theoretically preferable to the monotonic case ($T=0$), its practical impact is limited: high temperatures prevent full refinement, while low temperatures barely improve the chance of escaping a local optimum. Worse, $T > 0$ risks abandoning a suboptimal configuration that lies on the path to the global optimum---for example, the $L = 4.6188$ arrangement for HEX~17 (\Cref{fig:cartoon_stages}). Recovering such a configuration can be prohibitively slow, particularly for problems like ACI~2 where each \texttt{improve} call is expensive.

\paragraph{Can an LLM replace human tuning?}
The \improvevolvepluse\ results relied on expert-guided hyperparameter edits, which raises the question of whether an LLM can match these autonomously. We prompted \emph{Gemini~3~Pro} to make minimal, non-structural modifications to the HEX~11 programs---adjusting constants, parameter values, or function arguments only (full prompt in \Cref{sec:llm_tuning_prompt}). For the \gigaevo\ baseline (edits marked \texttt{LLM-EDIT} in \Cref{sec:hex_baseline_code}), the LLM \emph{increased} Langevin noise to escape local minima and enlarged the batch size; the result was overlapping hexagons---the noise should have been \emph{decreased}. For the \improvevolve\ program (\Cref{sec:hex_improver_code}), edits included softening penalty weights and switching the lattice orientation from point- to flat-topped; while the latter helped dense packings at $n=13,21,31$, it suppressed the stochastic exploration Stage~A relies on, and both the fully edited program and a hybrid variant underperformed the unedited version (\Cref{tab:stage_ablation}).

These results expose a common failure mode: even advanced LLMs suggest plausible-sounding but empirically harmful edits. The damage is exacerbated in monolithic programs and mitigated by modular ones. Evolution can partially compensate by feeding validation results back into context, but models still lack a reliable sense for choosing optimization hyperparameters; agentic loops that evaluate candidate edits under different configurations are a more promising route.

\section{Discussion and Conclusion}\label{sec:discussion}
\improvevolve\ demonstrates that decomposing an optimization task into modular subproblems substantially improves the output of an evolutionary coding agent. Decomposition is a well-established principle, but its significance for LLM-powered discovery deserves emphasis. Current models produce a complete response in a single turn regardless of problem difficulty, while task decomposition, already a staple of AI-assisted code editors, can bring comparable benefit to AI for mathematics, with researchers specifying the modules the program must contain and agents validating and debugging them so the LLM-guided evolution can focus on the conceptual core.

Decomposition is especially important in mathematics, where a promising idea may give a consistently poor signal until its implementation is fully correct. For example, reproducing the 592-sphere 11D kissing number construction took roughly 20 chat exchanges with \emph{Gemini 3 Pro} provided with the \LaTeX source of the original article \citep{ganzhinov2022highlysymmetriclines}, and intermediate versions yielded fewer than 100 valid spheres. Such a complex construction is unlikely to arise through evolution alone, because the sparse reward signal would discard the idea before a correct implementation is obtained. Human experts can tailor the decomposition, for example, by delegating non-local search to basin-hopping. Combining these tailored decompositions with agentic workflows that autonomously decide which modules to refine and when to invoke external solvers represents a promising direction for scaling LLM-powered mathematical discovery.

\section*{Limitations}

Our main quantitative evaluation is limited to three mathematical optimization problems: hexagon packing, the second autocorrelation inequality, and spherical codes. While the results on the reported tasks are strong, broader generalisability to other problem classes remains to be demonstrated. Beyond the multi-run study in \Cref{tab:rebuttal_stats}, the headline state-of-the-art numbers are single-run best results and may vary across runs given the stochastic nature of evolution and basin-hopping.
\section*{Impact Statement}

This paper aims to advance machine learning by applying LLM-guided evolutionary methods to mathematical optimization. The primary ethical consideration is computational cost: each evolutionary run consumes significant energy and API resources. We mitigate this with a two-tier LLM choice strategy and a modular framework that reduces the iterations needed to reach competitive results. Broader societal impacts beyond standard scientific research are not expected to differ materially from other optimization and automated-discovery work.

\bibliography{evolve}
\bibliographystyle{icml2026}

\newpage
\appendix
\crefalias{section}{appendix}

\onecolumn
\section{Hexagon Packing Problem: Task Description and Evolved Solution}
\label{sec:hexagons_expanded}

\begin{figure*}[h] % changed [t] to [h] so it appears right at the start of the section
    \centering
    \includegraphics[width=0.95\linewidth]{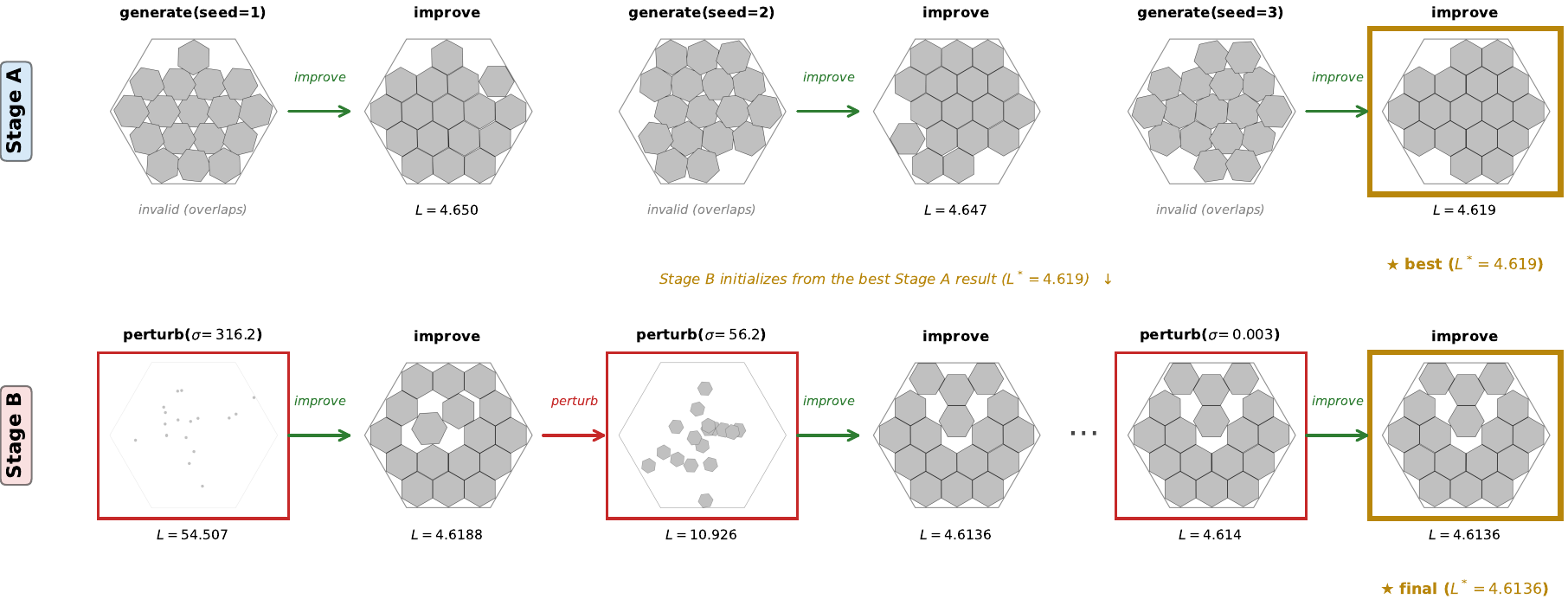}
    \caption{Illustration of the two-stage validation scheme (Algorithm~\ref{alg:validation}) on the HEX~17 problem. \emph{Stage~A (top):} \texttt{generate} proposes initial configurations that typically contain overlapping hexagons (invalid); \texttt{improve} resolves these overlaps and produces valid packings. The best result ($L^* = 4.619$, gold border) is selected. \emph{Stage~B (bottom):} basin-hopping alternates \texttt{perturb} (red border) and \texttt{improve}. Even an extreme perturbation ($\sigma = 316$, round~4) that scatters hexagons across a vast enclosing hexagon ($L = 54.5$) is recovered by \texttt{improve} back to $L = 4.619$. The breakthrough occurs at round~8 ($\sigma = 56.2$), where \texttt{improve} discovers a new, tighter basin at $L = 4.6136$---a structural improvement inaccessible from the Stage~A initialization. Subsequent rounds with small $\sigma$ confirm stability.}
    \label{fig:cartoon_stages}
\end{figure*}

\subsection{Prompts}
\begin{paracol}{2}
\subsubsection{Baseline}

\noindent \textbf{TASK DEFINITION -- HEXAGON PACKING PROBLEM}

\vspace{0.3cm}
\noindent \textbf{Challenge:} Implement a Python function that minimizes the side length of a flat-topped regular enclosing hexagon required to contain $N$ non-overlapping unit regular hexagons.

\vspace{0.3cm}
\noindent \textbf{Specific Benchmark:} $N=11$.\\
\textbf{General Requirement:} The code should be valid for any $N$ from 1 to 25, defaulting to 11.

\vspace{0.3cm}
\noindent \textbf{GEOMETRIC SPECIFICATIONS}
\begin{enumerate}
    \item \textbf{Unit Hexagons (Items):}
    \begin{itemize}
        \item Regular hexagons with circumradius $r = 1.0$.
        \item Side length $= 1.0$.
        \item May be arbitrarily placed $(x, y)$ and rotated $(\theta)$.
    \end{itemize}
    \item \textbf{Enclosing Hexagon (Container):}
    \begin{itemize}
        \item Regular, flat-topped, centered at $(0,0)$.
        \item ``Flat-topped'' vertices are located at angles $[0^\circ, \allowbreak 60^\circ, \allowbreak 120^\circ, \allowbreak 180^\circ, \allowbreak 240^\circ, \allowbreak 300^\circ]$ relative to the center.
        \item Objective: Minimize the enclosing hexagon's side length $L$ (which is equal to its circumradius).
    \end{itemize}
\end{enumerate}

\noindent \textbf{OBJECTIVE FUNCTION}\\
Maximize Fitness: $-L$ (Minimize side length $L$).\\
\textbf{Target for $N=11$:} $L \le 3.92$.

\vspace{0.3cm}
\noindent \textbf{CONSTRAINTS}
\begin{enumerate}
    \item \textbf{Containment:} All vertices of every unit hexagon must lie inside or on the boundary of the enclosing hexagon.
    \item \textbf{Non-overlap:} The intersection area between any pair of unit hexagons must be zero (touching is allowed).
    \item \textbf{Count:} Exactly \texttt{hex\_num} unit hexagons must be placed.
\end{enumerate}

\vspace{0.3cm}
\noindent \textbf{IMPLEMENTATION REQUIREMENTS}

\noindent \textbf{Libraries:}\\
Use \texttt{numpy}, \texttt{scipy}, \texttt{sklearn}, \texttt{jax}, \texttt{optax}, \texttt{jaxopt} or standard Python libraries.

\noindent \textbf{Function Interface:}\\
You must implement the following function structure exactly:

\begin{lstlisting}[language=Python, frame=single, basicstyle=\ttfamily\scriptsize]
import numpy as np

def entrypoint(hex_num=11, seed=0) -> tuple[np.ndarray, np.ndarray]:
    """
    Calculates the optimal packing configuration.
    
    Args:
        hex_num: Number of hexagons to pack.
        seed: Random seed for reproducibility.

    Returns:
        centers: np.ndarray of shape (hex_num, 2) representing (x, y) coordinates.
        angles: np.ndarray of shape (hex_num,) in radians [0, 2pi).
    """
    # Implement optimization logic here
    pass
\end{lstlisting}

\noindent \textbf{CRITICAL FAILURE MODES}
\begin{itemize}
    \item \textbf{Wrong Shapes:} Returning arrays of shape \texttt{(12, ...)} when \texttt{hex\_num=11}.
    \item \textbf{Geometry Errors:} Assuming point-topped enclosing hexagon instead of flat-topped.
    \item \textbf{Invalid Output:} Hexagons overlapping or protruding outside the boundary in the final result.
    \item \textbf{Wrong Return Type:} Returning a class or object instead of the \texttt{(centers, angles)} tuple.
\end{itemize}

\noindent \textbf{DOCUMENTATION}\\
Supply your code with a brief explanation of the algorithm using \textbf{LaTeX} in the introductory comment.

\switchcolumn
\subsubsection{With Improver}

\noindent \textbf{TASK DEFINITION -- HEXAGON PACKING OPTIMIZER}

\vspace{0.3cm}
\noindent \textbf{Challenge:} Implement a Python class that minimizes the side length of a flat-topped regular enclosing hexagon required to contain $N$ non-overlapping unit regular hexagons.

\vspace{0.3cm}
\noindent \textbf{Specific Benchmark:} $N=11$.\\
\textbf{General Requirement:} The code must be valid for any $N$ from 1 to 25.

\vspace{0.3cm}
\noindent \textbf{GEOMETRIC SPECIFICATIONS}
\begin{enumerate}
    \item \textbf{Unit Hexagons (Items):}
    \begin{itemize}
        \item Regular hexagons with circumradius $r = 1.0$.
        \item Side length $= 1.0$.
        \item May be arbitrarily placed $(x, y)$ and rotated $(\theta)$.
    \end{itemize}
    \item \textbf{Enclosing Hexagon (Container):}
    \begin{itemize}
        \item Regular, flat-topped, centered at $(0,0)$.
        \item ``Flat-topped'' vertices are located at angles $[0^\circ,\allowbreak 60^\circ,\allowbreak 120^\circ,\allowbreak 180^\circ,\allowbreak 240^\circ,\allowbreak 300^\circ]$ relative to the center.
        \item Objective: Minimize the enclosing hexagon's side length $L$ (which is equal to its circumradius).
    \end{itemize}
\end{enumerate}

\noindent \textbf{OBJECTIVE FUNCTION}\\
Maximize Fitness: $-L$ (Minimize side length $L$).\\
\textbf{Target for $N=11$:} $L \le 3.92$.

\vspace{0.3cm}
\noindent \textbf{CONSTRAINTS}
\begin{enumerate}
    \item \textbf{Containment:} All vertices of every unit hexagon must lie inside or on the boundary of the enclosing hexagon.
    \item \textbf{Non-overlap:} The intersection area between any pair of unit hexagons must be zero (touching is allowed).
    \item \textbf{Count:} Exactly \texttt{hex\_num} unit hexagons must be placed.
\end{enumerate}

\vspace{0.3cm}
\noindent \textbf{IMPLEMENTATION REQUIREMENTS}

\noindent \textbf{Libraries:}\\
Use \texttt{numpy}, \texttt{scipy}, \texttt{sklearn}, \texttt{jax}, \texttt{optax}, \texttt{jaxopt} or standard Python libraries.

\noindent \textbf{Class Interface:}\\
You must implement the following class structure exactly:

\begin{lstlisting}[language=Python, basicstyle=\ttfamily\scriptsize, frame=single]
import numpy as np

class Improver:
    def __init__(self, hex_num=11, seed: int = 0):
        self.hex_num = hex_num
        self.seed = seed

    def improve(self, input_config: tuple[np.ndarray, np.ndarray], seed=None) -> tuple[np.ndarray, np.ndarray]:
        """
        Refines the configuration to minimize L.
        """
        pass

    def perturb(self, input_config, intensity: float, seed=None) -> tuple[np.ndarray, np.ndarray]:
        """
        Perturbs the configuration (position/rotation).
        """
        pass

    def generate_config(self, seed=None) -> tuple[np.ndarray, np.ndarray]:
        """
        Generates a valid starting configuration.
        """
        pass

def entrypoint():
    return Improver
\end{lstlisting}

\noindent \textbf{CRITICAL FAILURE MODES}
\begin{itemize}
    \item \textbf{Wrong Shapes:} Returning arrays of shape \texttt{(12, ...)} when \texttt{hex\_num=11}.
    \item \textbf{Geometry Errors:} Assuming point-topped enclosing hexagon instead of flat-topped.
    \item \textbf{Invalid Output:} Hexagons overlapping or protruding outside the boundary in the final result.
\end{itemize}

\noindent \textbf{DOCUMENTATION}\\
Supply your code with a brief explanation of the algorithm using \textbf{LaTeX} in the introductory comment.
\end{paracol}

\onecolumn
%\newpage
\begin{figure*}[h]
    \centering
    \begin{subfigure}[b]{0.24\textwidth}
        \centering
        \includegraphics[width=\linewidth]{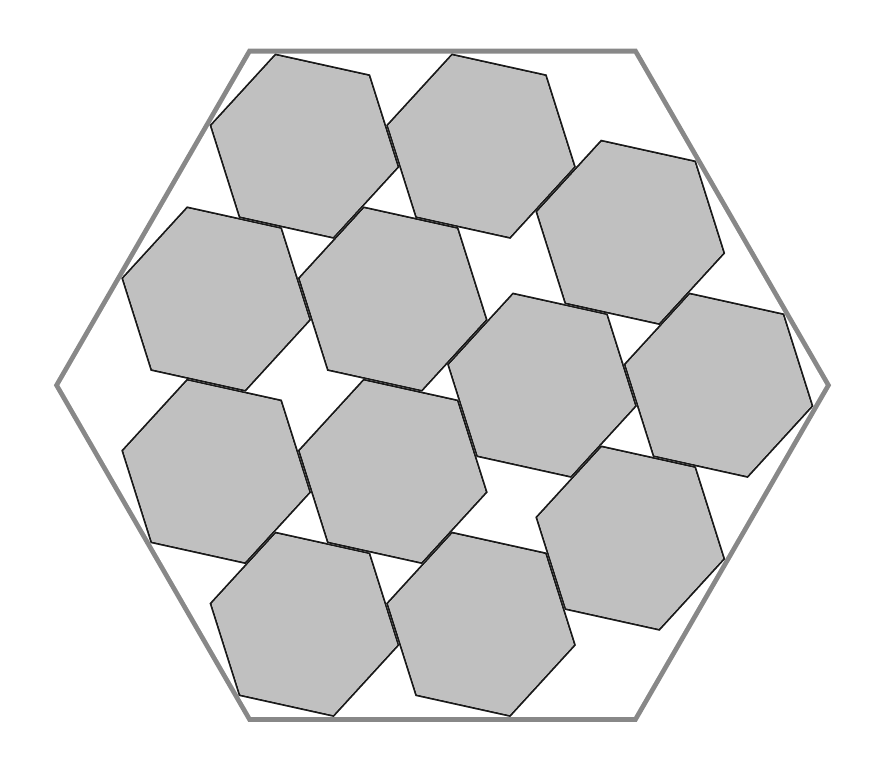}
        \caption{$n = 12$}
        \label{fig:hex12}
    \end{subfigure}\hfill
    \begin{subfigure}[b]{0.24\textwidth}
        \centering
        \includegraphics[width=\linewidth]{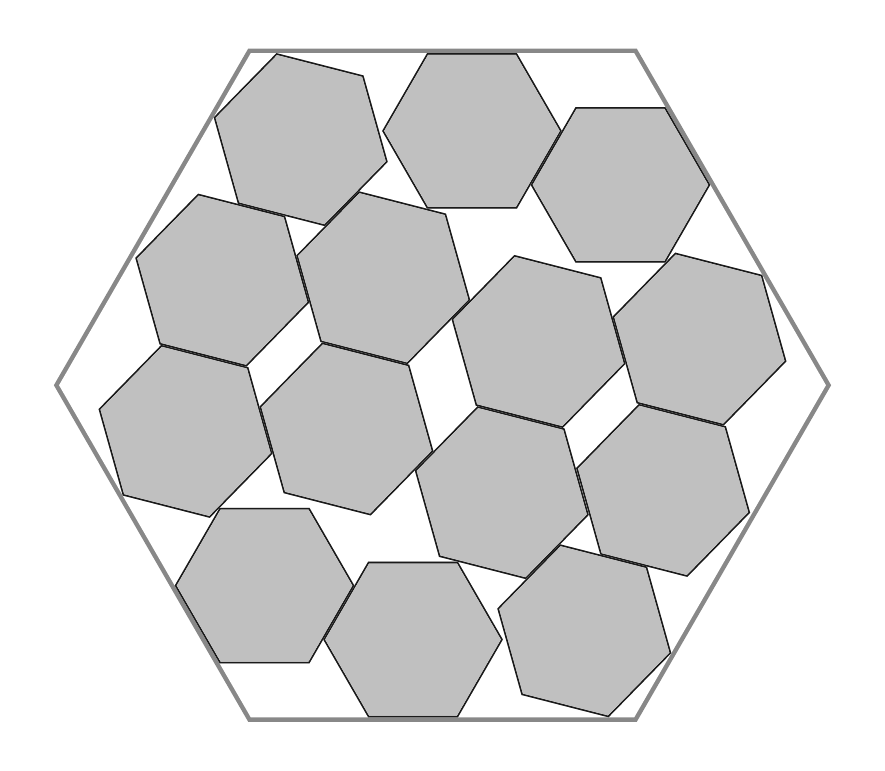}
        \caption{$n = 14$}
        \label{fig:hex14}
    \end{subfigure}\hfill
    \begin{subfigure}[b]{0.24\textwidth}
        \centering
        \includegraphics[width=\linewidth]{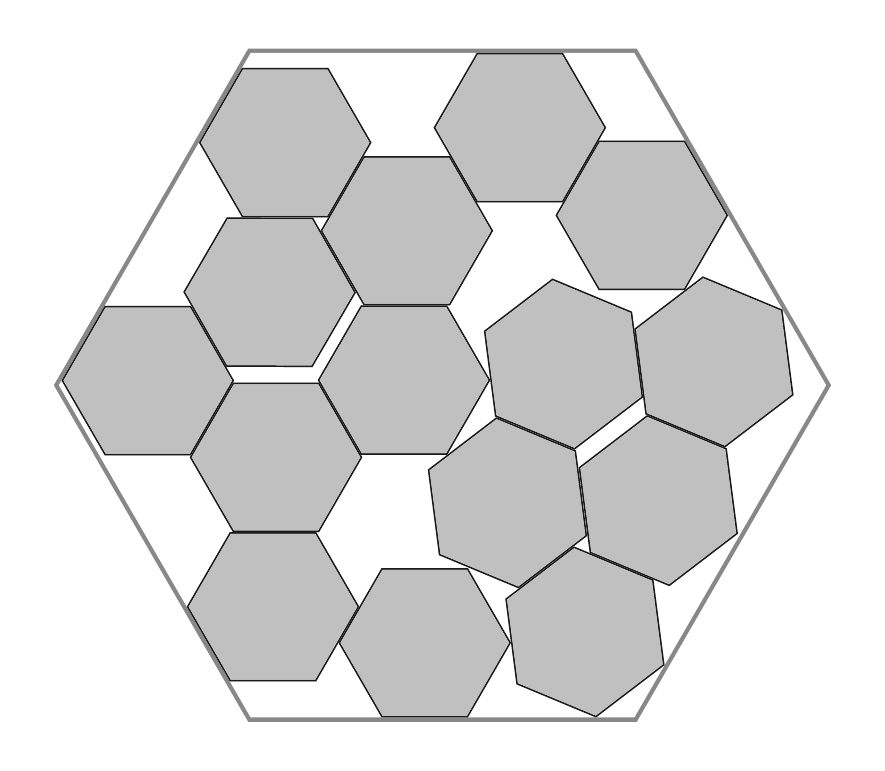}
        \caption{$n = 15$}
        \label{fig:hex15}
    \end{subfigure}\hfill
    \begin{subfigure}[b]{0.24\textwidth}
        \centering
        \includegraphics[width=\linewidth]{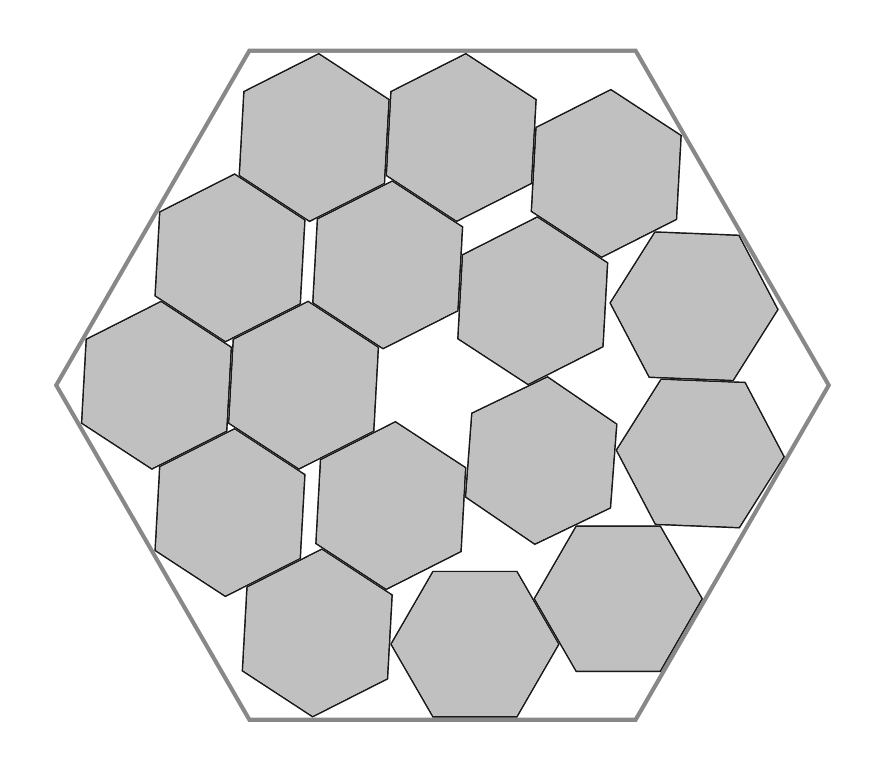}
        \caption{$n = 16$}
        \label{fig:hex16}
    \end{subfigure}

    \vspace{0.8em}

    \begin{subfigure}[b]{0.24\textwidth}
        \centering
        \includegraphics[width=\linewidth]{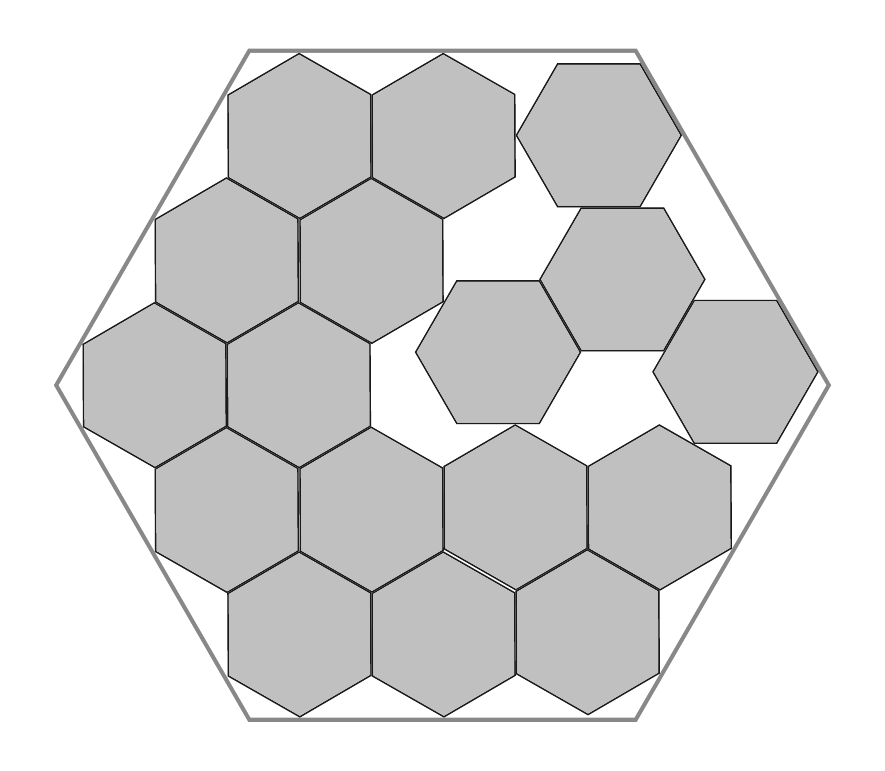}
        \caption{$n = 17$}
        \label{fig:hex17}
    \end{subfigure}\hfill
    \begin{subfigure}[b]{0.24\textwidth}
        \centering
        \includegraphics[width=\linewidth]{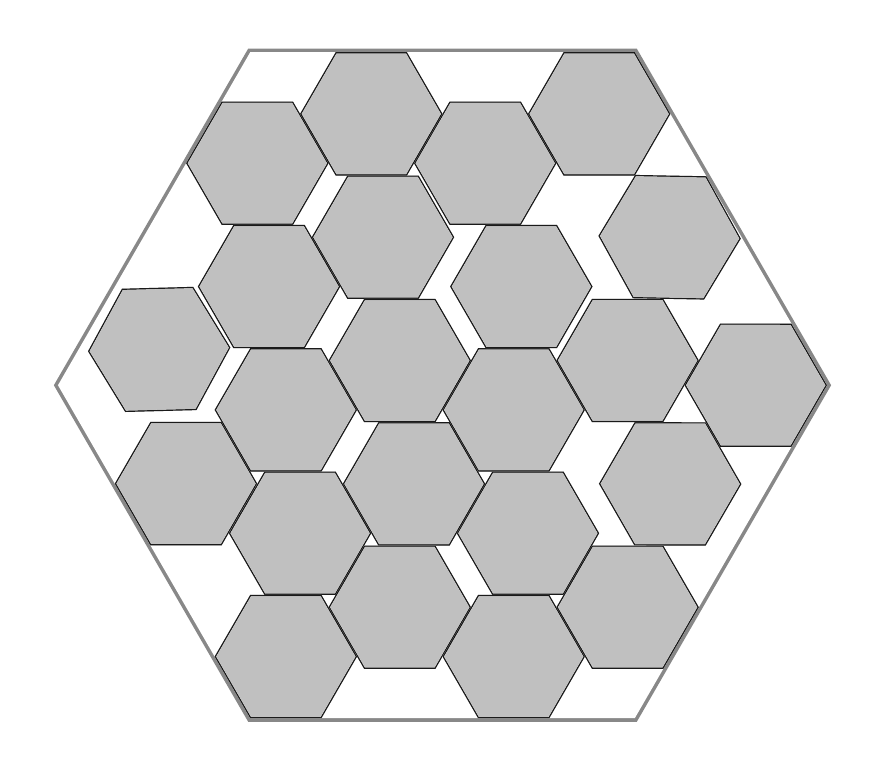}
        \caption{$n = 23$}
        \label{fig:hex23}
    \end{subfigure}\hfill
    \begin{subfigure}[b]{0.24\textwidth}
        \centering
        \includegraphics[width=\linewidth]{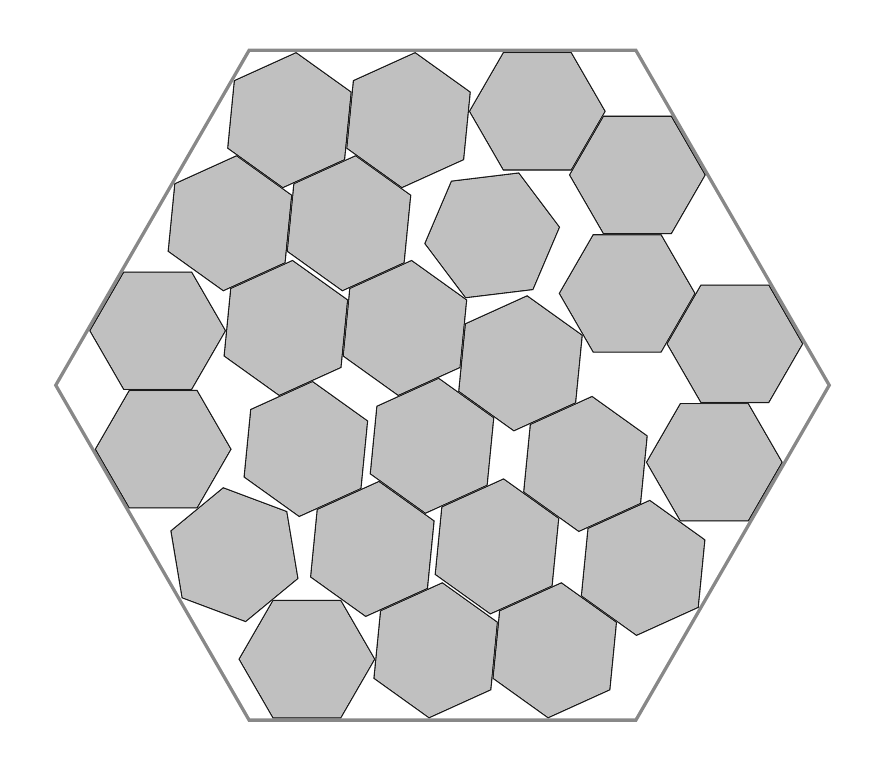}
        \caption{$n = 25$}
        \label{fig:hex25}
    \end{subfigure}\hfill
    \begin{subfigure}[b]{0.24\textwidth}
        \centering
        \includegraphics[width=\linewidth]{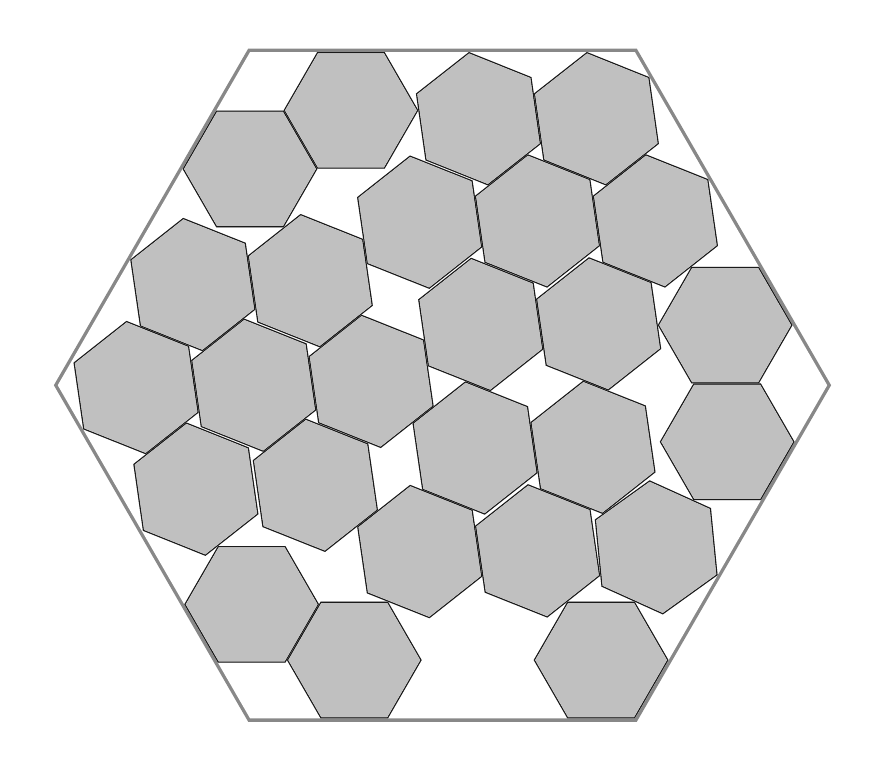}
        \caption{$n = 26$}
        \label{fig:hex26}
    \end{subfigure}

    \vspace{0.8em}

    \begin{subfigure}[b]{0.24\textwidth}
        \centering
        \includegraphics[width=\linewidth]{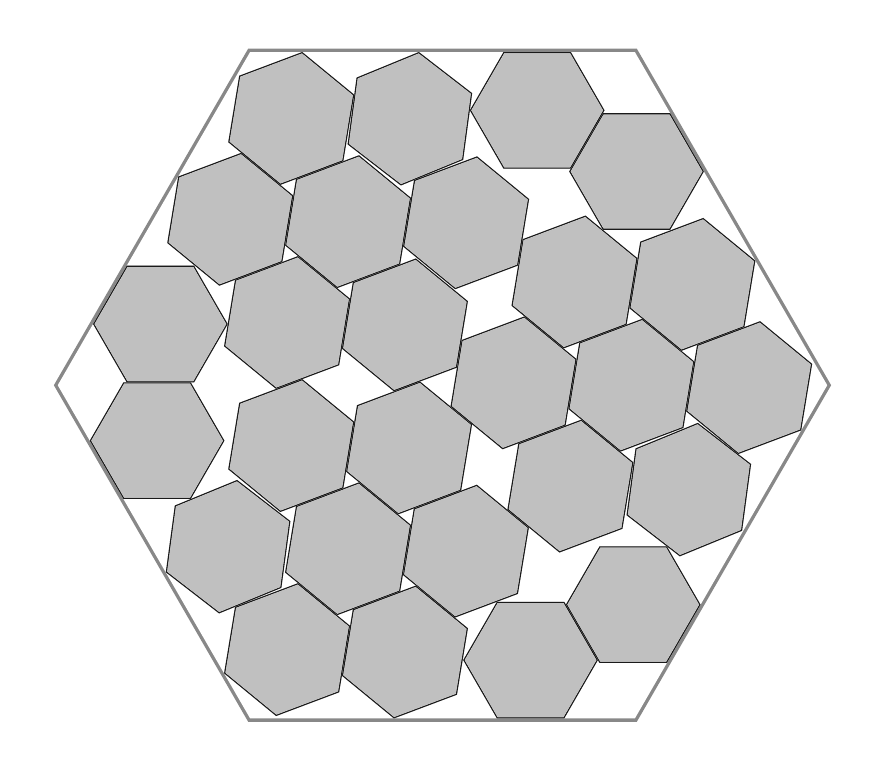}
        \caption{$n = 27$}
        \label{fig:hex27}
    \end{subfigure}\hfill
    \begin{subfigure}[b]{0.24\textwidth}
        \centering
        \includegraphics[width=\linewidth]{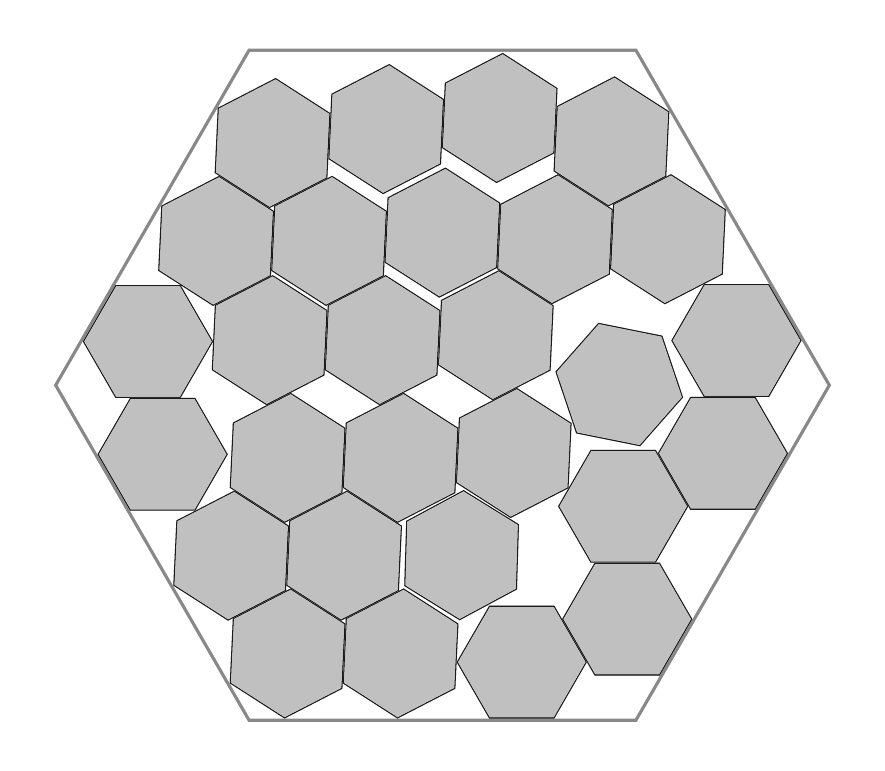}
        \caption{$n = 28$}
        \label{fig:hex28}
    \end{subfigure}\hfill
    \begin{subfigure}[b]{0.24\textwidth}
        \centering
        \includegraphics[width=\linewidth]{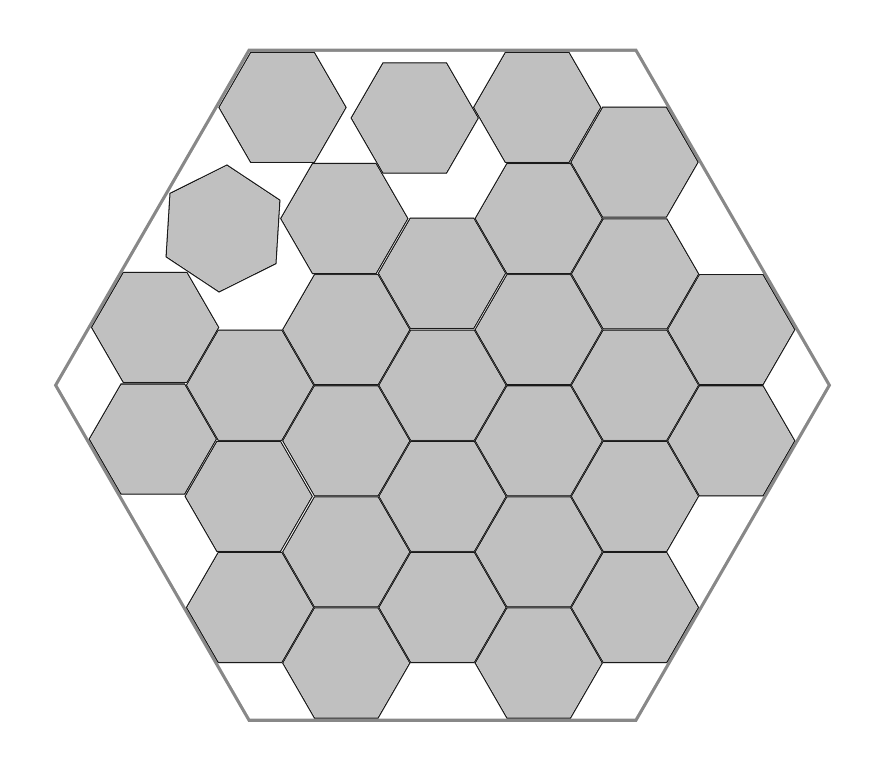}
        \caption{$n = 29-30$}
        \label{fig:hex29}
    \end{subfigure}\hfill
    \begin{subfigure}[b]{0.24\textwidth}
        % Empty 4th slot to balance the final row
    \end{subfigure}

    \caption{Extended state-of-the-art hexagon packings discovered by \improvevolve\ and \improvevolvepluse. New best-known configurations are shown for $n = 12$--$17$ and $n = 23$. For $n = 25$--$30$, no prior packings have been reported; the discovered configurations exhibit structured, non-chaotic arrangements. The $n = 30$ packing (shown) achieves the same side length $L = 6.0$ as $n = 29$: any single hexagon can be removed to obtain a valid $n = 29$ packing with identical~$L$. Side lengths are listed in \Cref{tab:new_sota_transposed} and \Cref{tab:hex_large_scale}.}
    \label{fig:all_hex_results}
\end{figure*}
\subsection{Evolved Solution Code}
\begin{paracol}{2}
\subsubsection{Baseline}\label{sec:hex_baseline_code}
The following program was evolved by \gigaevo\ (Program ID: 80ed10a3-7d87-42f7-a610-068c6d69411e, Fitness: -3.9327, Generation: 13).
\begin{minted}{python}
import numpy as np
import jax
import jax.numpy as jnp
import optax

def entrypoint(hex_num=11, seed=0) -> tuple[np.ndarray, np.ndarray]:
    """""
    Algorithm: Gradient-based packing optimization with SAT for non-overlap
    and explicit projection constraints for a flat-topped enclosing hexagon.
    L is the side length (and circumradius) of the container.
    """""
    jax.config.update("jax_enable_x64", True)

    # Flat-topped container normals at 30, 90, 150, 210, 270, 330 degrees
    cont_angles = jnp.deg2rad(jnp.array([30, 90, 150, 210, 270, 330]))
    container_normals = jnp.stack([jnp.cos(cont_angles), jnp.sin(cont_angles)], axis=1)

    # Unit hexagon local geometry
    unit_angles = jnp.arange(6) * (jnp.pi / 3.0)
    unit_verts_local = jnp.stack([jnp.cos(unit_angles), jnp.sin(unit_angles)], axis=1)
    sat_axes_local = jnp.array([0, jnp.pi/3, 2*jnp.pi/3]) + jnp.pi/6  # Midpoints of edges

    def get_world_data(centers, angles):
        c, s = jnp.cos(angles), jnp.sin(angles)
        rot = jnp.stack([jnp.stack([c, -s], axis=-1), jnp.stack([s, c], axis=-1)], axis=-2)
        verts = jnp.einsum('nij,kj->nki', rot, unit_verts_local) + centers[:, None, :]
        axes_angles = angles[:, None] + sat_axes_local[None, :]
        axes = jnp.stack([jnp.cos(axes_angles), jnp.sin(axes_angles)], axis=2)
        return verts, axes

    def containment_loss(verts, L):
        # For flat-topped hexagon with side L, apothem is L * cos(30 deg)
        h = L * jnp.cos(jnp.pi / 6.0)
        # Projections of unit hexagon vertices onto container normals must be <= h
        projections = jnp.dot(verts.reshape(-1, 2), container_normals.T)
        return jnp.sum(jax.nn.relu(projections - h)**2)

    def overlap_loss(verts, axes_all):
        N = verts.shape[0]
        idx_i, idx_j = jnp.triu_indices(N, 1)
        
        def pair_penalty(i, j):
            v_i, v_j = verts[i], verts[j]
            axes = jnp.concatenate([axes_all[i], axes_all[j]], axis=0)
            p_i = jnp.dot(v_i, axes.T)
            p_j = jnp.dot(v_j, axes.T)
            gaps = jnp.maximum(jnp.min(p_j, axis=0) - jnp.max(p_i, axis=0), 
                               jnp.min(p_i, axis=0) - jnp.max(p_j, axis=0))
            return jax.nn.relu(-jnp.max(gaps) + 1e-9)**2

        return jnp.sum(jax.vmap(pair_penalty)(idx_i, idx_j))

    def optimize_instance(sub_key):
        total_steps = 22000
        # LLM-EDIT: total_steps = 30000
        k1, k2, k3 = jax.random.split(sub_key, 3)
        
        # Insight: Expanded range based on expected L ~ 4.0
        centers = jax.random.uniform(k1, (hex_num, 2), minval=-3.5, maxval=3.5)
        # LLM_EDIT: centers = jax.random.uniform(k1, (hex_num, 2), minval=-3.0, maxval=3.0)
        init_angles = jax.random.uniform(k2, (hex_num,), minval=0.0, maxval=jnp.pi/3.0)
        params = {'centers': centers, 'angles': init_angles, 'log_L': jnp.log(4.5)}

        lr = optax.cosine_decay_schedule(0.04, total_steps, alpha=1e-3)
        optimizer = optax.chain(optax.clip_by_global_norm(1.0), optax.adam(learning_rate=lr))
        opt_state = optimizer.init(params)

        def step_fn(carry, i):
            p, state, rng = carry
            prog = i / total_steps
            # Smoother, more gradual weight ramping to allow rearrangement
            weight = 50.0 + 5e6 * jnp.power(prog, 2.8)
            
            def loss_fn(par):
                L = jnp.exp(par['log_L'])
                # Maintain symmetry constraint
                ang = par['angles'] % (jnp.pi / 3.0)
                v, a = get_world_data(par['centers'], ang)
                c_err = containment_loss(v, L)
                o_err = overlap_loss(v, a)
                return L + weight * (c_err + 5.0 * o_err), (L, c_err + o_err)

            (val, (L, err)), grads = jax.value_and_grad(loss_fn, has_aux=True)(p)
            
            rng, n_rng = jax.random.split(rng)
            # Langevin-style noise for escaping local minima
            temp = 1e-3 * jnp.exp(-4.0 * prog)
            # LLM-EDIT: temp = 2e-3 * jnp.exp(-3.5 * prog)
            grads['centers'] += jax.random.normal(n_rng, grads['centers'].shape) * temp
            
            updates, state = optimizer.update(grads, state)
            p = optax.apply_updates(p, updates)
            return (p, state, rng), (L, err)

        (final_p, _, _), (h_L, h_e) = jax.lax.scan(step_fn, (params, opt_state, k3), jnp.arange(total_steps))
        return final_p, h_L[-1], h_e[-1]

    batch_size = 96
    # LLM-EDIT: batch_size = 256
    rng_key = jax.random.PRNGKey(seed)
    keys = jax.random.split(rng_key, batch_size)
    all_p, all_L, all_e = jax.jit(jax.vmap(optimize_instance))(keys)
    
    # Precision check
    valid = all_e < 1e-7
    best_idx = jnp.where(valid, all_L, 1e10).argmin()
    best_idx = jax.lax.cond(valid[best_idx], lambda: best_idx, lambda: all_e.argmin())
    
    res = jax.tree_util.tree_map(lambda x: x[best_idx], all_p)
    return np.array(res['centers']), np.array(res['angles']) % (2*np.pi)
\end{minted}

\switchcolumn
\subsubsection{With Improver}\label{sec:hex_improver_code}
The following program was evolved by \improvevolve\ (Program ID: e98264be-75e4-4160-aaaa-530ec55f7848, Fitness: -3.92467, Generation: 13). Lines marked with \texttt{EDIT} show the human modifications applied in the \improvevolvepluse\ variant: the optimizer was changed from \texttt{L-BFGS-B} to \texttt{SLSQP}, initial penalty weights were softened, variable bounds were relaxed, and the iteration limit was increased.

\begin{minted}{python}
import numpy as np
import jax
import jax.numpy as jnp
from scipy.optimize import minimize

jax.config.update("jax_enable_x64", True)

@jax.jit
def get_unit_hex_verts(center, theta):
    # Regular hexagon vertices, circumradius 1.0
    base_angles = jnp.array([0.0, jnp.pi/3, 2*jnp.pi/3, jnp.pi, 4*jnp.pi/3, 5*jnp.pi/3], dtype=jnp.float64)
    angles = base_angles + theta
    return center + jnp.stack([jnp.cos(angles), jnp.sin(angles)], axis=1)

@jax.jit
def get_container_normals():
    # Normals for flat-topped enclosing hexagon (face-centered directions)
    angles = jnp.array([jnp.pi/6, jnp.pi/2, 5*jnp.pi/6, 7*jnp.pi/6, 3*jnp.pi/2, 11*jnp.pi/6], dtype=jnp.float64)
    return jnp.stack([jnp.cos(angles), jnp.sin(angles)], axis=1)

@jax.jit
def compute_separation(c1, theta1, c2, theta2):
    v1 = get_unit_hex_verts(c1, theta1)
    v2 = get_unit_hex_verts(c2, theta2)
    axes_angles = jnp.array([jnp.pi/6, jnp.pi/2, 5*jnp.pi/6], dtype=jnp.float64)
    n1 = jnp.stack([jnp.cos(axes_angles + theta1), jnp.sin(axes_angles + theta1)], axis=1)
    n2 = jnp.stack([jnp.cos(axes_angles + theta2), jnp.sin(axes_angles + theta2)], axis=1)
    axes = jnp.concatenate([n1, n2], axis=0)
    p1 = jnp.dot(v1, axes.T)
    p2 = jnp.dot(v2, axes.T)
    gaps = jnp.maximum(jnp.min(p1, axis=0) - jnp.max(p2, axis=0), jnp.min(p2, axis=0) - jnp.max(p1, axis=0))
    return jnp.max(gaps)

@jax.jit
def objective_fn(params, idx_i, idx_j, weight_cont, weight_ov, margin):
    N = (len(params) - 1) // 3
    centers = params[:2*N].reshape((N, 2))
    thetas = params[2*N:3*N]
    L = params[-1]
    
    # Flat-topped container containment
    H = L * (jnp.sqrt(3.0) / 2.0)
    normals = get_container_normals()
    
    def hex_cont(c, t):
        v = get_unit_hex_verts(c, t)
        return jnp.sum(jax.nn.relu(jnp.dot(v, normals.T) - H)**2)
    
    total_cont = jnp.sum(jax.vmap(hex_cont)(centers, thetas))

    def hex_ov(i, j):
        sep = compute_separation(centers[i], thetas[i], centers[j], thetas[j])
        return jax.nn.relu(margin - sep)**2

    total_ov = jnp.sum(jax.vmap(hex_ov)(idx_i, idx_j))
    
    return L + weight_cont * total_cont + weight_ov * total_ov

_grad_jit = jax.jit(jax.value_and_grad(objective_fn))

class Improver:
    def __init__(self, hex_num=11, seed: int = 0):
        self.hex_num = hex_num
        self.seed = seed
        self.idx_i, self.idx_j = map(jnp.array, np.triu_indices(hex_num, k=1))

    def generate_config(self, seed=None) -> tuple[np.ndarray, np.ndarray]:
        rng = np.random.default_rng(seed if seed is not None else self.seed)
        centers = []
        for q in range(-5, 6):
            for r in range(-5, 6):
                if abs(q+r) <= 5:
                    # Hexagonal lattice base
                    x = (q + r/2.0) * np.sqrt(3)
                    y = r * 1.5
                    # LLM-EDIT: x = q * 1.5
                    # LLM-EDIT: y = (r + q/2.0) * np.sqrt(3)
                    centers.append([x, y])
        centers = np.array(centers)
        dists = np.linalg.norm(centers, axis=1)
        # Start with slightly jittered central points
        idx = np.argsort(dists + rng.uniform(0, 0.5, size=len(dists)))[:self.hex_num]
        return centers[idx], rng.uniform(0, np.pi/3, self.hex_num)
        # LLM-EDIT: return centers[idx], rng.normal(0, 0.01, self.hex_num)

    def perturb(self, input_config, intensity, seed=None):
        rng = np.random.default_rng(seed if seed is not None else self.seed)
        c, a = map(np.copy, input_config)
        if intensity > 2.0:
            # Structural swap to escape local optima
            i, j = rng.choice(self.hex_num, 2, replace=False)
            c[i], c[j] = c[j].copy(), c[i].copy()
            a[i], a[j] = a[j].copy(), a[i].copy()
        
        c += rng.normal(0, 0.05 * intensity, c.shape)
        a += rng.normal(0, 0.02 * intensity, a.shape)
        return c, np.mod(a, np.pi/3)

    def improve(self, input_config, seed=None):
        centers, angles = input_config
        L_est = np.max(np.linalg.norm(centers, axis=1)) + 1.0
        x = np.concatenate([centers.flatten(), angles, [L_est]]).astype(np.float64)
        
        # Graduated hardening schedule for N=11 precision
        stages = [
            # EDIT: {'w_cont': 1e2, 'w_ov': 10, 'margin': 1e-3}
            {'w_cont': 1e3, 'w_ov': 1e2, 'margin': 1e-3},
            {'w_cont': 1e4, 'w_ov': 1e4, 'margin': 1e-4},
            {'w_cont': 1e6, 'w_ov': 1e6, 'margin': 1e-5},
            {'w_cont': 1e9, 'w_ov': 1e8, 'margin': 1e-7}
        ]
        # LLM-EDIT:
        # stages = [
        #     {'w_cont': 1e1, 'w_ov': 1e1, 'margin': 1e-1},
        #     {'w_cont': 1e3, 'w_ov': 1e3, 'margin': 1e-3},
        #     {'w_cont': 1e6, 'w_ov': 1e6, 'margin': 1e-5},
        #     {'w_cont': 1e9, 'w_ov': 1e9, 'margin': 1e-7}
        # ]

        for st in stages:
            res = minimize(
                lambda p: tuple(map(np.array, _grad_jit(p, self.idx_i, self.idx_j, st['w_cont'], st['w_ov'], st['margin']))),
                x, method='L-BFGS-B', jac=True,
                # EDIT: x, method='SLSQP', jac=True,
                bounds=[(-15, 15)]*(2*self.hex_num) + [(0.0, np.pi/3)]*self.hex_num + [(1.0, 15.0)],
                # EDIT: bounds=[(None, None)]*(2*self.hex_num) + [(None, None)]*self.hex_num + [(1.0, 15.0)],
                options={'maxiter': 800, 'ftol': 1e-10}
                # EDIT: options={'maxiter': 10000, 'ftol': 1e-11}
                # LLM-EDIT: options={'maxiter': 1000, 'ftol': 1e-10}

            )
            x = res.x

        final_centers = x[:2*self.hex_num].reshape((self.hex_num, 2))
        final_angles = np.mod(x[2*self.hex_num:3*self.hex_num], 2*np.pi)
        return final_centers, final_angles

def entrypoint():
    return Improver
\end{minted}
\end{paracol}

\section{Second Autocorrelation Inequality: Prompt and Evolved Solution}
\label{sec:autocorr_expanded}

\begin{figure}[h]
    \centering
    \begin{subfigure}[b]{0.48\linewidth}
        \centering
        \includegraphics[width=\linewidth]{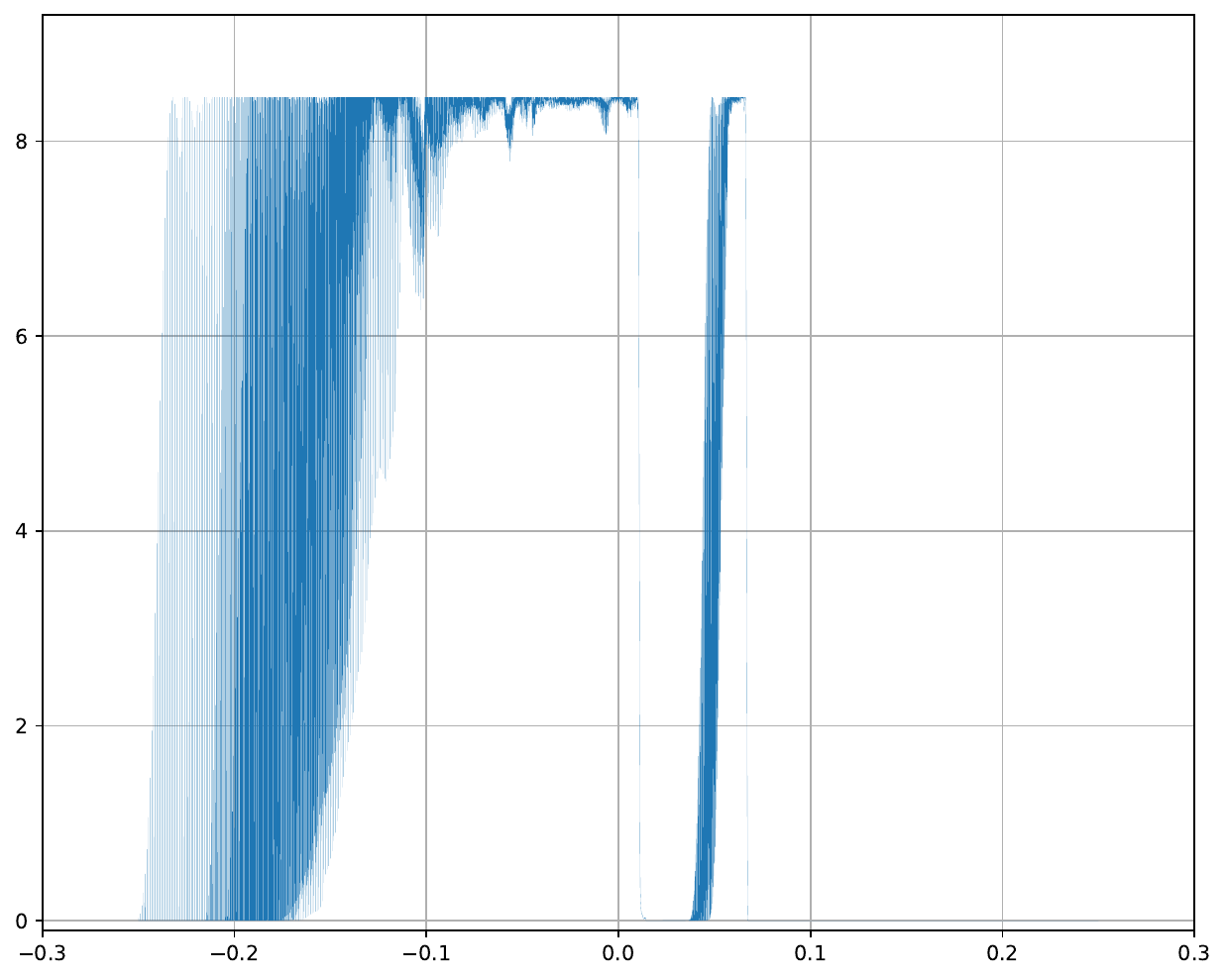}
        \caption{\alphaevolve}
        \label{fig:alpha_auto}
    \end{subfigure}
    \hfill
    \begin{subfigure}[b]{0.48\linewidth}
        \centering
        \includegraphics[width=\linewidth]{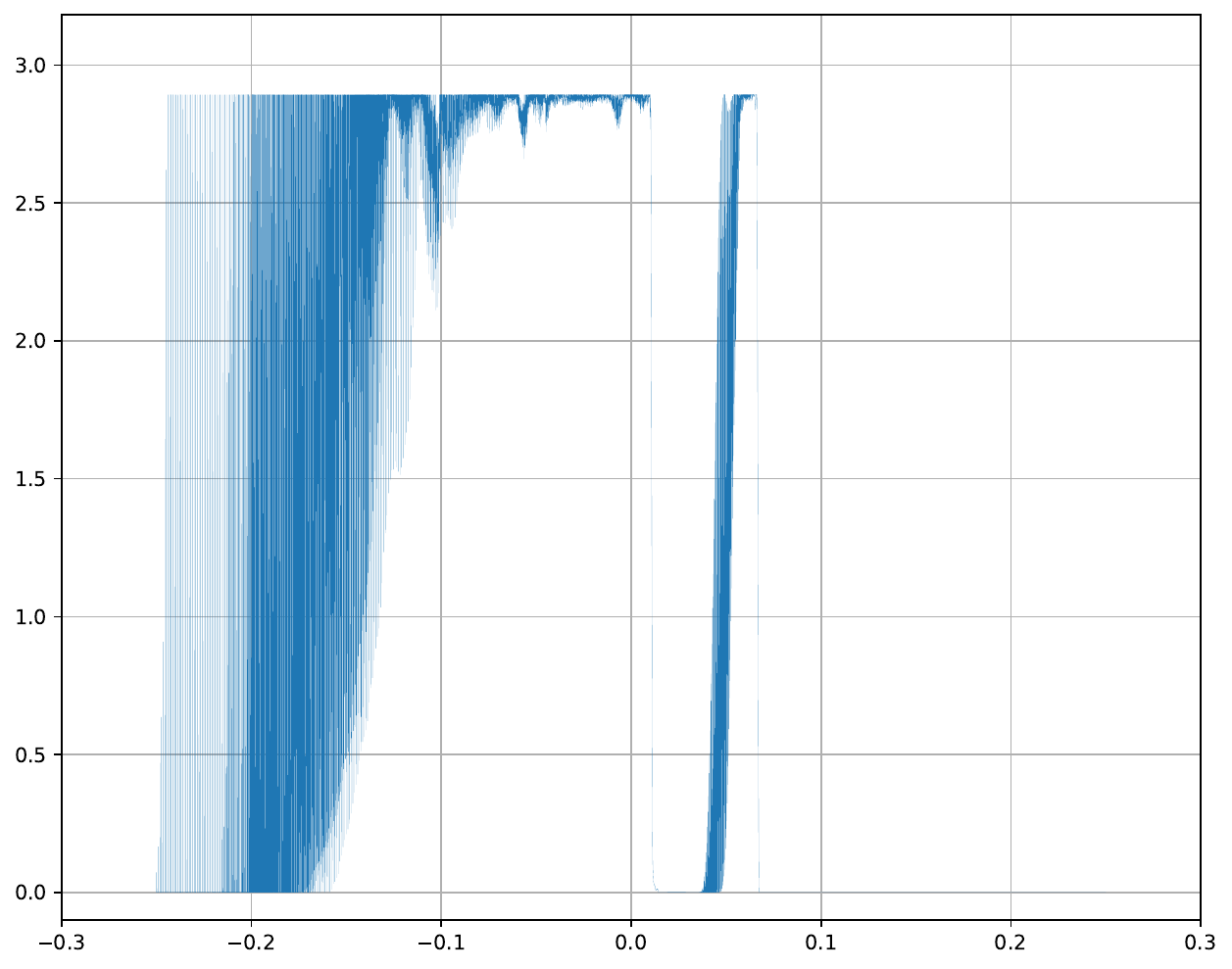}
        \caption{\improvevolvepluse}
        \label{fig:improv_auto}
    \end{subfigure}
    \caption{Autoconvolutions $f * f$ of the extremal functions from \Cref{fig:func_comparison}. The \improvevolvepluse\ solution achieves a higher ratio $C(f) = \|f*f\|_2^2 / (\|f*f\|_1 \|f*f\|_\infty)$.}
    \label{fig:autoconv_comparison}
\end{figure}

\begin{figure*}[h]
    \centering
    \includegraphics[width=0.95\linewidth]{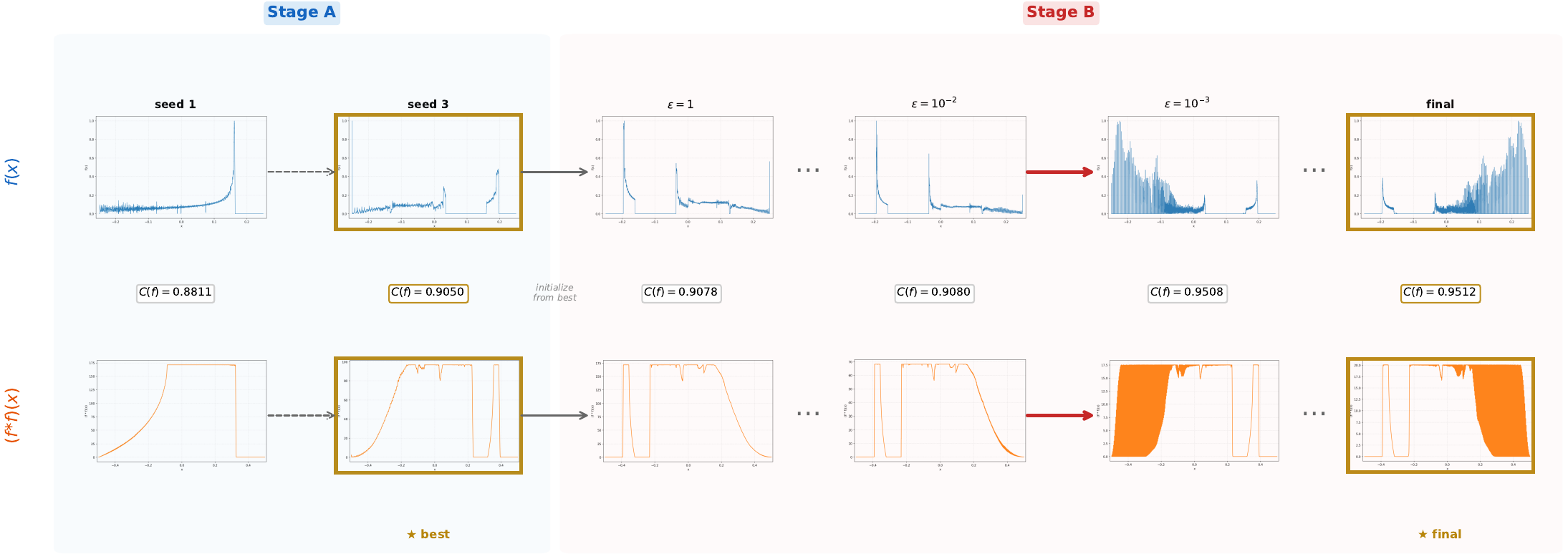}
    \caption{Optimization path of the \improvevolve\ validation scheme on the ACI~2 problem. Each column shows the evolved function $f(x)$ (top row) and its autoconvolution $(f\!*\!f)(x)$ (bottom row) at a key moment; $C(f)$ values are displayed between the rows. \emph{Stage~A} (blue background): two random seeds are generated and improved; seed~3 yields the best initial result ($C(f) = 0.9050$, gold border). \emph{Stage~B} (red background): the \texttt{perturb}--\texttt{improve} cycle is applied with geometrically decreasing perturbation intensity~$\varepsilon$. Iterations at $\varepsilon = 1$ and $\varepsilon = 10^{-2}$ produce only marginal gains ($C(f) \approx 0.908$). At $\varepsilon = 10^{-3}$ (red arrow), $C(f)$ jumps to $0.9508$, accompanied by a qualitative change in both~$f$ and~$f\!*\!f$. Further refinement yields the final result of $C(f) = 0.9512$ (gold border).}
    \label{fig:cartoon_aci_stages}
\end{figure*}

\onecolumn
\subsection{Prompt}
\begin{paracol}{2}
\subsubsection{Baseline}
\noindent \textbf{TASK DEFINITION -- SECOND AUTOCORRELATION INEQUALITY}

\vspace{0.3cm}
\noindent \textbf{Challenge:} Functional optimization in analysis. Implement a Python function that generates a non-negative function $f: \mathbb{R} \to [0, \infty)$ to maximize the constant $C$ in the second autocorrelation inequality.

\vspace{0.3cm}
\noindent \textbf{Adaptive Resolution:} The solver is \textbf{allowed and encouraged} to choose the optimal resolution (array size $N$) of the function.
\begin{itemize}
    \item \textit{Coarse Grids (Lower $N$):} faster iteration and global structure search.
    \item \textit{Fine Grids (Higher $N$):} higher precision and higher potential fitness $C$.
    \item Your solution should ideally be scalable for $N$ up to 100,000.
\end{itemize}

\vspace{0.3cm}
\noindent \textbf{MATHEMATICAL SPECIFICATIONS}
\begin{enumerate}
    \item \textbf{The Function ($f$):}
    \begin{itemize}
        \item A discrete 1D array representing a function $f(x)$ on a finite support.
        \item \textbf{Non-negativity:} $f(x) \ge 0$ for all $x$.
        \item \textbf{Non-triviality:} $\sum f(x) > 0$.
    \end{itemize}
    \item \textbf{The Convolution ($g$):}
    \begin{itemize}
        \item $g = f \star f$. In the discrete domain, this is the discrete convolution of the array $f$ with itself (linear convolution).
    \end{itemize}
    \item \textbf{The Objective:}
    \begin{itemize}
        \item Maximize the ratio involving the $L_1, L_2,$ and $L_\infty$ norms of the autocorrelation $g$.
    \end{itemize}
\end{enumerate}

\noindent \textbf{OBJECTIVE FUNCTION}\\
Maximize the constant $C$:
\[
C(f) = \frac{\|f \star f\|_2^2}{\|f \star f\|_1 \|f \star f\|_\infty}
\]
\textbf{Target:} $C \ge 0.962$.

\vspace{0.3cm}
\noindent \textbf{CONSTRAINTS}
\begin{enumerate}
    \item \textbf{Validity:} All elements of the output array must be non-negative finite floats.
    \item \textbf{Bounds:} $0.961 \le C \le 1$.
    \item \textbf{Minimum Resolution:} The final output array must have length $N \ge 1024$ to ensure the discretization approximates the continuous function reasonably well.
\end{enumerate}

\vspace{0.3cm}
\noindent \textbf{IMPLEMENTATION REQUIREMENTS}

\noindent \textbf{Libraries:}\\
Use \texttt{numpy}, \texttt{jax}, \texttt{optax}, \texttt{scipy}, \texttt{jaxopt}, or standard Python libraries. \textbf{JAX is preferred} for automatic differentiation and GPU acceleration.

\noindent \textbf{Interface:}\\
You must implement the following function structure exactly:

\begin{lstlisting}[language=Python, frame=single, basicstyle=\ttfamily\scriptsize]
import numpy as np

def entrypoint() -> np.ndarray:
    """
    Generates and optimizes a function f to maximize the autocorrelation constant C.

    The function should perform the optimization internally (using JAX/SciPy/etc.)
    and return the final optimized 1D array.

    Returns:
        f: 1D NumPy array (float) representing the optimized function. 
           Shape should be at least (1024,).
    """
    pass
\end{lstlisting}

\noindent \textbf{CRITICAL FAILURE MODES}
\begin{itemize}
    \item \textbf{Zero Function:} Returning an array of all zeros.
    \item \textbf{Negative Values:} Returning $f(x) < 0$.
    \item \textbf{Dimensionality Errors:} Returning a 2D array.
    \item \textbf{Excessively Low Resolution:} Returning $N < 1000$ where convolution properties degrade significantly.
    \item \textbf{Delusions:} Trying to ``guess'' optimal function rather than searching for it. Believe me, the example for $C = 0.961$ is a really strange function.
\end{itemize}

\noindent \textbf{DOCUMENTATION}\\
Supply your code with a brief explanation of the algorithm, using \textbf{LaTeX} in the introductory comment.

\switchcolumn
\subsubsection{For Improver}

\noindent \textbf{TASK DEFINITION -- SECOND AUTOCORRELATION INEQUALITY}

\vspace{0.3cm}
\noindent \textbf{Challenge:} Functional optimization in analysis. Implement a Python class that generates a non-negative function $f: \mathbb{R} \to [0, \infty)$ to maximize the constant $C$ in the second autocorrelation inequality.

\vspace{0.3cm}
\noindent \textbf{Adaptive Resolution:} The solver is \textbf{allowed and encouraged} to change the resolution (array size $N$) of the function.
\begin{itemize}
    \item \textit{Coarse Grids (Lower $N$):} faster iteration and global structure search.
    \item \textit{Fine Grids (Higher $N$):} higher precision and higher potential fitness $C$.
    \item Your solution must be scalable for $N$ up to $100\,000$ (but, of course, with larger time limits).
\end{itemize}

\vspace{0.3cm}
\noindent \textbf{MATHEMATICAL SPECIFICATIONS}
\begin{enumerate}
    \item \textbf{The Function ($f$):}
    \begin{itemize}
        \item A discrete 1D array representing a function $f(x)$ on a finite support.
        \item \textbf{Non-negativity:} $f(x) \ge 0$ for all $x$.
        \item \textbf{Non-triviality:} $\sum f(x) > 0$.
    \end{itemize}
    \item \textbf{The Convolution ($g$):}
    \begin{itemize}
        \item $g = f \star f$. In the discrete domain, this is the discrete convolution of the array $f$ with itself (linear convolution).
    \end{itemize}
    \item \textbf{The Objective:}
    \begin{itemize}
        \item Maximize the ratio involving the $L_1, L_2,$ and $L_\infty$ norms of the autocorrelation $g$.
    \end{itemize}
\end{enumerate}

\noindent \textbf{OBJECTIVE FUNCTION}\\
Maximize the constant $C$:
\[
C(f) = \frac{\|f \star f\|_2^2}{\|f \star f\|_1 \|f \star f\|_\infty}
\]
\textbf{Target:} $C \ge 0.962$.

\vspace{0.3cm}
\noindent \textbf{CONSTRAINTS}
\begin{enumerate}
    \item \textbf{Validity:} All elements of the output array must be non-negative finite floats.
    \item \textbf{Bounds:} $0.961 \le C \le 1$.
    \item \textbf{Minimum Resolution:} The final output array must have length $N \ge 1024$ to ensure the discretization approximates the continuous function reasonably well.
\end{enumerate}

\vspace{0.3cm}
\noindent \textbf{IMPLEMENTATION REQUIREMENTS}

\noindent \textbf{Libraries:}\\
Use \texttt{numpy}, \texttt{jax}, \texttt{optax}, \texttt{scipy}, \texttt{jaxopt}, or standard Python libraries. \textbf{JAX is preferred} for automatic differentiation and GPU acceleration.

\noindent \textbf{Class Interface:}\\
You must implement the following class structure exactly:

\begin{lstlisting}[language=Python, frame=single, basicstyle=\ttfamily\scriptsize]
import numpy as np

class Improver:
    def __init__(self, seed: int = 0):
        """
        Initialize with reproducible random state.
        """
        self.seed = seed
        # Initialize JAX keys or RNG here

    def improve(self, input_f: np.ndarray) -> np.ndarray:
        """
        Refines an existing function f using rigorous local optimization.
        
        ADAPTIVE RESOLUTION:
        The output array does NOT need to match the input size.

        Args:
            input_f: 1D NumPy array of shape (N_in,)

        Returns:
            improved_f: 1D NumPy array of shape (N_out,)
        """
        pass

    def perturb(self, input_f: np.ndarray, intensity: float) -> np.ndarray:
        """
        Applies DISCRETE, structural, or RESOLUTION modifications.
        Args:
            input_f: 1D NumPy array.
            intensity: Float (1e-3 to 1e3).

        Returns:
            perturbed_f: 1D NumPy array (size may differ from input).
        """
        pass

    def generate_config(self, initial_resolution: int = 1000) -> np.ndarray:
        """
        Generates a valid starting function f with a maximum possible fitness.

        Args:
            initial_resolution: Suggested starting size N.

        Returns:
            f: 1D NumPy array of shape (initial_resolution,)
        """
        pass

def entrypoint():
    return Improver
\end{lstlisting}

\noindent \textbf{CRITICAL FAILURE MODES}
\begin{itemize}
    \item \textbf{Zero Function:} Returning an array of all zeros.
    \item \textbf{Negative Values:} Returning $f(x) < 0$.
    \item \textbf{Dimensionality Errors:} Returning a 2D array.
    \item \textbf{Excessively Low Resolution:} Returning $N < 1000$ where convolution properties degrade significantly.
    \item \textbf{Excessively High Resolution:} Returning $N > 20000$ where optimization is very slow. Your solution must work even for this case, but we will not validate it now, so put in the comments to your program how it should be scaled or program a ``self.scale\_mode'' for your class. 
    \item \textbf{Inefficiency:} Ignoring the benefits of Multigrid strategies (not utilizing lower $N$ for fast initial convergence), using a wrong optimizer, or resorting to extremely large $N$.
    \item \textbf{Delusions:} Trying to ``guess'' optimal function rather than searching for it. Believe me, the example for $C = 0.961$ is a really strange function.
\end{itemize}

\noindent \textbf{DOCUMENTATION}\\
Supply your code with a brief explanation of the algorithm, using \textbf{LaTeX} in the introductory comment.
\end{paracol}

\onecolumn
\subsection{Evolved Solution Code}
\begin{paracol}{2}
\subsubsection{Baseline}
The following program was evolved by \gigaevo\ (Program ID: 175c1208-29b0-4c6a-9323-c73d52804f2f, Fitness: 0.9406, Generation: 18).

\begin{minted}{python}
import numpy as np
import jax
import jax.numpy as jnp
import optax
from jax.image import resize
from scipy.optimize import minimize

jax.config.update('jax_enable_x64', True)

def entrypoint() -> np.ndarray:
    """""
    Maximizes C(f) = ||f*f||_2^2 / (||f*f||_1 * ||f*f||_inf)
    
    Algorithm: 
    A hierarchical multi-scale approach targeting fractal-like function supports. 
    The strategy uses a stochastic pulse-mixture initialization followed by 
    coarse-to-fine optimization (N=1024 up to N=65536) using Adam and L-BFGS-B. 
    The L-infinity norm is approximated by an adaptive p-norm proxy to stabilize 
    gradients across resolution jumps.
    
    $$ C(f) = \frac{\sum |(f \star f)(x)|^2}{(\sum (f \star f)(x)) \cdot \max (f \star f)(x)} $$
    """""

    def get_objective(p, alpha=100.0, exact=False):
        f = jnp.square(p)
        N = f.shape[0]
        n_fft = 2**((2 * N - 1).bit_length())
        f_fft = jnp.fft.rfft(f, n=n_fft)
        g = jnp.fft.irfft(f_fft * f_fft, n=n_fft)[:2*N-1]
        g = jnp.maximum(g, 0.0)
        
        l1 = jnp.sum(f)**2
        l2_sq = jnp.sum(jnp.square(g))
        
        if exact:
            linf = jnp.max(g)
        else:
            # Detached-max p-norm for stable L-infinity approximation
            g_max_stop = jax.lax.stop_gradient(jnp.max(g))
            g_normed = g / (g_max_stop + 1e-18)
            linf = jnp.power(jnp.sum(jnp.power(g_normed, alpha)), 1.0/alpha) * g_max_stop
        
        # Non-negativity and sum constraint are implicit
        return -(l2_sq / (l1 * linf + 1e-15))

    # 1. Stochastic Initialization (Diversity improvement)
    N_base = 1024
    key = jax.random.PRNGKey(np.random.randint(0, 10000))
    k1, k2, k3, k4 = jax.random.split(key, 4)
    
    # Mixture of spikes and low-frequency components
    p = jnp.zeros(N_base)
    spikes = jax.random.choice(k1, N_base, shape=(64,), replace=False)
    p = p.at[spikes].set(jax.random.uniform(k2, (64,)))
    noise = 0.1 * jax.random.normal(k3, (N_base,))
    low_freq = jnp.abs(jnp.fft.irfft(jax.random.normal(k4, (N_base//4,)), n=N_base))
    p = p + noise + low_freq

    # 2. Multi-scale Adam Phase
    # Resolutions and alpha schedules for progressive refinement
    resolutions = [1024, 4096, 16384]
    alphas = [64.0, 128.0, 256.0]
    lrs = [0.01, 0.005, 0.002]
    
    current_p = p
    for res, alpha, base_lr in zip(resolutions, alphas, lrs):
        if current_p.shape[0] != res:
            current_p = resize(current_p, (res,), method='cubic')
        
        optimizer = optax.adam(optax.cosine_decay_schedule(base_lr, 1000))
        opt_state = optimizer.init(current_p)

        @jax.jit
        def train_step(params, state, a):
            loss, grads = jax.value_and_grad(get_objective)(params, a, False)
            updates, next_state = optimizer.update(grads, state, params)
            return optax.apply_updates(params, updates), next_state

        for _ in range(1000):
            current_p, opt_state = train_step(current_p, opt_state, alpha)

    # 3. High-Resolution Polish (L-BFGS-B)
    N_final = 65536
    current_p = resize(current_p, (N_final,), method='cubic')

    @jax.jit
    def val_and_grad_fn(p_in):
        # Use higher alpha for more exactness but still differentiable
        return jax.value_and_grad(get_objective)(p_in, alpha=512.0, exact=False)

    def scipy_interface(p_np):
        p_jax = jnp.array(p_np)
        v, g = val_and_grad_fn(p_jax)
        return np.array(v, dtype=np.float64), np.array(g, dtype=np.float64)

    res_opt = minimize(
        scipy_interface,
        np.array(current_p, dtype=np.float64),
        method='L-BFGS-B', 
        jac=True,
        options={'maxiter': 2500, 'ftol': 1e-12}
    )

    f_final = np.square(res_opt.x)
    f_final = np.maximum(f_final, 0.0)
    v_max = np.max(f_final)
    if v_max > 0:
        f_final /= v_max
    else:
        f_final = np.ones(N_final)

    return f_final.astype(np.float64)
\end{minted}

\switchcolumn
\subsubsection{For Improver}\label{sec:autocorr_improver_code}
The following program was evolved by \improvevolve\ (Program ID: b8dadcb7-a2a7-452d-93ec-386e06ea1cb8, Fitness: 0.9512, Generation: 11). Lines marked with \texttt{EDIT START/END} show the human modifications applied in the \improvevolvepluse\ variant: the multi-grid schedule was scaled up to accommodate \alphaevolve's 50\,000-step solution and progressively refine it up to 1.6 million steps.

\begin{minted}{python}
import numpy as np
import jax
import jax.numpy as jnp
import optax
import jaxopt

# LaTeX: C(f) = \frac{\|f \star f\|_2^2}{\|f \star f\|_1 \|f \star f\|_\infty}
# We use a multi-grid strategy with JAX and an adaptive LogSumExp approximation.
# Resolution scales up to 32768, and Beta scales to 50000 for high-precision convergence.

class Improver:
    def __init__(self, seed: int = 0):
        self.seed = seed
        self.key = jax.random.PRNGKey(seed)

    @staticmethod
    @jax.jit
    def calculate_c_loss(w, beta):
        f = jnp.square(w)
        n = f.shape[0]
        n_fft = 2**((2*n - 1).bit_length())
        f_fft = jnp.fft.rfft(f, n=n_fft)
        g = jnp.fft.irfft(f_fft * f_fft, n=n_fft)[:2*n-1]
        g = jnp.maximum(g, 0.0)
        
        l2_sq = jnp.sum(jnp.square(g))
        l1 = jnp.sum(g) + 1e-10
        
        # Stabilized LogSumExp to approximate L_infinity
        g_max = jnp.max(g)
        linf_proxy = jax.nn.logsumexp(beta * (g / (g_max + 1e-10))) / beta * g_max
        
        # Objective: Maximize C -> Minimize -C
        return -(l2_sq / (l1 * linf_proxy + 1e-10))

    def improve(self, input_f: np.ndarray) -> np.ndarray:
        w = jnp.sqrt(jnp.maximum(jnp.array(input_f), 1e-10))
        
        # Multi-grid schedule: increase resolution and beta (sharpness) concurrently
        # Final stage at 32k resolution with very high beta for precision
        stages = [
            {"res": 4096, "beta": 800, "iters": 1000},
            {"res": 16384, "beta": 5000, "iters": 2000},
            {"res": 32768, "beta": 50000, "iters": 3000}
        ]
        # Our own edit to effectively start from the AlphaEvolve's solution:
        # EDIT START
        # w = jnp.sqrt(jnp.maximum(jnp.array(input_f), 0)) # removed 1e-10
        # stages = [
        #   {"res": max(100000, len(input_f)),  "iters": 10000, "beta": 20000,},
        #   {"res": max(200000, len(input_f)),  "iters": 10000, "beta": 20000,},
        #   {"res": max(400000, len(input_f)),  "iters": 10000, "beta": 20000,},
        #   {"res": max(800000, len(input_f)),  "iters": 10000, "beta": 20000,},
        #   {"res": max(1600000, len(input_f)),  "iters": 10000, "beta": 20000,},
        # ]
        # EDIT END

        for st in stages:
            if len(w) != st["res"]:
                # Use linear resizing to avoid Fourier/Gibbs artifacts
                w = jax.image.resize(w, (st["res"],), method='linear')
            
            def obj_fn(weights):
                return self.calculate_c_loss(weights, st["beta"])

            lbfgs = jaxopt.LBFGS(fun=obj_fn, maxiter=st["iters"], tol=1e-12)
            w = lbfgs.run(w).params

        final_f = np.square(np.array(w))
        return final_f / (np.max(final_f) + 1e-15)

    def perturb(self, input_f: np.ndarray, intensity: float) -> np.ndarray:
        self.key, subkey = jax.random.split(self.key)
        n = len(input_f)
        f = jnp.array(input_f)
        # Randomly roll or apply mirror symmetry to find better global shapes
        if jax.random.uniform(subkey) > 0.5:
            f = jnp.flip(f)
        
        noise = jax.random.uniform(subkey, (n,)) * intensity * 0.1
        f = jnp.maximum(f + noise, 0.0)
        return np.array(f)

    def generate_config(self, initial_resolution: int = 2048) -> np.ndarray:
        # Start with a denser initialization to provide more gradient paths
        n = max(initial_resolution, 1024)
        self.key, subkey1, subkey2 = jax.random.split(self.key, 3)
        
        # Mix of uniform noise and specific pulses
        f = jax.random.uniform(subkey1, (n,), minval=0.0, maxval=0.1)
        spike_pos = jax.random.randint(subkey2, (32,), 0, n)
        f = f.at[spike_pos].set(jax.random.uniform(subkey2, (32,), minval=0.5, maxval=1.0))
        
        # Small smoothing to help initial L-BFGS
        x = jnp.linspace(-2, 2, 5)
        kernel = jnp.exp(-x**2)
        f = jnp.convolve(f, kernel / jnp.sum(kernel), mode='same')
        return np.array(f / (jnp.max(f) + 1e-12))

def entrypoint():
    return Improver
\end{minted}
\end{paracol}

\section{LLM-Based Program Tuning: Prompt}
\label{sec:llm_tuning_prompt}

The following prompt was used to evaluate whether \emph{Gemini~3~Pro} could replicate the expert-guided edits of \improvevolvepluse\ (see \Cref{tab:stage_ablation} for results).

\begin{quote}
\small
Your objective is to improve the program that solves the given mathematical challenge while making only minimal modifications to the existing code.
Allowed changes include adjusting numerical constants, modifying parameter values, or replacing function arguments.

Do not rewrite the program or restructure its logic; prioritize the smallest possible edits that lead to improved performance or correctness.

\medskip
\noindent\textbf{Program:} \textit{[evolved program inserted here]}

\medskip
\noindent\textbf{Challenge:} \textit{[task description inserted here]}
\end{quote}

\section{Spherical Codes: Experimental Details, Diagnostics, and Source Code}
\label{sec:spherical_edits}

This appendix collects the experimental details, distribution diagnostics, dimensionality scan, and source code for the spherical-codes benchmark of \Cref{sec:spherical_codes}.

\subsection{Experimental setup}\label{sec:spherical_setup}
\paragraph{Evolution.}
The spherical-codes \texttt{Improver} was evolved by \gigaevo\ with \emph{Gemini~3.5~Flash} as the proposer LLM on the single configuration $(N,d)=(600,11)$, using the default evolution hyperparameters of \Cref{sec:method}. The program with the highest evolution fitness $-\mu(\mathcal{X})$ was kept as the \emph{LLM-evolved} \texttt{Improver} reported in \Cref{sec:spherical_codes_baseline}. The \improvevolvepluse\ variant (\Cref{sec:spherical_codes_edited}) was obtained by passing the evolved program to \emph{Gemini~3.1~Pro} together with the minimal-edits prompt of \Cref{sec:llm_tuning_prompt}; the resulting changes are concentrated in three places---rewritten upper-triangular LSE / $s$-energy surrogates, an extended LogSumExp continuation with tighter L-BFGS-B options, and a concurrent / geodesic / hemispheric-twist \texttt{perturb}---and are shown as a unified diff in \Cref{sec:spherical_codes_diff}.

\paragraph{Validation against the Cohn catalogue.}
For the main configuration $(N,d)=(600,11)$ the validation pool of $K=10$ Stage-A seeds is replaced by a single starting point: the best-known $(600,11)$ configuration from~\citet{cohnSphericalCodes}, directly exercising the resume-from-input interface. Stage~B then refines it with a geometric $\sigma$ schedule under monotone acceptance ($T=0$), so each round can only reduce $\mu(\mathcal{X})$; the calibrated search budget and restart schedule are those of protocol $P^\star$ (\Cref{sec:spherical_calibrated_search}).

\paragraph{Dimensionality scan.}
For the $d \in [8,16]$ scan in \Cref{sec:spherical_dim_scan} we keep all three \texttt{Improver}s fixed (no re-evolution per~$d$) and evaluate $10$ configurations per dimension with $N$ drawn from $[26, 1021]$, each initialised from the corresponding entry in the catalogue of~\citet{cohnSphericalCodes}. Every program is run under the same calibrated maximum-effort protocol $P^\star$ (\Cref{sec:spherical_calibrated_search}), so the comparison isolates the program from the search budget. The full per-configuration table of $\mu$ values is reported in \Cref{tab:spherical_cohn}; the aggregate plots discussed below treat any non-improvement as zero gain so that the success-rate and average-gain summaries are not skewed by negative values produced when basin-hopping fails to improve upon the catalogue baseline.

\subsection{Calibrated general-improver search}\label{sec:spherical_calibrated_search}
\improvevolvextwo\ is a single \texttt{Improver}, evolved by \gigaevo\ to improve the catalogue across the whole range $d\in[8,16]$ at once rather than for one $(N,d)$, and reported alongside the two single-configuration programs of \Cref{sec:spherical_codes_baseline,sec:spherical_codes_edited} (the \improvevolvextwo\ program is listed in \Cref{sec:spherical_codes_champion}). Its fitness signal during evolution was a \emph{panel} of $14$ high-headroom configurations---those on which the catalogue leaves the most slack as suggested by an LLM; the remaining configurations of the scan are essentially saturated and carry little signal. We calibrated the validation search on this same panel and then applied it, unchanged, to the full configuration benchmark of 90 codes.

\paragraph{Metric.} For every configuration we report the relative reduction of the maximum cosine below the catalogue value, floored at zero,
\[
g(\mathcal{X}) = \max\!\Big(0,\ \tfrac{\mu_{\mathrm{Cohn}}-\mu(\mathcal{X})}{|\mu_{\mathrm{Cohn}}|}\Big),
\qquad
\% = 100\cdot\operatorname*{mean}_{\text{configs}} g(\mathcal{X}),
\]
and the headline number is the mean of $g$ over the configurations. This quantity is computed identically by the evolution grader and by the validation harness; only the \emph{search protocol} that drives the \texttt{improve}/\texttt{perturb} loop differs between them.

\paragraph{Budget is the dominant lever.} The canonical grader, a single monotone basin-hopping chain ($R=1$, $M=10$ noising steps, monotone acceptance, fixed seed), reproduces the plotted panel fitness of ${\approx}\,0.46\%$ when run at the evolution's true per-configuration wall of $3540$\,s ($0.4645\%$), and degrades smoothly as the wall shrinks ($0.4125\%$ at $600$\,s). Fitness is wall-limited and monotone in compute, so the per-configuration budget is the primary thing that buys improvement.

\paragraph{Protocol $P^\star$.} Holding the wall fixed, we swept the basin-hopping knobs on the panel: the number of noising steps $M$ per restart, the perturbation schedule $\sigma_{\max}\!\to\!\sigma_{\min}$, the fresh-restart period, and the restart budget. Fewer noising steps per restart win consistently---a smaller $M$ lets more restarts fit a fixed wall, yielding more independent basin hops---while the fresh-restart period is immaterial. This selects $P^\star$: unbounded restarts, $M=10$, $\sigma:1\!\to\!10^{-6}$, a fresh restart every $5$ rounds, and monotone acceptance ($T=0$). At an identical $3540$\,s per-configuration wall, $P^\star$ scores $0.4797\%$ on the panel against the canonical chain's $0.4645\%$---a $+3.3\%$ relative gain attributable to the search protocol alone.

\paragraph{Full-benchmark evaluation.} The full-$90$ head-to-head of \Cref{tab:spherical_cohn} and \Cref{fig:spherical_dim_box,fig:spherical_dim_success,fig:spherical_dim_gain} applies $P^\star$ to all three programs under the same large budget---$3540$\,s per configuration, best of $3$ random seeds, $185$ parallel workers, so that the program is the only thing that varies. The $61$ configurations with measurable headroom are searched directly; the $29$ saturated ones, on which no program beats Cohn, are carried at the catalogue value (zero gain). Under this protocol \improvevolvextwo\ strictly improves $53/90$ configurations for a mean relative gain of $\mathbf{0.1765\%}$, against $0.0752\%$ ($54/90$) for \improvevolve\ and $0.0618\%$ ($36/90$) for \improvevolvepluse.

\subsection{Inner-product distribution at $(296,16)$}\label{sec:spherical_validation}
\begin{figure}[t]
\centering
\includegraphics[width=\linewidth]{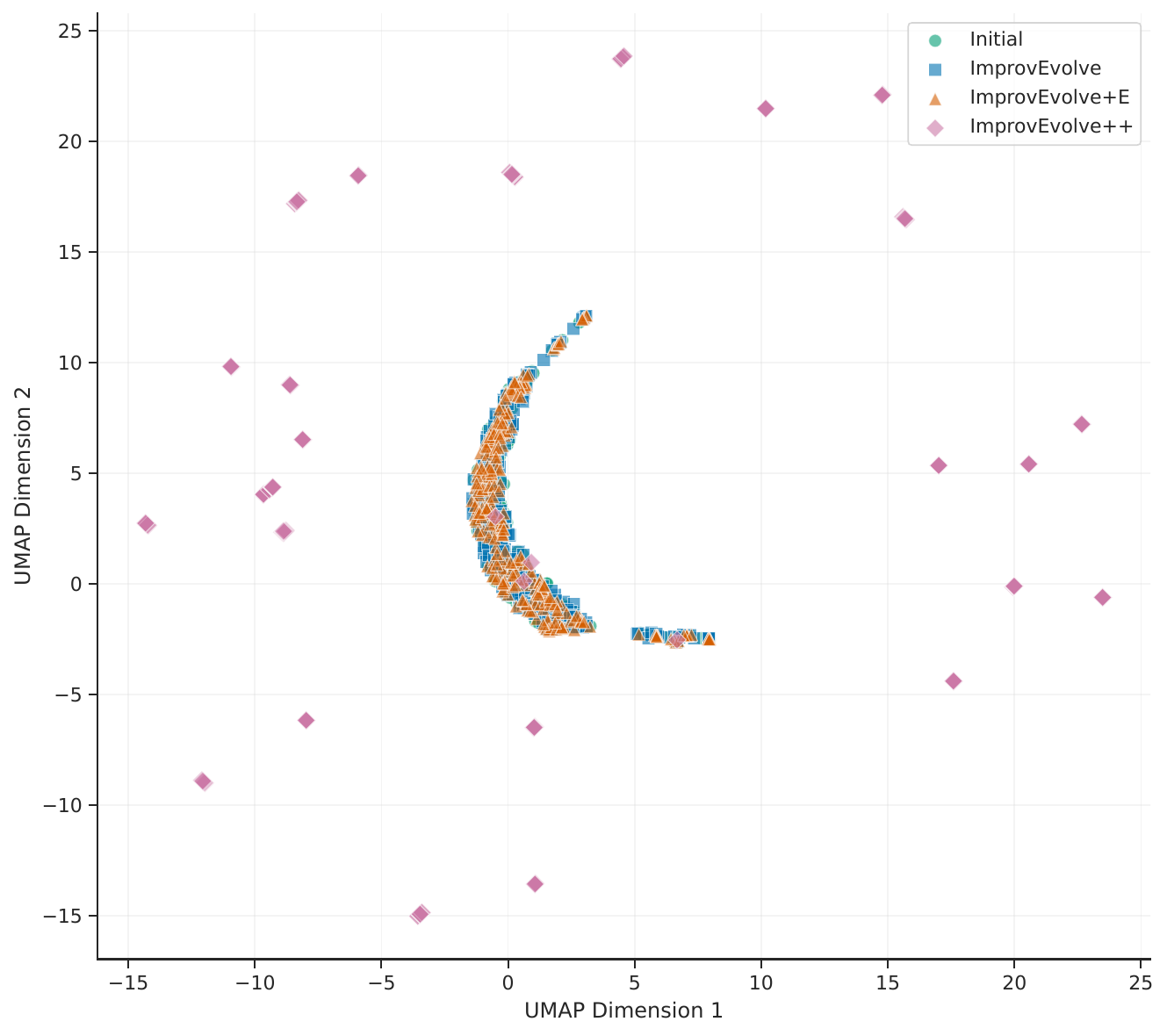}
\caption{UMAP projection of the final $(N,d)=(296,16)$ configurations after validation. \improvevolve\ and \improvevolvepluse\ overlap the Cohn starting configuration almost exactly, whereas \improvevolvextwo\ relocates a substantial fraction of the points into regions of the mapping disjoint from the initial configuration, a hallmark of global rather than local optimization.}
\label{fig:spherical_umap}
\end{figure}

\begin{figure*}[t]
\centering
\begin{subfigure}[b]{0.49\linewidth}
    \centering
    \includegraphics[width=\linewidth]{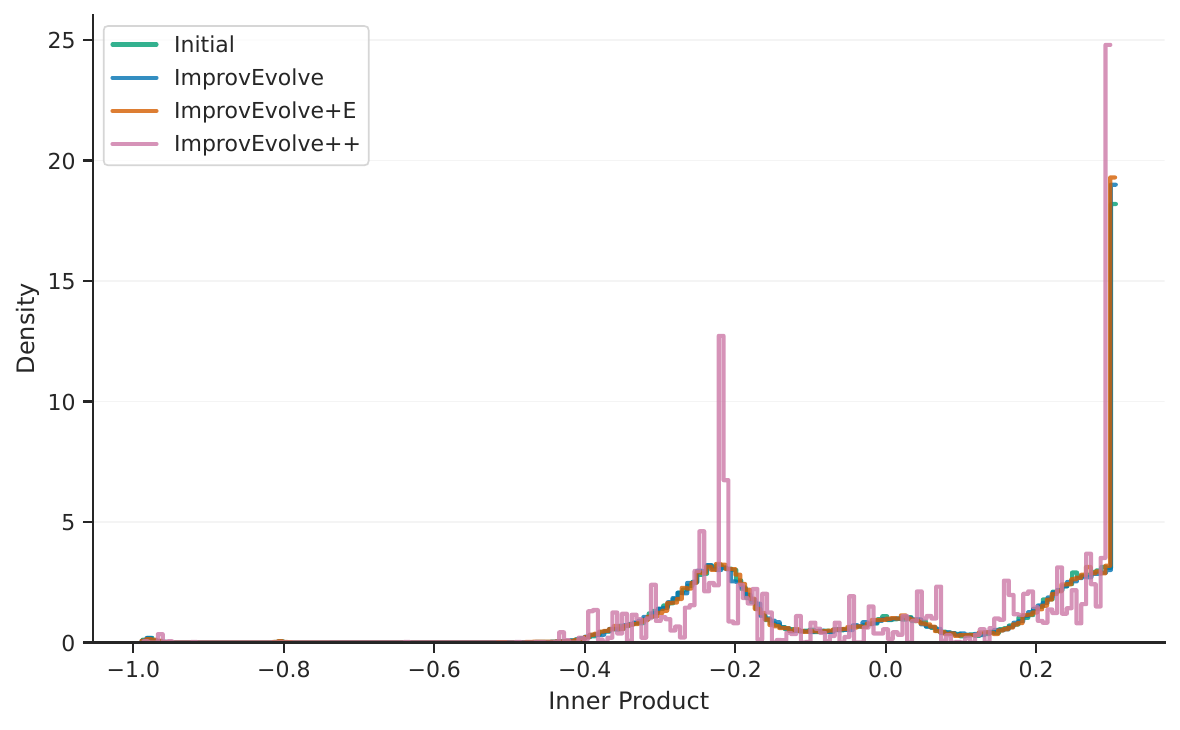}
    \caption{Full pairwise inner-product distribution.}
    \label{fig:spherical_full_dist}
\end{subfigure}
\hfill
\begin{subfigure}[b]{0.49\linewidth}
    \centering
    \includegraphics[width=\linewidth]{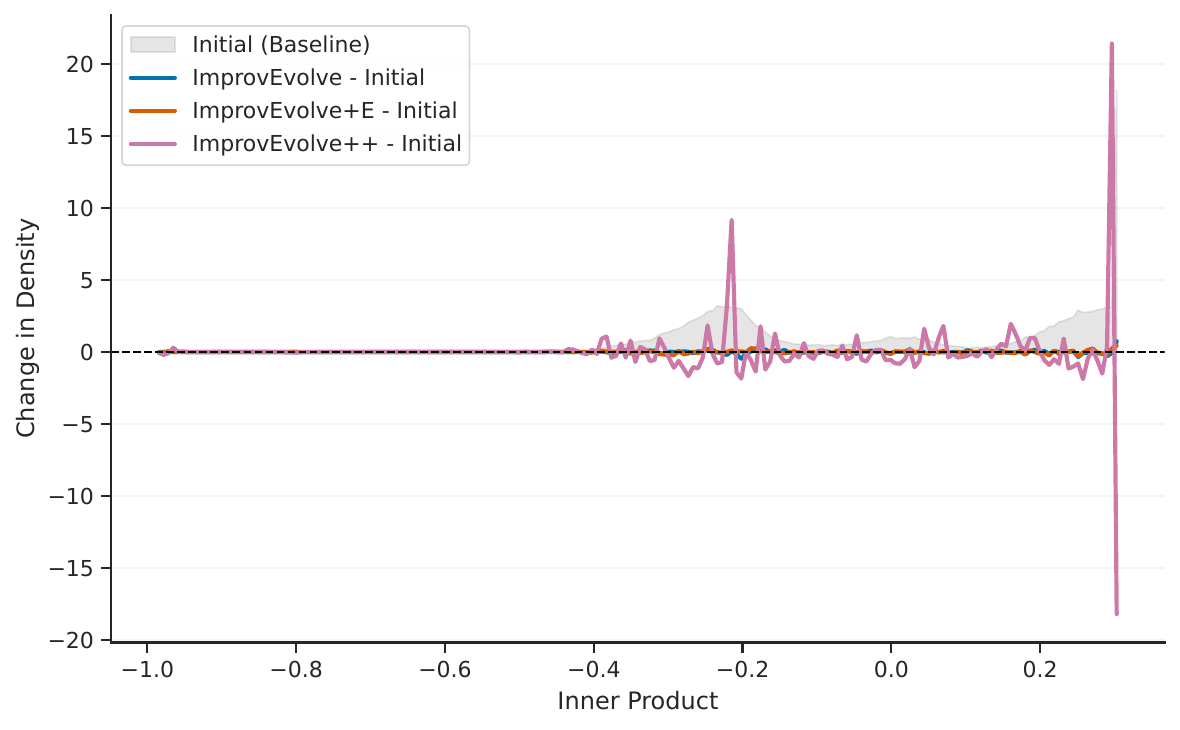}
    \caption{Shift relative to the Cohn baseline.}
    \label{fig:spherical_shift}
\end{subfigure}
\caption{Pairwise inner-product diagnostics at $(N,d)=(296,16)$, validated from the Cohn catalogue. (a)~The full distribution: \improvevolve\ and \improvevolvepluse\ closely follow Cohn, while \improvevolvextwo\ produces a sparser, sharper density distribution---a structural signature of a more discrete code. (b)~Subtracting the Cohn density from each improver curve: positive values mean the improver places more mass than Cohn at that inner-product value, negative values the opposite. The \improvevolve\ and \improvevolvepluse\ differences are negligible, whereas \improvevolvextwo\ removes mass from the catalogue's maximum cosine (near $0.30$) and adds it at smaller inner products.}
\label{fig:spherical_distributions}
\end{figure*}

\Cref{fig:spherical_full_dist} shows the full pairwise inner-product distribution after validation, complementing the right-tail view of \Cref{fig:spherical_main}. \improvevolve\ and \improvevolvepluse\ track the Cohn spectrum almost exactly, consistent with their marginal gains at this configuration; \improvevolvextwo, by contrast, concentrates the inner products onto a sparser, more discrete set of values---a pronounced spike at the (now lower) maximum cosine and a sharp secondary peak at negative inner product---the hallmark of a structured code rather than a generic local optimum. \Cref{fig:spherical_shift} subtracts the Cohn baseline from each curve and makes this explicit: the \improvevolve\ and \improvevolvepluse\ differences are essentially flat, whereas \improvevolvextwo\ removes a substantial block of mass from the catalogue's maximum cosine and redeposits it at smaller inner products, matching the UMAP picture of \Cref{fig:spherical_umap}.

\subsection{Dimensionality scan, $d \in [8,16]$}\label{sec:spherical_dim_scan}
\begin{figure}[t]
\centering
\includegraphics[width=\linewidth]{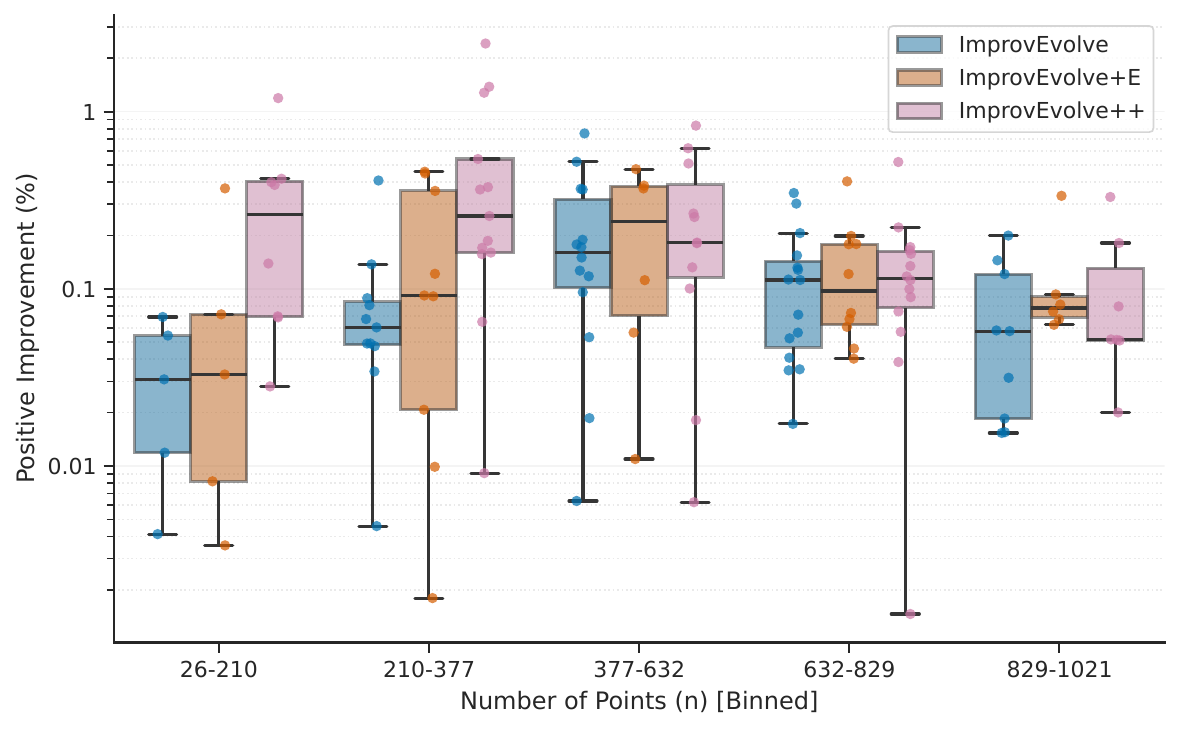}
\caption{Positive improvement on $\mu(\mathcal{X})$ as a function of the number of points $N$, binned into five ranges spanning $[26,1021]$. Validation starts from the Cohn catalogue under the calibrated protocol $P^\star$ (\Cref{sec:spherical_calibrated_search}); the $y$-axis is on a log scale. Configurations on which the improver fails to beat Cohn are scored as zero gain (negative improvements are disregarded).}
\label{fig:spherical_dim_box}
\end{figure}

\begin{figure*}[t]
\centering
\begin{subfigure}[b]{0.49\linewidth}
    \centering
    \includegraphics[width=\linewidth]{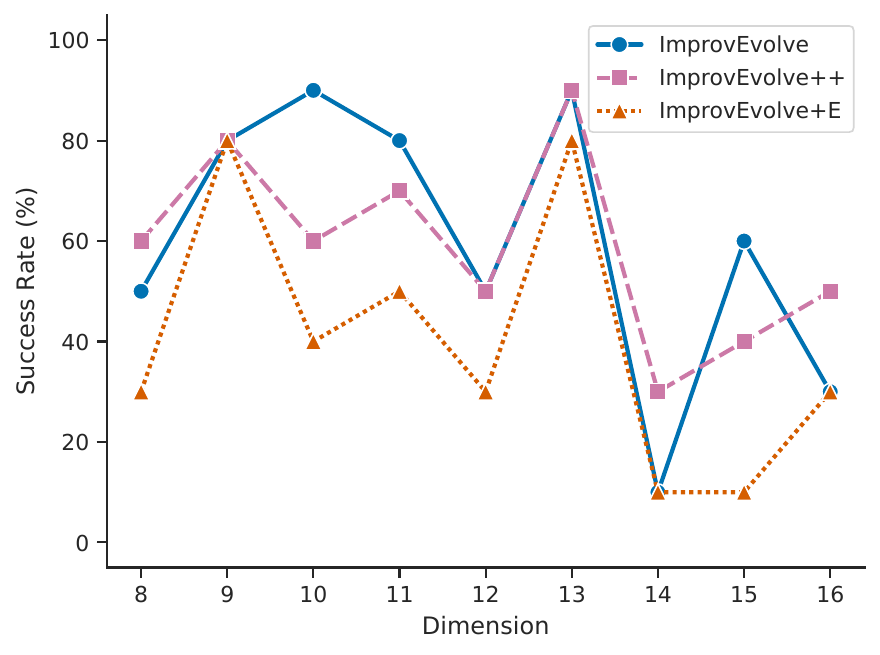}
    \caption{Success rate per dimension.}
    \label{fig:spherical_dim_success}
\end{subfigure}
\hfill
\begin{subfigure}[b]{0.49\linewidth}
    \centering
    \includegraphics[width=\linewidth]{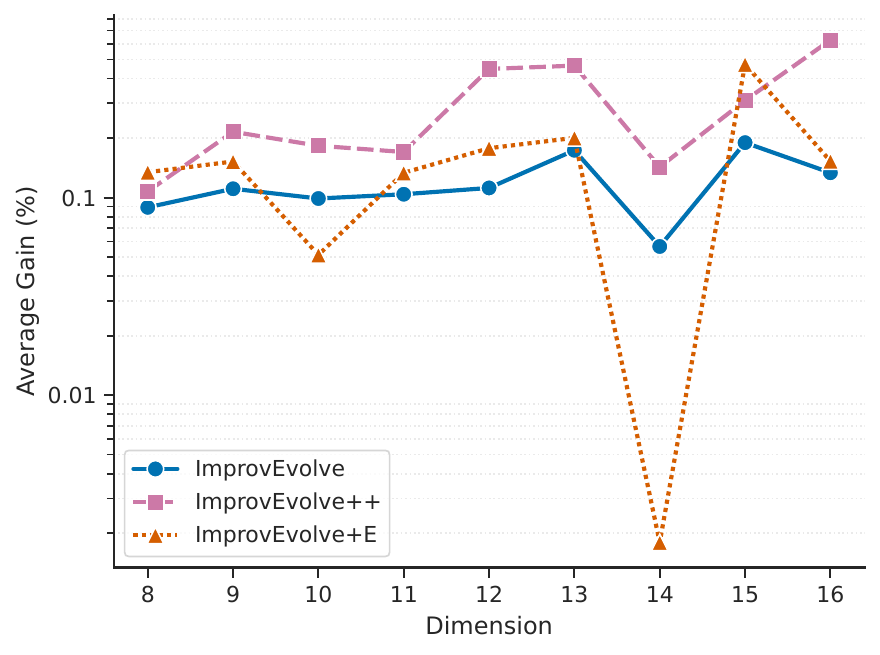}
    \caption{Average gain in $\mu(\mathcal{X})$ per dimension.}
    \label{fig:spherical_dim_gain}
\end{subfigure}
\caption{Dimensionality scan across $d \in [8, 16]$ ($10$ configurations per dimension, validated from the Cohn catalogue under the calibrated protocol $P^\star$). (a)~Success rate: fraction of configurations on which the improver strictly improves on Cohn. (b)~Average gain on $\mu(\mathcal{X})$ in percent. In both panels, non-improvements are scored as zero.}
\label{fig:spherical_dim_summary}
\end{figure*}

\Cref{fig:spherical_dim_box,fig:spherical_dim_success,fig:spherical_dim_gain} aggregate the dimensionality scan described in \Cref{sec:spherical_setup,sec:spherical_calibrated_search} and \Cref{tab:spherical_cohn} lists the corresponding per-configuration $\mu(\mathcal{X})$ values: $90$ configurations in total ($10$ per dimension), all validated from the Cohn catalogue under the calibrated protocol $P^\star$. \Cref{fig:spherical_dim_box} bins the configurations by $N$ and shows the positive-improvement percentage on $\mu(\mathcal{X})$; for \improvevolvextwo\ the gains are largest at small-to-moderate $N$ and shrink as the configurations saturate towards $N=1021$, whereas \improvevolve\ and \improvevolvepluse\ peak in the middle bins. \Cref{fig:spherical_dim_success} reports the success rate per dimension (fraction of configurations on which the improver strictly improves on Cohn); \improvevolve\ and \improvevolvextwo\ have comparable success rates (near $60\%$ averaged over dimensions), both above \improvevolvepluse\ (near $40\%$) at most dimensions. \Cref{fig:spherical_dim_gain} reports the average gain across the same configurations under the convention that non-improvements contribute zero; here \improvevolvextwo\ dominates at nearly every dimension---most decisively at $d=13$ and $d=16$---showing that its advantage lies in the \emph{size} of its improvements rather than how often they fire.

\begingroup
\small     % Apply your smaller font size
\begin{longtable}{cccccc}
\caption{Per-configuration head-to-head of the spherical-codes \texttt{Improver}s against the catalogue of~\citet{cohnSphericalCodes} across $d \in [8,16]$ and $N \in [26, 1021]$ ($10$ configurations per dimension), under the calibrated maximum-effort search protocol $P^\star$ of \Cref{sec:spherical_calibrated_search} (best of $3$ seeds, $3540$\,s wall per configuration), applied identically to every program. \emph{ImprovEvolve++} is the general improver discovered by this protocol. Each cell is the final maximum cosine $\mu(\mathcal{X})$; smaller is better. \textbf{Bold} marks the best of the three improver columns within a row; \underline{underline} marks an improvement over Cohn that is not the best. Configurations on which a variant fails to clear the Cohn baseline are left unmarked.} \label{tab:spherical_cohn} \\
\toprule
D & N & Cohn & ImprovEvolve & ImprovEvolve+E & ImprovEvolve++ \\
\midrule
\endfirsthead

% This section defines what appears at the top of page 2, 3, etc.
\multicolumn{6}{c}{{\bfseries \tablename\ \thetable{} -- Continued from previous page}} \\
\toprule
D & N & Cohn & ImprovEvolve & ImprovEvolve+E & ImprovEvolve++ \\
\midrule
\endhead

% This section defines what appears at the bottom of the page before breaking
\midrule 
\multicolumn{6}{r}{\textit{Continued on next page}} \\
\endfoot

% This section defines the absolute bottom of the final page
\bottomrule
\endlastfoot

% --- TABLE DATA ---
8 & 26 & 0.17265 & 0.17265 & 0.17265 & 0.17265 \\
8 & 51 & 0.31644 & 0.31644 & 0.31644 & 0.31644 \\
8 & 86 & 0.40533 & 0.40533 & 0.40533 & 0.40533 \\
8 & 283 & 0.58281 & 0.58281 & 0.58281 & 0.582\textbf{43} \\
8 & 322 & 0.59702 & 0.59702 & 0.59702 & 0.59702 \\
8 & 528 & 0.65577 & 0.65\textbf{479} & 0.65504 & 0.65\underline{490} \\
8 & 656 & 0.66736 & 0.667\underline{01} & 0.66736 & 0.66\textbf{661} \\
8 & 669 & 0.67014 & 0.66928 & 0.6\textbf{6881} & 0.6\underline{6924} \\
8 & 835 & 0.69934 & 0.69\textbf{893} & 0.69934 & 0.699\underline{20} \\
8 & 873 & 0.70384 & 0.70343 & 0.703\underline{18} & 0.70\textbf{256} \\
9 & 32 & 0.15022 & 0.15022 & 0.15022 & 0.15022 \\
9 & 44 & 0.23355 & 0.23355 & 0.23355 & 0.23355 \\
9 & 521 & 0.59486 & 0.59270 & 0.59\underline{258} & 0.5\textbf{8990} \\
9 & 625 & 0.61687 & 0.61461 & 0.61\underline{460} & 0.61\textbf{373} \\
9 & 750 & 0.64604 & 0.64577 & 0.64\textbf{564} & 0.64\underline{567} \\
9 & 786 & 0.65067 & 0.65044 & 0.6\textbf{4988} & 0.650\underline{09} \\
9 & 838 & 0.65624 & 0.65612 & 0.65\textbf{575} & 0.65\underline{590} \\
9 & 869 & 0.65963 & 0.65942 & 0.659\textbf{09} & 0.659\underline{10} \\
9 & 880 & 0.66063 & 0.66053 & 0.660\textbf{22} & 0.660\underline{29} \\
9 & 886 & 0.66124 & 0.66114 & 0.66\textbf{079} & 0.66\underline{090} \\
10 & 91 & 0.32653 & 0.32635 & 0.326\textbf{29} & 0.326\underline{30} \\
10 & 136 & 0.39348 & 0.39343 & 0.393\underline{35} & 0.39\textbf{293} \\
10 & 271 & 0.48032 & 0.480\underline{16} & 0.48028 & 0.4\textbf{7853} \\
10 & 314 & 0.49205 & 0.49\underline{181} & 0.49205 & 0.49\textbf{127} \\
10 & 324 & 0.49474 & 0.49444 & 0.494\underline{29} & 0.49\textbf{382} \\
10 & 785 & 0.57554 & 0.57\textbf{380} & 0.57554 & 0.57\underline{455} \\
10 & 827 & 0.58832 & 0.58\textbf{766} & 0.58832 & 0.58832 \\
10 & 862 & 0.59325 & 0.59\textbf{253} & 0.59325 & 0.59325 \\
10 & 868 & 0.59411 & 0.59\textbf{325} & 0.59411 & 0.59411 \\
10 & 912 & 0.59829 & 0.59829 & 0.59829 & 0.59829 \\
11 & 84 & 0.26522 & 0.265\textbf{03} & 0.2652\underline{0} & 0.265\textbf{03} \\
11 & 268 & 0.43926 & 0.43\underline{897} & 0.43926 & 0.43\textbf{852} \\
11 & 352 & 0.47738 & 0.47738 & 0.47738 & 0.47738 \\
11 & 373 & 0.48223 & 0.48\underline{180} & 0.48213 & 0.48\textbf{048} \\
11 & 407 & 0.48715 & 0.48\underline{668} & 0.48715 & 0.48\textbf{626} \\
11 & 521 & 0.49965 & 0.49965 & 0.49965 & 0.49965 \\
11 & 590 & 0.49998 & 0.499\underline{71} & 0.499\textbf{69} & 0.49989 \\
11 & 613 & 0.51325 & 0.51\textbf{236} & 0.51325 & 0.51325 \\
11 & 692 & 0.54282 & 0.54198 & 0.54\textbf{185} & 0.54\underline{192} \\
11 & 740 & 0.55428 & 0.55355 & 0.55\textbf{204} & 0.55\underline{305} \\
12 & 95 & 0.26509 & 0.26509 & 0.26509 & 0.26509 \\
12 & 240 & 0.39827 & 0.39\underline{795} & 0.39827 & 0.39\textbf{278} \\
12 & 331 & 0.43098 & 0.43098 & 0.43098 & 0.43098 \\
12 & 343 & 0.43298 & 0.43122 & 0.43\underline{105} & 0.43\textbf{065} \\
12 & 416 & 0.45118 & 0.4511\underline{0} & 0.45118 & 0.45\textbf{036} \\
12 & 642 & 0.49303 & 0.49286 & 0.49\underline{280} & 0.49\textbf{254} \\
12 & 734 & 0.49681 & 0.49672 & 0.496\textbf{61} & 0.496\underline{62} \\
12 & 855 & 0.52037 & 0.52037 & 0.52037 & 0.52037 \\
12 & 899 & 0.53252 & 0.53252 & 0.53252 & 0.53252 \\
12 & 913 & 0.53527 & 0.53527 & 0.53527 & 0.53527 \\
13 & 69 & 0.20843 & 0.20842 & 0.20\underline{766} & 0.20\textbf{759} \\
13 & 90 & 0.24398 & 0.2439\underline{0} & 0.24398 & 0.24\textbf{107} \\
13 & 162 & 0.31368 & 0.31368 & 0.3136\underline{7} & 0.31\textbf{248} \\
13 & 244 & 0.37319 & 0.37301 & 0.37\underline{148} & 0.3\textbf{6843} \\
13 & 351 & 0.41208 & 0.41\underline{152} & 0.41158 & 0.41\textbf{102} \\
13 & 592 & 0.46509 & 0.46\textbf{159} & 0.46509 & 0.46509 \\
13 & 691 & 0.48288 & 0.482\underline{54} & 0.48256 & 0.482\textbf{52} \\
13 & 725 & 0.48554 & 0.48\underline{499} & 0.48518 & 0.48\textbf{497} \\
13 & 818 & 0.49044 & 0.4\textbf{8943} & 0.4\underline{8957} & 0.48967 \\
13 & 882 & 0.49736 & 0.49637 & 0.49\textbf{570} & 0.49\underline{572} \\
14 & 102 & 0.22829 & 0.22829 & 0.22829 & 0.22\textbf{734} \\
14 & 211 & 0.31596 & 0.31596 & 0.31596 & 0.31596 \\
14 & 212 & 0.31640 & 0.31640 & 0.316\underline{39} & 0.316\textbf{37} \\
14 & 642 & 0.43419 & 0.43419 & 0.43419 & 0.4341\textbf{8} \\
14 & 692 & 0.45402 & 0.45\textbf{377} & 0.45402 & 0.45402 \\
14 & 913 & 0.45520 & 0.45520 & 0.45520 & 0.45520 \\
14 & 922 & 0.45535 & 0.45535 & 0.45535 & 0.45535 \\
14 & 925 & 0.45535 & 0.45535 & 0.45535 & 0.45535 \\
14 & 970 & 0.45562 & 0.45562 & 0.45562 & 0.45562 \\
14 & 1021 & 0.45758 & 0.45758 & 0.45758 & 0.45758 \\
15 & 208 & 0.26553 & 0.26553 & 0.26553 & 0.26553 \\
15 & 243 & 0.28637 & 0.28637 & 0.28637 & 0.28637 \\
15 & 380 & 0.36054 & 0.3\underline{5866} & 0.35883 & 0.3\textbf{5830} \\
15 & 425 & 0.37138 & 0.3713\underline{5} & 0.37138 & 0.37\textbf{043} \\
15 & 436 & 0.37646 & 0.37646 & 0.37646 & 0.37646 \\
15 & 456 & 0.37646 & 0.37646 & 0.37646 & 0.37646 \\
15 & 491 & 0.39306 & 0.39\textbf{260} & 0.39306 & 0.39306 \\
15 & 514 & 0.39554 & 0.39\textbf{484} & 0.39554 & 0.395\underline{14} \\
15 & 517 & 0.39605 & 0.39\textbf{555} & 0.39605 & 0.39605 \\
15 & 527 & 0.39763 & 0.39\underline{688} & 0.39763 & 0.39\textbf{657} \\
16 & 88 & 0.16867 & 0.16867 & 0.16867 & 0.16867 \\
16 & 126 & 0.19999 & 0.19999 & 0.19999 & 0.19999 \\
16 & 160 & 0.20498 & 0.20498 & 0.20498 & 0.20498 \\
16 & 197 & 0.24996 & 0.24996 & 0.24996 & 0.249\textbf{89} \\
16 & 296 & 0.30606 & 0.30591 & 0.30\underline{497} & 0.\textbf{29865} \\
16 & 341 & 0.33038 & 0.33037 & 0.330\underline{08} & 0.3\textbf{2985} \\
16 & 534 & 0.36744 & 0.36744 & 0.3674\textbf{0} & 0.3674\underline{1} \\
16 & 688 & 0.39726 & 0.39726 & 0.39726 & 0.39726 \\
16 & 770 & 0.40000 & 0.40000 & 0.40000 & 0.40000 \\
16 & 807 & 0.41152 & 0.41\underline{009} & 0.41152 & 0.4\textbf{0939} \\

\end{longtable}
\endgroup

\subsection{Task description prompt}\label{sec:spherical_task_prompt}
The following prompt was given to \gigaevo\ to evolve the spherical-codes \texttt{Improver}. It defines the objective $\mu(\mathcal{X})$, the class interface, and a non-trivial set of perturbation heuristics for the proposer LLM to draw on (active-set targeting, surrogate relaxations, manifold / subspace moves, intensity orchestration).

\medskip
\noindent\textbf{TASK DEFINITION -- D-DIMENSIONAL SPHERICAL CODE}

\medskip
\noindent\textbf{Challenge:} Implement a Python class that generates a set of $n$ points on the surface of a unit hypersphere in $\mathbb{R}^d$ ($S^{d-1}$) such that the maximum cosine similarity (inner product) between any pair of distinct points is minimized.

\medskip
\noindent\textbf{Specific Benchmark:} $n=600, d=11$.\\
\textbf{General Requirement:} The code must be valid for any natural number $n \geq 2$ and dimension $d \geq 2$.

\medskip
\noindent\textbf{GEOMETRIC SPECIFICATIONS}
\begin{enumerate}
\item \textbf{Points (Items):}
\begin{itemize}
    \item A collection of $n$ points $X = \{x_1, x_2, \dots, x_n\}$ with $x_i \in \mathbb{R}^d$.
\end{itemize}
\item \textbf{Constraint Surface:}
\begin{itemize}
    \item Unit $(d-1)$-sphere: $\|x_i\|_2 = 1$ for all $i = 1, \dots, n$.
    \item Points must lie strictly on the surface of the hypersphere.
\end{itemize}
\item \textbf{Objective:} Minimize the maximum pairwise inner product.
\end{enumerate}

\medskip
\noindent\textbf{OBJECTIVE FUNCTION}\\
Minimize the coherence (maximum cosine similarity)
\[
\mu(X) := \max_{1 \leq i < j \leq n} \langle x_i, x_j \rangle,
\]
i.e.\ produce a spherical code with the smallest possible maximum inner product---points should be as spread out / orthogonal as possible.

\medskip
\noindent\textbf{CONSTRAINTS}
\begin{enumerate}
\item \textbf{Spherical Constraint:} Every returned point must satisfy $\|x_i\| \approx 1.0$ (within floating-point tolerance).
\item \textbf{Count:} Exactly $n$ points must be returned.
\item \textbf{Dimensions:} Points must have exactly $d$ coordinates.
\item \textbf{Real Values:} No \texttt{NaN} or \texttt{Inf} values allowed.
\end{enumerate}

\medskip
\noindent\textbf{TARGETED PERTURBATION \& LANDSCAPE TRAVERSAL (CRITICAL)}\\
A highly competitive \texttt{perturb} method must navigate a notoriously difficult, non-convex energy landscape filled with shallow local minima and saddle points. However, high-quality spherical codes are extremely delicate: applying naive global random noise will destroy the carefully packed sub-structures. Your perturbation must be \emph{surgical, structure-preserving, and highly non-trivial}.

To achieve state-of-the-art performance, your \texttt{perturb} method should act as an orchestrator for a portfolio of advanced heuristics. To inspire your design, consider combining approaches such as:
\begin{itemize}
\item \textbf{Active-Set Targeting:} Identifying the specific subset of points acting as the ``bottleneck'' (the pairs realising the maximum cosine) and applying localised, repelling vector fields or geometric reflections exclusively to them, leaving the well-packed majority intact.
\item \textbf{Surrogate Relaxations:} Temporarily morphing the objective function (e.g.\ evaluating the configuration under LogSumExp, Riesz $s$-energy, or $p$-norms with varying temperatures) to dissolve the current local minimum and let the points smoothly slide into a deeper basin.
\item \textbf{Manifold \& Subspace Moves:} Exploiting the geometry of $S^{d-1}$ via localised topological twists, tangent-space momentum, or subset rotations that alter the configuration's state without destroying the existing invariants.
\item \textbf{Intensity Orchestration:} Using the \texttt{intensity} parameter to intelligently scale from micro-adjustments of bottleneck points to meso-scale structural shifts.
\end{itemize}

\medskip
\noindent\textbf{IMPLEMENTATION REQUIREMENTS}\\
\textbf{Libraries:} \texttt{numpy}, \texttt{scipy}, \texttt{sklearn}, \texttt{jax}, \texttt{optax}, \texttt{jaxopt}, \texttt{optuna}, or standard Python libraries.

\medskip
\noindent\textbf{Class Interface:} The class must implement \texttt{Improver.\_\_init\_\_(n, d, seed=0)}, \texttt{Improver.improve(points, seed=None)}, \texttt{Improver.perturb(points, intensity, seed=None)}, and \texttt{Improver.generate\_config(seed=None)}, together with a module-level \texttt{entrypoint()} that returns the class. The \texttt{improve} method must strictly enforce $\|x_i\| = 1$ in its output; \texttt{perturb} aggregates multiple perturbation heuristics keyed off the \texttt{intensity} argument (in $[0,1]$, higher = larger meso-scale shifts); \texttt{generate\_config} returns a valid initial configuration of $n$ points on $S^{d-1}$.

\medskip
\noindent\textbf{Optimization Context:} \texttt{perturb()} and \texttt{improve()} are called in an iterative loop (e.g.\ $50$ iterations); a $5$-second time limit per \texttt{improve()} call is assumed.

\medskip
\noindent\textbf{CRITICAL FAILURE MODES:} wrong shape (returning $(n,3)$ when $d=4$), constraint violation ($\|x_i\| \neq 1$), and dimension hardcoding (assuming $d=3$ instead of supporting general $d \geq 2$).

\medskip
\noindent\textbf{DOCUMENTATION:} the proposer is asked to supply a brief LaTeX docstring explaining the \texttt{improve} optimization method and the orchestrating \texttt{perturb} logic.

\subsection{Improver class and LLM edits}
The remaining subsections document the \texttt{Improver} class itself. It exposes the three \improvevolve\ operators on $S^{d-1}$: \texttt{improve} runs a LogSumExp continuation on the max-cosine objective $\mu(\mathcal{X})$ with L-BFGS-B; \texttt{perturb} combines active-set repulsion at the current min-distance pairs, Riesz $s$-energy relaxation, and a global rotation step; and \texttt{generate\_config} draws uniformly from $S^{d-1}$ via Gaussian normalisation. JAX provides JIT-compiled value-and-gradient functions for both the LogSumExp and $s$-energy surrogates.

In the \improvevolvepluse\ variant, the LLM edits are concentrated in three places, none of which alter the high-level algorithmic skeleton:
\begin{enumerate}
\item \textbf{Cleaner LSE / $s$-energy formulation.} Both surrogates are rewritten in terms of upper-triangular pair indices and \texttt{jax.nn.logsumexp}, eliminating diagonal masking with \texttt{-jnp.inf} and the manual \mbox{max-subtraction} trick. The change removes a double-counting factor (each pair was counted twice in the original) and reduces the gradient's numerical sensitivity to near-collinear pairs.
\item \textbf{Wider LogSumExp continuation in \texttt{improve}.} The temperature schedule is extended from five steps $[100, \dots, 25600]$ to seven steps $[10, \dots, 40960]$. The softer initial $\alpha$ allows the smooth surrogate to globally untangle the configuration before the sharper stages localise the active-set; the larger terminal $\alpha$ pushes the final surrogate closer to the true minimax. The inner L-BFGS-B budget is also raised (\texttt{maxiter} $500\!\to\!1500$, \texttt{gtol} $10^{-9}\!\to\!10^{-10}$).
\item \textbf{Concurrent, geodesic \texttt{perturb}.} The pair-by-pair sequential repulsion is replaced by a concurrent active-set accumulation (each point sums repulsion contributions from all of its active neighbours), followed by a single tangent-space projection and a geodesic update $x_i \leftarrow x_i\cos\theta + \hat{v}_i\sin\theta$. The structured tangent noise is moved out of the inner loop and applied once per point. The large-intensity branch is changed from a random-subset SO$(d)$ rotation to a \emph{hemispheric twist}: a random hyperplane splits the sphere and the rotation is applied only to one hemisphere, which preserves local cluster structure while breaking global symmetries.
\end{enumerate}

The remainder of this section renders, first, the unified diff between the two programs and, then, the full source of each. In the diff, insertions are shown in green and deletions in red; the diff is line-aligned so that unchanged context appears in plain typewriter font. The three edit clusters listed above are clearly visible in the diff: the rewritten upper-triangular LSE / $s$-energy bodies, the extended \texttt{alphas} continuation schedule with tightened L-BFGS-B options, and the concurrent / geodesic / hemispheric-twist reformulation of \texttt{perturb}.

\subsection{LLM-evolved program}\label{sec:spherical_codes_baseline}
The following program was evolved by \gigaevo\ using \emph{Gemini~3.5~Flash} as the proposer LLM. It is the starting point for the \emph{Gemini 3.1 Pro} edits described above.
\begin{minted}{python}
r"""
=============================================================================
D-DIMENSIONAL SPHERICAL CODE OPTIMIZATION
=============================================================================

Methodology:
We frame the spherical code problem as a continuous min-max optimization
problem over the unit sphere $S^{d-1}$:
    \min_{X \in \mathbb{R}^{n \times d}} \max_{i \neq j} \frac{x_i \cdot x_j}{\|x_i\| \|x_j\|}

Optimization ('improve' method):
We employ a continuation method over the LogSumExp temperature \alpha. By 
progressively increasing \alpha from 100 to 25600, we morph the smooth LSE 
surrogate into the sharp minimax objective. We use L-BFGS-B with tight 
tolerances to ensure convergence in the high-dimensional (d=11) landscape.

Perturbation ('perturb' method):
1. Active-Set Targeting: Surgical repulsion for pairs within a tight 0.005 
   cosine margin of the current maximum.
2. Riesz s-Energy: Meso-scale relaxation using s=d to simulate hard-sphere 
   repulsion dynamics.
3. Subspace Rotations: Random SO(d) rotations applied to random subsets to
   break interlocking chains of points.
=============================================================================
"""

import jax
import jax.numpy as jnp
import numpy as np
import scipy.linalg
import scipy.optimize

jax.config.update("jax_enable_x64", True)

class Improver:
    def __init__(self, n: int, d: int, seed: int = 0):
        self.n = n
        self.d = d
        self.seed = seed
        self.rng = np.random.default_rng(seed)
        n_static = self.n

        @jax.jit
        def objective_lse(V, alpha):
            norms = jnp.linalg.norm(V, axis=1, keepdims=True)
            X = V / jnp.maximum(norms, 1e-8)
            ips = jnp.dot(X, X.T)
            idx = jnp.arange(n_static)
            ips_masked = ips.at[idx, idx].set(-jnp.inf)
            max_ip = jnp.max(ips_masked)
            sum_exp = jnp.sum(jnp.exp(alpha * (ips_masked - max_ip)))
            return max_ip + jnp.log(sum_exp) / alpha

        self.val_grad_lse = jax.jit(jax.value_and_grad(objective_lse, argnums=0))

        @jax.jit
        def s_energy(V, s):
            norms = jnp.linalg.norm(V, axis=1, keepdims=True)
            X = V / jnp.maximum(norms, 1e-8)
            ips = jnp.dot(X, X.T)
            dists_sq = jnp.maximum(2.0 - 2.0 * ips, 1e-10)
            idx = jnp.arange(n_static)
            dists_sq = dists_sq.at[idx, idx].set(jnp.inf)
            return jnp.sum(1.0 / jnp.power(dists_sq, s / 2.0))

        self.val_grad_s_energy = jax.jit(jax.value_and_grad(s_energy, argnums=0))

    def generate_config(self, seed=None) -> np.ndarray:
        rng = np.random.default_rng(seed) if seed is not None else self.rng
        points = rng.standard_normal((self.n, self.d))
        points /= np.linalg.norm(points, axis=1, keepdims=True)
        return points

    def improve(self, points: np.ndarray, seed=None) -> np.ndarray:
        V = points.copy()
        alphas = [100.0, 400.0, 1600.0, 6400.0, 25600.0]
        for alpha in alphas:
            def fun(v_flat):
                v_reshaped = v_flat.reshape((self.n, self.d))
                val, grad = self.val_grad_lse(v_reshaped, float(alpha))
                return float(val), np.array(grad, dtype=np.float64).flatten()

            res = scipy.optimize.minimize(
                fun,
                V.flatten(),
                method="L-BFGS-B",
                jac=True,
                options={"maxiter": 500, "ftol": 1e-12, "gtol": 1e-9},
            )
            V = res.x.reshape((self.n, self.d))
        norms = np.linalg.norm(V, axis=1, keepdims=True)
        return V / norms

    def perturb(self, points: np.ndarray, intensity: float, seed=None) -> np.ndarray:
        rng = np.random.default_rng(seed) if seed is not None else self.rng
        new_points = points.copy()
        ips = np.dot(new_points, new_points.T)
        np.fill_diagonal(ips, -np.inf)
        max_ip = np.max(ips)
        threshold = max_ip - 0.005 * (1.0 + intensity)
        active_pairs = np.argwhere(ips > threshold)

        for i, j in active_pairs:
            if i > j: continue
            direction = new_points[i] - new_points[j]
            norm_dir = np.linalg.norm(direction)
            if norm_dir > 1e-8:
                direction /= norm_dir
                step_size = rng.uniform(0.01, 0.05) * (intensity + 0.1)
                new_points[i] += step_size * direction
                new_points[j] -= step_size * direction
            for idx in (i, j):
                noise = rng.standard_normal(self.d)
                noise -= np.dot(noise, new_points[idx]) * new_points[idx]
                new_points[idx] += rng.uniform(0.005, 0.02) * (intensity + 0.1) * (noise / np.linalg.norm(noise))

        new_points /= np.linalg.norm(new_points, axis=1, keepdims=True)
        s_val = float(self.d) + 20.0 * (1.0 - intensity)
        def fun_s(v_flat):
            v_reshaped = v_flat.reshape((self.n, self.d))
            val, grad = self.val_grad_s_energy(v_reshaped, float(s_val))
            return float(val), np.array(grad, dtype=np.float64).flatten()

        res = scipy.optimize.minimize(fun_s, new_points.flatten(), method="L-BFGS-B", jac=True, options={"maxiter": int(5 + 15 * intensity)})
        new_points = res.x.reshape((self.n, self.d))
        new_points /= np.linalg.norm(new_points, axis=1, keepdims=True)

        if intensity > 0.3:
            num_to_rotate = int(self.n * rng.uniform(0.1, 0.5 * intensity))
            if num_to_rotate > 0:
                subset_indices = rng.choice(self.n, num_to_rotate, replace=False)
                U_random = rng.standard_normal((self.d, self.d))
                Q, _ = np.linalg.qr(U_random)
                A = Q - Q.T
                A /= np.linalg.norm(A) + 1e-8
                theta = rng.uniform(0.1, 0.5) * intensity
                R = scipy.linalg.expm(theta * A)
                new_points[subset_indices] = np.dot(new_points[subset_indices], R.T)
        
        new_points /= np.linalg.norm(new_points, axis=1, keepdims=True)
        return new_points

def entrypoint():
    return Improver

\end{minted}

\subsection{Gemini 3.1 Pro edited program (\improvevolvepluse)}\label{sec:spherical_codes_edited}
The full source of the expert-edited \texttt{Improver} after the three changes of \Cref{sec:spherical_edits} have been applied.
\begin{minted}{python}
r"""
=============================================================================
D-DIMENSIONAL SPHERICAL CODE OPTIMIZATION
=============================================================================

Methodology:
We frame the spherical code problem as a continuous min-max optimization
problem over the unit sphere $S^{d-1}$:
    \min_{X \in \mathbb{R}^{n \times d}} \max_{i \neq j} \frac{x_i \cdot x_j}{\|x_i\| \|x_j\|}

Optimization ('improve' method):
We employ a continuation method over the LogSumExp temperature \alpha. By 
progressively increasing \alpha from 10 to 40960, we morph the smooth LSE 
surrogate into the sharp minimax objective. We use L-BFGS-B with tight 
tolerances. JAX's logsumexp and upper-triangular masking avoid double-counting.

Perturbation ('perturb' method):
1. Active-Set Targeting: Concurrent surgical repulsion mapped to tangent 
   planes to prevent sequential interference.
2. Riesz s-Energy: Meso-scale relaxation using s=d to simulate hard-sphere 
   repulsion dynamics.
3. Subspace Rotations: Tectonic hemispheric twists applied to geometrically 
   separated halves to preserve local sub-structures while breaking global locks.
=============================================================================
"""

import jax
import jax.numpy as jnp
import numpy as np
import scipy.linalg
import scipy.optimize

jax.config.update("jax_enable_x64", True)

class Improver:
    def __init__(self, n: int, d: int, seed: int = 0):
        self.n = n
        self.d = d
        self.seed = seed
        self.rng = np.random.default_rng(seed)
        n_static = self.n

        @jax.jit
        def objective_lse(V, alpha):
            norms = jnp.linalg.norm(V, axis=1, keepdims=True)
            X = V / jnp.maximum(norms, 1e-8)
            ips = jnp.dot(X, X.T)
            # Use upper triangular indices to strictly avoid double-counting pairs
            i, j = jnp.triu_indices(n_static, k=1)
            return jax.nn.logsumexp(alpha * ips[i, j]) / alpha

        self.val_grad_lse = jax.jit(jax.value_and_grad(objective_lse, argnums=0))

        @jax.jit
        def s_energy(V, s):
            norms = jnp.linalg.norm(V, axis=1, keepdims=True)
            X = V / jnp.maximum(norms, 1e-8)
            ips = jnp.dot(X, X.T)
            i, j = jnp.triu_indices(n_static, k=1)
            dists_sq = jnp.maximum(2.0 - 2.0 * ips[i, j], 1e-10)
            return jnp.sum(1.0 / jnp.power(dists_sq, s / 2.0))

        self.val_grad_s_energy = jax.jit(jax.value_and_grad(s_energy, argnums=0))

    def generate_config(self, seed=None) -> np.ndarray:
        rng = np.random.default_rng(seed) if seed is not None else self.rng
        points = rng.standard_normal((self.n, self.d))
        points /= np.linalg.norm(points, axis=1, keepdims=True)
        return points

    def improve(self, points: np.ndarray, seed=None) -> np.ndarray:
        V = points.copy()
        # Softer start schedule to globally untangle points early on
        alphas = [10.0, 40.0, 160.0, 640.0, 2560.0, 10240.0, 40960.0]

        for alpha in alphas:
            def fun(v_flat):
                v_reshaped = v_flat.reshape((self.n, self.d))
                val, grad = self.val_grad_lse(v_reshaped, float(alpha))
                return float(val), np.array(grad, dtype=np.float64).flatten()

            res = scipy.optimize.minimize(
                fun,
                V.flatten(),
                method="L-BFGS-B",
                jac=True,
                options={"maxiter": 1500, "ftol": 1e-12, "gtol": 1e-10},
            )
            V = res.x.reshape((self.n, self.d))
        norms = np.linalg.norm(V, axis=1, keepdims=True)
        return V / norms

    def perturb(self, points: np.ndarray, intensity: float, seed=None) -> np.ndarray:
        rng = np.random.default_rng(seed) if seed is not None else self.rng
        new_points = points.copy()
        ips = np.dot(new_points, new_points.T)
        np.fill_diagonal(ips, -2.0)
        max_ip = np.max(ips)
        threshold = max_ip - 0.02 * (1.0 + intensity)
        active_mask = ips > threshold

        # Concurrent Active-Set accumulation to prevent sequential chaos
        repulsions = np.zeros_like(new_points)
        for i in range(self.n):
            neighbors = new_points[active_mask[i]]
            if len(neighbors) > 0:
                repulsions[i] = np.sum(new_points[i] - neighbors, axis=0)

        # Apply tangent-projected geodesic steps
        for i in range(self.n):
            if np.any(repulsions[i]):
                direction = repulsions[i]
                direction -= np.dot(direction, new_points[i]) * new_points[i]
                norm_dir = np.linalg.norm(direction)
                if norm_dir > 1e-8:
                    direction /= norm_dir
                    step_size = rng.uniform(0.01, 0.1) * intensity
                    new_points[i] = new_points[i] * np.cos(step_size) + direction * np.sin(step_size)
                    
        # Apply structured stochastic noise outside the active pairs loop
        noise = rng.standard_normal((self.n, self.d))
        for i in range(self.n):
            n_vec = noise[i] - np.dot(noise[i], new_points[i]) * new_points[i]
            n_norm = np.linalg.norm(n_vec)
            if n_norm > 1e-8:
                new_points[i] += rng.uniform(0.005, 0.02) * (intensity + 0.1) * (n_vec / n_norm)

        new_points /= np.linalg.norm(new_points, axis=1, keepdims=True)
        
        s_val = float(self.d) + 20.0 * (1.0 - intensity)
        def fun_s(v_flat):
            v_reshaped = v_flat.reshape((self.n, self.d))
            val, grad = self.val_grad_s_energy(v_reshaped, float(s_val))
            return float(val), np.array(grad, dtype=np.float64).flatten()

        res = scipy.optimize.minimize(fun_s, new_points.flatten(), method="L-BFGS-B", jac=True, options={"maxiter": int(5 + 15 * intensity)})
        new_points = res.x.reshape((self.n, self.d))
        new_points /= np.linalg.norm(new_points, axis=1, keepdims=True)

        if intensity > 0.3:
            # Tectonic Hemispheric Twist (Replaces fully random scatter)
            split_v = rng.standard_normal(self.d)
            split_v /= np.linalg.norm(split_v)
            subset_indices = np.where(np.dot(new_points, split_v) > 0)[0]
            
            if len(subset_indices) > 0:
                U_random = rng.standard_normal((self.d, self.d))
                Q, _ = np.linalg.qr(U_random)
                A = Q - Q.T
                A /= np.linalg.norm(A) + 1e-8
                theta = rng.uniform(0.1, 0.5) * intensity
                R = scipy.linalg.expm(theta * A)
                new_points[subset_indices] = np.dot(new_points[subset_indices], R.T)
        
        new_points /= np.linalg.norm(new_points, axis=1, keepdims=True)
        return new_points

def entrypoint():
    return Improver
\end{minted}

\subsection{Unified diff (\improvevolve\ vs.\ \improvevolvepluse)}\label{sec:spherical_codes_diff}
\begin{minted}{diff}
--- spherical_codes_improver_baseline.py
+++ spherical_codes_improver_edited.py
@@ -10,17 +10,17 @@
 
 Optimization ('improve' method):
 We employ a continuation method over the LogSumExp temperature \alpha. By 
-progressively increasing \alpha from 100 to 25600, we morph the smooth LSE 
+progressively increasing \alpha from 10 to 40960, we morph the smooth LSE 
 surrogate into the sharp minimax objective. We use L-BFGS-B with tight 
-tolerances to ensure convergence in the high-dimensional (d=11) landscape.
+tolerances. JAX's logsumexp and upper-triangular masking avoid double-counting.
 
 Perturbation ('perturb' method):
-1. Active-Set Targeting: Surgical repulsion for pairs within a tight 0.005 
-   cosine margin of the current maximum.
+1. Active-Set Targeting: Concurrent surgical repulsion mapped to tangent 
+   planes to prevent sequential interference.
 2. Riesz s-Energy: Meso-scale relaxation using s=d to simulate hard-sphere 
    repulsion dynamics.
-3. Subspace Rotations: Random SO(d) rotations applied to random subsets to
-   break interlocking chains of points.
+3. Subspace Rotations: Tectonic hemispheric twists applied to geometrically 
+   separated halves to preserve local sub-structures while breaking global locks.
 =============================================================================
 """
 
@@ -45,11 +45,9 @@
             norms = jnp.linalg.norm(V, axis=1, keepdims=True)
             X = V / jnp.maximum(norms, 1e-8)
             ips = jnp.dot(X, X.T)
-            idx = jnp.arange(n_static)
-            ips_masked = ips.at[idx, idx].set(-jnp.inf)
-            max_ip = jnp.max(ips_masked)
-            sum_exp = jnp.sum(jnp.exp(alpha * (ips_masked - max_ip)))
-            return max_ip + jnp.log(sum_exp) / alpha
+            # Use upper triangular indices to strictly avoid double-counting pairs
+            i, j = jnp.triu_indices(n_static, k=1)
+            return jax.nn.logsumexp(alpha * ips[i, j]) / alpha
 
         self.val_grad_lse = jax.jit(jax.value_and_grad(objective_lse, argnums=0))
 
@@ -58,9 +56,8 @@
             norms = jnp.linalg.norm(V, axis=1, keepdims=True)
             X = V / jnp.maximum(norms, 1e-8)
             ips = jnp.dot(X, X.T)
-            dists_sq = jnp.maximum(2.0 - 2.0 * ips, 1e-10)
-            idx = jnp.arange(n_static)
-            dists_sq = dists_sq.at[idx, idx].set(jnp.inf)
+            i, j = jnp.triu_indices(n_static, k=1)
+            dists_sq = jnp.maximum(2.0 - 2.0 * ips[i, j], 1e-10)
             return jnp.sum(1.0 / jnp.power(dists_sq, s / 2.0))
 
         self.val_grad_s_energy = jax.jit(jax.value_and_grad(s_energy, argnums=0))
@@ -73,7 +70,9 @@
 
     def improve(self, points: np.ndarray, seed=None) -> np.ndarray:
         V = points.copy()
-        alphas = [100.0, 400.0, 1600.0, 6400.0, 25600.0]
+        # Softer start schedule to globally untangle points early on
+        alphas = [10.0, 40.0, 160.0, 640.0, 2560.0, 10240.0, 40960.0]
+
         for alpha in alphas:
             def fun(v_flat):
                 v_reshaped = v_flat.reshape((self.n, self.d))
@@ -85,7 +84,7 @@
                 V.flatten(),
                 method="L-BFGS-B",
                 jac=True,
-                options={"maxiter": 500, "ftol": 1e-12, "gtol": 1e-9},
+                options={"maxiter": 1500, "ftol": 1e-12, "gtol": 1e-10},
             )
             V = res.x.reshape((self.n, self.d))
         norms = np.linalg.norm(V, axis=1, keepdims=True)
@@ -95,26 +94,39 @@
         rng = np.random.default_rng(seed) if seed is not None else self.rng
         new_points = points.copy()
         ips = np.dot(new_points, new_points.T)
-        np.fill_diagonal(ips, -np.inf)
+        np.fill_diagonal(ips, -2.0)
         max_ip = np.max(ips)
-        threshold = max_ip - 0.005 * (1.0 + intensity)
-        active_pairs = np.argwhere(ips > threshold)
+        threshold = max_ip - 0.02 * (1.0 + intensity)
+        active_mask = ips > threshold
 
-        for i, j in active_pairs:
-            if i > j: continue
-            direction = new_points[i] - new_points[j]
-            norm_dir = np.linalg.norm(direction)
-            if norm_dir > 1e-8:
-                direction /= norm_dir
-                step_size = rng.uniform(0.01, 0.05) * (intensity + 0.1)
-                new_points[i] += step_size * direction
-                new_points[j] -= step_size * direction
-            for idx in (i, j):
-                noise = rng.standard_normal(self.d)
-                noise -= np.dot(noise, new_points[idx]) * new_points[idx]
-                new_points[idx] += rng.uniform(0.005, 0.02) * (intensity + 0.1) * (noise / np.linalg.norm(noise))
+        # Concurrent Active-Set accumulation to prevent sequential chaos
+        repulsions = np.zeros_like(new_points)
+        for i in range(self.n):
+            neighbors = new_points[active_mask[i]]
+            if len(neighbors) > 0:
+                repulsions[i] = np.sum(new_points[i] - neighbors, axis=0)
 
+        # Apply tangent-projected geodesic steps
+        for i in range(self.n):
+            if np.any(repulsions[i]):
+                direction = repulsions[i]
+                direction -= np.dot(direction, new_points[i]) * new_points[i]
+                norm_dir = np.linalg.norm(direction)
+                if norm_dir > 1e-8:
+                    direction /= norm_dir
+                    step_size = rng.uniform(0.01, 0.1) * intensity
+                    new_points[i] = new_points[i] * np.cos(step_size) + direction * np.sin(step_size)
+                    
+        # Apply structured stochastic noise outside the active pairs loop
+        noise = rng.standard_normal((self.n, self.d))
+        for i in range(self.n):
+            n_vec = noise[i] - np.dot(noise[i], new_points[i]) * new_points[i]
+            n_norm = np.linalg.norm(n_vec)
+            if n_norm > 1e-8:
+                new_points[i] += rng.uniform(0.005, 0.02) * (intensity + 0.1) * (n_vec / n_norm)
+
         new_points /= np.linalg.norm(new_points, axis=1, keepdims=True)
+        
         s_val = float(self.d) + 20.0 * (1.0 - intensity)
         def fun_s(v_flat):
             v_reshaped = v_flat.reshape((self.n, self.d))
@@ -126,9 +138,12 @@
         new_points /= np.linalg.norm(new_points, axis=1, keepdims=True)
 
         if intensity > 0.3:
-            num_to_rotate = int(self.n * rng.uniform(0.1, 0.5 * intensity))
-            if num_to_rotate > 0:
-                subset_indices = rng.choice(self.n, num_to_rotate, replace=False)
+            # Tectonic Hemispheric Twist (Replaces fully random scatter)
+            split_v = rng.standard_normal(self.d)
+            split_v /= np.linalg.norm(split_v)
+            subset_indices = np.where(np.dot(new_points, split_v) > 0)[0]
+            
+            if len(subset_indices) > 0:
                 U_random = rng.standard_normal((self.d, self.d))
                 Q, _ = np.linalg.qr(U_random)
                 A = Q - Q.T
@@ -142,4 +157,3 @@
 
 def entrypoint():
     return Improver
-
\end{minted}

\subsection{Calibrated general improver (\improvevolvextwo)}\label{sec:spherical_codes_champion}
\improvevolvextwo\ is a distinct evolved program---not an edit of the two above---discovered by the general-improver run and search calibration of \Cref{sec:spherical_calibrated_search}. Its \texttt{improve} runs an eight-stage LogSumExp continuation on the \emph{geodesic} distances between points (rather than on the raw inner products), annealing the surrogate temperature $\alpha$ from $20$ to $1.5\times10^{5}$ with a tangent-space-projected L-BFGS-B at each stage. Its \texttt{perturb} combines a hemispheric $\mathrm{SO}(d)$ twist, a concurrent geodesic repulsion of the active set, a worst-node destroy-and-greedily-reinsert step, and a Riesz $s$-energy polish. The full source follows.
\begin{minted}{python}
import jax
import jax.numpy as jnp
import numpy as np
import scipy.linalg
import scipy.optimize

jax.config.update("jax_enable_x64", True)


class Improver:
    def __init__(self, n: int, d: int, seed: int = 0):
        self.n = n
        self.d = d
        self.seed = seed
        self.rng = np.random.default_rng(seed)
        n_static = self.n

        @jax.jit
        def objective_lse(V, alpha):
            norms = jnp.linalg.norm(V, axis=1, keepdims=True)
            X = V / jnp.maximum(norms, 1e-8)
            ips = jnp.dot(X, X.T)
            i, j = jnp.triu_indices(n_static, k=1)
            clamped_ips = jnp.clip(ips[i, j], -1.0 + 1e-12, 1.0 - 1e-12)
            geodesic_dists = jnp.arccos(clamped_ips)
            return jax.nn.logsumexp(-alpha * geodesic_dists) / alpha

        self.val_grad_lse = jax.jit(jax.value_and_grad(objective_lse, argnums=0))

        @jax.jit
        def s_energy(V, s):
            norms = jnp.linalg.norm(V, axis=1, keepdims=True)
            X = V / jnp.maximum(norms, 1e-8)
            ips = jnp.dot(X, X.T)
            i, j = jnp.triu_indices(n_static, k=1)
            dists_sq = jnp.maximum(2.0 - 2.0 * ips[i, j], 1e-10)
            return jnp.sum(1.0 / jnp.power(dists_sq, s / 2.0))

        self.val_grad_s_energy = jax.jit(jax.value_and_grad(s_energy, argnums=0))

    def generate_config(self, seed=None) -> np.ndarray:
        rng = np.random.default_rng(seed) if seed is not None else self.rng
        points = rng.standard_normal((self.n, self.d))
        points /= np.linalg.norm(points, axis=1, keepdims=True)
        return points

    def improve(self, points: np.ndarray, seed=None) -> np.ndarray:
        V = points.copy()
        V /= np.linalg.norm(V, axis=1, keepdims=True)

        def compute_mu(X):
            ips = X @ X.T
            np.fill_diagonal(ips, -1.0)
            return float(np.max(ips))

        best_V = V.copy()
        best_mu = compute_mu(best_V)

        def make_callback():
            nonlocal best_V, best_mu

            def callback(xk):
                nonlocal best_V, best_mu
                v_reshaped = xk.reshape((self.n, self.d))
                norms = np.linalg.norm(v_reshaped, axis=1, keepdims=True)
                v_normed = v_reshaped / np.maximum(norms, 1e-8)
                current_mu = compute_mu(v_normed)
                if current_mu < best_mu:
                    best_mu = current_mu
                    best_V = v_normed.copy()

            return callback

        alphas = np.geomspace(20.0, 150000.0, num=8)
        maxiters = [200, 300, 400, 500, 800, 1000, 1200, 1500]
        for alpha, maxiter in zip(alphas, maxiters):

            def fun(v_flat):
                v_reshaped = v_flat.reshape((self.n, self.d))
                val, grad = self.val_grad_lse(v_reshaped, float(alpha))
                grad = np.array(grad, dtype=np.float64)
                norms = np.linalg.norm(v_reshaped, axis=1, keepdims=True)
                v_normed = v_reshaped / np.maximum(norms, 1e-8)
                grad_proj = (
                    grad - np.sum(grad * v_normed, axis=1, keepdims=True) * v_normed
                )
                return float(val), grad_proj.flatten()

            res = scipy.optimize.minimize(
                fun,
                V.flatten(),
                method="L-BFGS-B",
                jac=True,
                callback=make_callback(),
                options={"maxiter": maxiter, "ftol": 1e-12, "gtol": 1e-10},
            )
            V = res.x.reshape((self.n, self.d))
            V = V / np.linalg.norm(V, axis=1, keepdims=True)
            current_mu = compute_mu(V)
            if current_mu < best_mu:
                best_mu = current_mu
                best_V = V.copy()

        return best_V

    def perturb(self, points: np.ndarray, intensity: float, seed=None) -> np.ndarray:
        rng = np.random.default_rng(seed) if seed is not None else self.rng
        new_points = points.copy()
        new_points /= np.linalg.norm(new_points, axis=1, keepdims=True)

        if intensity > 0.05:
            split_v = rng.standard_normal(self.d)
            split_v /= np.linalg.norm(split_v)
            subset_indices = np.where(np.dot(new_points, split_v) > 0)[0]
            if len(subset_indices) > 0:
                U_random = rng.standard_normal((self.d, self.d))
                Q, _ = np.linalg.qr(U_random)
                A = Q - Q.T
                A /= np.linalg.norm(A) + 1e-8
                theta = rng.uniform(0.1, 0.5) * intensity
                R = scipy.linalg.expm(theta * A)
                new_points[subset_indices] = np.dot(new_points[subset_indices], R.T)
            new_points /= np.linalg.norm(new_points, axis=1, keepdims=True)

        ips = np.dot(new_points, new_points.T)
        np.fill_diagonal(ips, -2.0)
        max_ip = np.max(ips)
        weights = np.exp((10.0 + 50.0 * intensity) * (ips - max_ip))
        np.fill_diagonal(weights, 0.0)

        repulsions = np.zeros_like(new_points)
        for i in range(self.n):
            diffs = (
                np.dot(new_points, new_points[i])[:, np.newaxis] * new_points[i]
                - new_points
            )
            repulsions[i] = np.sum(weights[i][:, np.newaxis] * diffs, axis=0)

        for i in range(self.n):
            if np.any(repulsions[i]):
                direction = repulsions[i]
                direction -= np.dot(direction, new_points[i]) * new_points[i]
                norm_dir = np.linalg.norm(direction)
                if norm_dir > 1e-8:
                    direction /= norm_dir
                    step_size = rng.uniform(0.01, 0.1) * intensity
                    new_points[i] = new_points[i] * np.cos(
                        step_size
                    ) + direction * np.sin(step_size)

        new_points /= np.linalg.norm(new_points, axis=1, keepdims=True)

        num_destroy = max(1, int(self.n * 0.15 * intensity))
        ips_temp = np.dot(new_points, new_points.T)
        np.fill_diagonal(ips_temp, -2.0)
        max_ips_per_node = np.max(ips_temp, axis=1)
        worst_nodes = np.argsort(max_ips_per_node)[::-1]
        destroy_idx = worst_nodes[:num_destroy]
        keep_idx = worst_nodes[num_destroy:]

        num_candidates = max(200, self.n * 5)
        candidates = rng.standard_normal((num_candidates, self.d))
        candidates /= np.linalg.norm(candidates, axis=1, keepdims=True)

        selected_points = list(new_points[keep_idx])
        for _ in range(num_destroy):
            current_points = np.array(selected_points)
            ips_candidates = np.dot(candidates, current_points.T)
            max_ips = np.max(ips_candidates, axis=1)
            best_idx = np.argmin(max_ips)
            selected_points.append(candidates[best_idx])
            candidates = np.delete(candidates, best_idx, axis=0)

        new_points[destroy_idx] = np.array(selected_points[-num_destroy:])
        new_points /= np.linalg.norm(new_points, axis=1, keepdims=True)

        s_val = float(self.d) + 5.0 * (1.0 - intensity)

        def fun_s(v_flat):
            v_reshaped = v_flat.reshape((self.n, self.d))
            val, grad = self.val_grad_s_energy(v_reshaped, float(s_val))
            grad = np.array(grad, dtype=np.float64)
            norms = np.linalg.norm(v_reshaped, axis=1, keepdims=True)
            v_normed = v_reshaped / np.maximum(norms, 1e-8)
            grad_proj = grad - np.sum(grad * v_normed, axis=1, keepdims=True) * v_normed
            return float(val), grad_proj.flatten()

        res = scipy.optimize.minimize(
            fun_s,
            new_points.flatten(),
            method="L-BFGS-B",
            jac=True,
            options={"maxiter": int(100 + 150 * intensity)},
        )
        new_points = res.x.reshape((self.n, self.d))
        new_points /= np.linalg.norm(new_points, axis=1, keepdims=True)
        return new_points


def entrypoint():
    return Improver
\end{minted}

\end{document}